%% file: top.tex
\documentclass[runningheads]{llncs}
\usepackage{graphicx}
\usepackage{amsmath,amssymb} 
\usepackage{mathrsfs} 
\usepackage{hyperref}
\usepackage{booktabs}
\usepackage{appendix}
\usepackage{subcaption}

\usepackage{color}
\usepackage[width=122mm,left=12mm,paperwidth=146mm,height=193mm,top=12mm,paperheight=217mm]{geometry}
\begin{document}

\pagestyle{headings}
\mainmatter

\title{Eigendecomposition-free Training of Deep Networks with Zero Eigenvalue-based Losses} 
\titlerunning{ }

\authorrunning{ }

\author{Zheng Dang\textsuperscript{1}, Kwang Moo Yi\textsuperscript{2}, Yinlin Hu\textsuperscript{3}, \\
Fei Wang\textsuperscript{1}, Pascal Fua\textsuperscript{3}, Mathieu Salzmann\textsuperscript{3}}
\institute{\textsuperscript{1}IAIR, Xi'an Jiaotong University, China\\
			\textsuperscript{2}University of Victoria, Canada\\
			\textsuperscript{3}CVLab, EPFL, Switzerland\\
			\vspace{1em}
			{\tt\small{dangzheng713@stu.xjtu.edu.cn, kyi@uvic.ca, wfx@mail.xjtu.edu.cn,\\
			\{yinlin.hu, pascal.fua, mathieu.salzmann\}@epfl.ch}}
}

\input{tex/defs}

\maketitle

\input{tex/abstract}
\input{tex/intro}
\input{tex/motivation}
\input{tex/related}
\input{tex/methodology}
\input{tex/experiments}
\input{tex/conclusion}
\input{tex/acknowledgement}
\clearpage

\bibliographystyle{splncs}
\bibliography{short,vision,learning,optim}
\clearpage

\input{tex/supplementary}

\end{document}

%% file: tex/defs.tex

\newif\ifdraft
\draftfalse
\drafttrue

\definecolor{orange}{rgb}{1,0.5,0}
\definecolor{violet}{RGB}{70,0,170}

\ifdraft
 \newcommand{\PF}[1]{{\color{red}{\bf PF: #1}}}
 \newcommand{\pf}[1]{{\color{red} #1}}
 \newcommand{\KY}[1]{{\color{blue}{\bf KY: #1}}}
 \newcommand{\ky}[1]{{\color{blue} #1}}
 \newcommand{\MS}[1]{{\color{green}{\bf MS: #1}}}
 \newcommand{\ms}[1]{{\color{green} #1}}
 \newcommand{\ZD}[1]{{\color{violet}{\bf ZD: #1}}}
 \newcommand{\zd}[1]{{\color{violet}{#1}}}
 \newcommand{\YH}[1]{{\color{orange}{\bf YH: #1}}}
 \newcommand{\yh}[1]{{\color{orange}{#1}}}
\else
 \newcommand{\PF}[1]{}
 \newcommand{\pf}[1]{ #1 }
 \newcommand{\KY}[1]{}
 \newcommand{\ky}[1]{ #1 }
 \newcommand{\MS}[1]{}
 \newcommand{\ms}[1]{ #1 }
 \newcommand{\ZD}[1]{}
 \newcommand{\zd}[1]{{#1}}
 \newcommand{\YH}[1]{}
 \newcommand{\yh}[1]{{#1}}
\fi

\newcommand{\comment}[1]{}

\newcommand{\bA}{\mathbf{A}}
\newcommand{\bU}{\mathbf{U}}
\newcommand{\bS}{\mathbf{S}}
\newcommand{\bM}{\mathbf{M}}
\newcommand{\bK}{\mathbf{K}}
\newcommand{\bR}{\mathbf{R}}
\newcommand{\btA}{\tilde{\mathbf{A}}}
\newcommand{\bbA}{\bar{\mathbf{A}}}
\newcommand{\bW}{\mathbf{W}}
\newcommand{\bX}{\mathbf{X}}
\newcommand{\bXe}{\tilde{\mathbf{X}}}
\newcommand{\bbX}{\bar{\mathbf{X}}}
\newcommand{\bI}{\mathbf{I}}
\newcommand{\bv}{\mathbf{v}}
\newcommand{\bw}{\mathbf{w}}
\newcommand{\bq}{\mathbf{q}}
\newcommand{\be}{\mathbf{e}}
\newcommand{\bte}{\tilde{\mathbf{e}}}
\newcommand{\bx}{\mathbf{x}}
\newcommand{\bp}{\mathbf{p}}
\newcommand{\bt}{\mathbf{t}}

\def\bSigma{\boldsymbol\Sigma}

\newcommand{\minimize}{\operatornamewithlimits{minimize}}

%% file: tex/abstract.tex

\begin{abstract}

Many classical Computer Vision problems, such as essential matrix computation and pose estimation from 3D to 2D correspondences, can be solved by finding the eigenvector corresponding to the smallest, or zero, eigenvalue of a matrix representing a linear system. Incorporating this in deep learning frameworks would allow us to explicitly encode known notions of geometry, instead of having the network implicitly learn them from data. However,  performing eigendecomposition within a network requires the ability to differentiate this operation. Unfortunately, while theoretically doable, this introduces numerical instability in the optimization process in practice. 
  
In this paper, we introduce an eigendecomposition-free approach to training a deep network whose loss depends on the eigenvector corresponding to a zero eigenvalue of a matrix predicted by the network. We demonstrate on several tasks, including keypoint matching and 3D pose estimation, that our approach is much more robust than explicit differentiation of the eigendecomposition, It has better convergence properties and yields state-of-the-art results on both tasks.

\keywords{End-to-end learning, eigendecomposition, singular value decomposition, geometric vision.}
\end{abstract}


%% file: tex/intro.tex

\section{Introduction}
\label{sec:intro}

In traditional Computer Vision, many tasks can be solved by finding the  singular- or eigen-vector corresponding to the smallest, often zero, singular- or eigen-value of the matrix encoding a linear system. Examples include estimating essential matrices or homographies from matched keypoints and computing pose from 3D to 2D correspondences. 

In the era of Deep Learning, there is growing interest in embedding these methods within a deep architecture to allow end-to-end training. For example, it has recently been shown that such an approach can be used to train networks to detect and match keypoints in image pairs while accounting for the global consistency of the correspondences~\cite{Yi18a}. More generally, this approach would allow us to explicitly encode notions of geometry within deep networks, thus sparing the network the need to re-learn what has been known for decades and making it possible to learn from smaller amounts of training data. 

One way to implement this approach is to design a network whose output defines a matrix and train it so that the smallest singluar- or eigen-vector of the matrices it produces are as close as possible to ground-truth ones. This is the strategy used in~\cite{Yi18a} to simultaneously establish correspondences and compute the corresponding Essential Matrix: The network's outputs are weights discriminating inlier correspondences from outliers and are used to assemble an auxiliary matrix whose smallest eigenvector is the sought-for Essential Matrix. 

The main obstacle to implementing this approach is that it requires being able to differentiate the singular value decomposition (SVD) or the eigendecomposition (ED) in a stable manner to train the network, a non-trivial problem that has already received considerable attention~\cite{Papadopoulo00,Giles08,Ionescu15}. As a result, these decompositions are already part of standard Deep Learning frameworks, such as TensorFlow~\cite{Tensorflow} or PyTorch~\cite{PyTorch}. However,
they
ignore two key practical issues. First, when optimizing with respect to the matrix itself or with respect to parameters defining it, the vector corresponding to the smallest singular value or eigenvalue may switch abruptly as the relative magnitudes of these values change, which is essentially non-differentiable. This is illustrated in the example of Fig.~\ref{fig:denoise}, discussed in detail in Section~\ref{sec:motivation}. Second, computing the gradient requires dividing by the difference between two singular values or eigenvalues, which could be zero.  While a solution to the latter was proposed in~\cite{Papadopoulo00}, the former is unavoidable.

\input{fig/denoise}

In this paper, we therefore introduce an approach to training a deep network whose loss depends on the eigenvector corresponding to a zero eigenvalue of a matrix $\bM$, which is either the output of the network or a function of it, {\it without} explicitly performing an SVD or ED. Our loss is fully differentiable, does {\it not} suffer from the instabilities the above-mentioned problems can cause, and can be naturally incorporated in a deep learning architecture. In practice, because image measurements are never perfect, the eigenvalue is never strictly zero. This, however, does not affect the computation either, which makes our approach robust to noise. 

To demonstrate this in a Deep Learning context, we evaluate our approach on the tasks of training a network to find globally-consistent keypoint correspondences using the essential matrix and training another to remove outliers for pose estimation when solving the Perspective-n-Point (PnP) problem. In both cases, our approach delivers state-of-the-art results, whereas using the standard implementation of singular- and eigen-value decomposition provided in TensorFlow results in either the learning procedure not converging or in significantly worse performance.


%% file: fig/denoise.tex

\begin{figure}[t]
    \centering
    \begin{tabular}{cc}
     \parbox[c]{0.35\linewidth}{\includegraphics[width=1.0\linewidth]{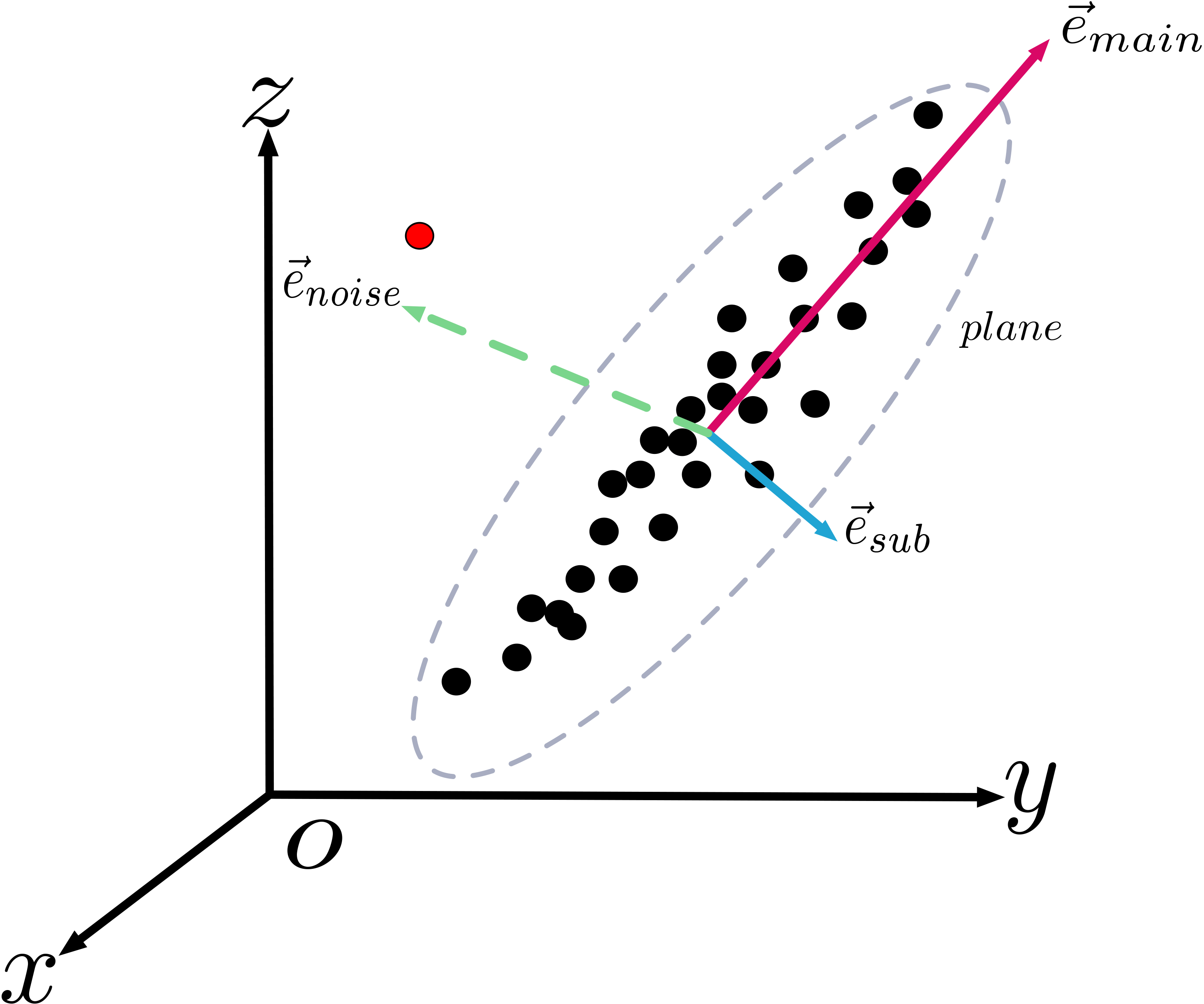}} &
     \parbox[c]{0.3\linewidth}{\includegraphics[width=1.0\linewidth]{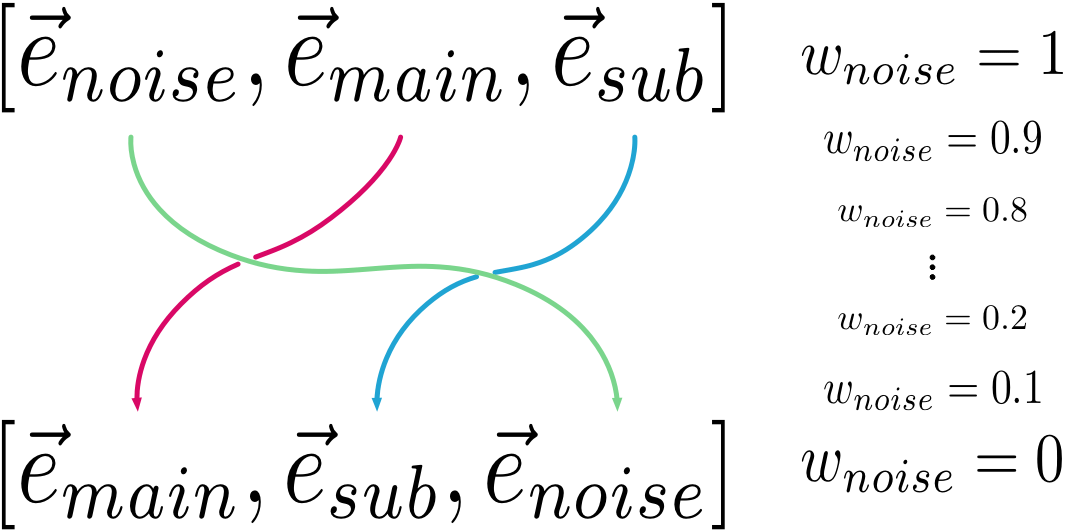}}\\[-2mm]
    (a) & (b)
    \end{tabular}{}
 \caption{{\bf Eigenvector switching.}
   {\bf (a)} 3D points lying on a plane in black and distant outlier in red. {\bf (b)} When the weights assigned to all the points are one, the eigenvector corresponding to the smallest eigenvalue is $\be_{sub}$, the vector shown in blue in (a), and on the right in the top portion of (b), where we sort the eigenvectors by decreasing eigenvalue. As the optimization progresses and the weight assigned to the outlier decreases, the eigenvector corresponding to the smallest eigenvalue switches to $\be_{noise}$, the vector shown in green in (a), which introduces a sharp change in the gradient values. }
\label{fig:denoise}
\vspace{-0.5cm}
\end{figure}


%% file: tex/motivation.tex

\section{Motivation}
\label{sec:motivation}

To illustrate the problems associated with differentiating eigenvectors and eigenvalues, consider the outlier rejection toy example depicted by Fig.~\ref{fig:denoise}. The inputs are 3D points lying on a plane and drawn in black, and an outlier 3D point shown in red, which we assume to be very far from the plane. Suppose we want to assign a binary weight to each point (1 for inliers, 0 for outliers) such that the eigenvector corresponding to the smallest eigenvalue of the weighted covariance matrix is close to the ground-truth one in the least-square sense. When the weight assigned to the outlier is 0, it would be $\be_{noise}$, which is also the normal to the plane and is shown in green.  However, if at some point during optimization, typically at initialization, we assign the weight 1 to the outlier, $\be_{noise}$ will correspond to the largest eigenvalue instead of the smallest, and the eigenvector corresponding to the smallest eigenvalue will be the vector $\be_{sub}$ shown in blue, which is perpendicular to $\be_{noise}$. As a result, if we initially set all weights to 1 and optimize them so that the smallest eigenvector approaches the plane normal, the gradient values will depend on the coordinates of $\be_{sub}$. At one point during the optimization, if everything goes well, the weight assigned to the outlier will become small enough so that the smallest eigenvector switches from being $\be_{sub}$ to being $\be_{noise}$, which introduces a large jump in the gradient vector whose values will now depend on the coordinates of $\be_{noise}$ instead of $\be_{sub}$.

In this simple case, this kind of instability does not preclude eventual convergence. However, in more complex situations, we found that it does, as evidenced by our experiments. This problem was already noted in~\cite{Yi18a} in the context of learning keypoint correspondences. To circumvent this issue, the algorithm in~\cite{Yi18a} had to first rely on a classification loss  to determine the potential inlier correspondences before incorporating the loss based on the essential matrix to impose geometric constraints, which requires eigendecomposition. This ensured that the  network weights were already good enough to prevent eigenvector switching when starting to minimize the geometry-based loss.


%% file: tex/related.tex

\section{Related Work}

In recent years, the need to integrate geometric methods and mathematical tools into Deep Learning frameworks has led to the reformulation of a number of them in network terms. For example,~\cite{Jaderberg15} considers spatial transformations of image regions with CNNs. The set of such transformations is extended in~\cite{Handa16}. In a different context,~\cite{Murray16} derives a differentiation of the Cholesky decomposition that could be integrated in Deep Learning frameworks.  

Unfortunately, the set of geometric Computer Vision problems that these methods can handle remains relatively limited. In particular, there is no widely accepted deep-learning way to solve the many geometric problems that reduce to finding least-square solution of linear systems. In this work, we consider two such problems: Computing the essential matrix from keypoint correspondences in an image pair and estimating the 3D pose of an object from 3D-to-2D correspondences, both of which we briefly discuss below.

\paragraph{\bf Estimating the Essential matrix from correspondences.} 

The eigenvalue-based solution to this problem has been known for decades~\cite{Longuet-Higgins81,Hartley97,Hartley00} and remains the standard way to compute Essential matrices~\cite{Nister03}. The real focus of research in this area has been to establish reliable keypoint correspondences  and to eliminate outliers. In this context, variations of RANSAC~\cite{Fischler81}, such as MLESAC~\cite{Torr00} and Least median of squared (LMeds)~\cite{Rousseeuw87}, and very recently GMS~\cite{Bian17}, have become popular. For a comprehensive study of such methods, we refer the interested reader to~\cite{Raguram13}. With the emergence of Deep Learning, there has been a trend towards moving away from this decades-old knowledge and apply instead a black-box approach where a Deep Network is trained to directly estimate the rotation and translation matrices~\cite{Zamir16,Ummenhofer17} without {\it a priori} geometrical knowledge. The very recent work of~\cite{Yi18a} attempts to reconcile these two opposing trends by embedding the geometric constraints into a Deep Net and has demonstrated superior performance for this task when the correspondences are hard to establish.

\paragraph{\bf Estimating 3D pose from 3D-to-2D correspondences.} This is known as the Perspective-n-Point (PnP) problem. It has also been investigated for decades and is also amenable to an eigendecomposition-based solution~\cite{Hartley00}, many variations of which have been proposed over the years~\cite{Lepetit09,Kneip11,Zheng13,Ferraz14}.  DSAC~\cite{Brachmann16b} is the only approach we know of that integrates the PnP solver into a Deep Network. As explicitly differentiating through the PnP solver is not optimization friendly, the authors apply the log trick used in the reinforcement learning literature. This amounts to using a numerical approximation of the derivative from random samples, which is not ideal, given that an analytical alternative exists. Moreover, DSAC only works for grid configurations and known scenes. By contrast, the method we propose in this work has an analytical form, with no need for stochastic sampling.

\paragraph{\bf Differentiating the eigen- and singular value decomposition}

Whether computing the essential matrix, estimating 3D pose, or solving any other least-squares problem, incorporating an eigendecomposition-solver into a deep network requires differentiating the eigendecomposition. Expressions for such derivatives have been given in~\cite{Papadopoulo00,Giles08} and reformulated in terms that are compatible with back-propagation in~\cite{Ionescu15}. Specifically, as shown in~\cite{Ionescu15}, for a matrix $\bM$ written as $\bM = \bU \bSigma \bU^T$, the variations of the eigenvectors $\bU$ with respect to the matrix, used to compute derivatives, are
\begin{equation}
d\bU = 2\bU\left(\bK \odot (\bU^T d\bM \bU)_{sym}\right)\;,
\end{equation}
where $\bS_{sym} = \frac{1}{2}(\bS^T + \bS)$, and
\begin{equation}
\bK_{ij} = \begin{cases} \frac{1}{\sigma_i - \sigma_j}, & i\neq j \\
      0, & i=j \end{cases}\;.
      \label{eq:K}
\end{equation}
As can be seen from Eq.~\ref{eq:K}, if two eigenvalues are equal, that is, $\sigma_i = \sigma_j$, the denominator becomes 0, thus creating numerical instabilities. The same can be said about singular value decomposition.

A solution to this was proposed in~\cite{Papadopoulo00}, and singular- and eigen-value decomposition have been used within deep networks for problems where all the singular values are used and their order is irrelevant~\cite{Huang17a,Huang17b}. In the context of spectral clustering, the approach of~\cite{Law17} also proposed a solution that eliminates  the need for explicit eigendecomposition. This solution, however, was dedicated to the scenario where one seeks to use all non-zero eigenvalues, assuming a matrix of constant rank. 

Here, by contrast, we tackle problems where what matters is a single eigen- or singular-value. In this case, the order of the eigenvalues is important. However, this order can change during training, which results in a non-differentiable switch from one eigenvector to another,  as in the toy example of Section~\ref{sec:motivation}. In turn, this leads to numerical instabilities, which can prevent convergence. In~\cite{Yi18a}, this problem is finessed by first training the network using a classification loss that does not depend on eigenvectors. Only once a sufficiently good solution is found, that is, a solution close enough to the correct one for vector switching not to happen anymore, is the loss term that depends on the eigenvector associated to the smallest eigenvalue turned on. 
As we will show later, we can achieve state-of-the-art results without the need for such a heuristic, by deriving a more robust, eigendecomposition-free loss function.


%% file: tex/methodology.tex

\section{Our Approach}

We introduce an approach that enables us to work with eigenvectors corresponding to zero eigenvalues within an end-to-end learning formalism, while being subject to neither the gradient instabilities due to vector switching discussed in Section~\ref{sec:motivation} nor to difficulties caused by repeated eigenvalues.
To this end, we derive a loss function that directly operates on the matrix whose eigen- or singular-vectors we are interested in but without explicitly performing an SVD or ED.

In this section, we first discuss the generic scenario in which the matrix of interest directly is the output of the network. We then consider the slightly more involved case where the network predicts weights that themselves define the matrix, which corresponds to our application scenarios. Note that, while we discuss our approach in the context of Deep Learning, it is applicable to \textit{any} optimization framework where one seeks to optimize a loss function based on the smallest eigenvector of a matrix with respect to the parameters that defining this matrix.

\subsection{Generic Scenario}
\label{sec:general}

Given an input measurement $\bx$, let us denote by $f_{\theta}(\bx)$ the output of a deep network with parameters $\theta$. Here, we consider the case where the output of the network is a matrix, which we write as $\bA_\theta = f_{\theta}(\bx)$. Our goal is to tackle problems where the loss function of the network depends on the smallest eigenvector $\be_\theta$ of 

$\bA^{T}_{\theta}\bA_\theta$, which ensures that the matrix is symmetric. 

Typically, one can use an $\ell_2$ loss of the form $\|\be_\theta - \bte\|^2$, where $\bte$ is the ground-truth smallest eigenvector. The standard approach to addressing this, as followed in~\cite{Ionescu15,Yi18a}, consists of explicitly differentiating this loss w.r.t. $\be_\theta$, then $\be_\theta$ w.r.t. $\bA_\theta$ and finally $\bA_\theta$ w.r.t. $\theta$ via backpropagation. As discussed above, however, this is not optimization friendly.

To overcome this, we propose to define a new loss motivated by the linear equation that defines eigenvectors and eigenvalues. Specifically, if $\be_\theta$ is an eigenvector of $\bA^{T}_\theta\bA_\theta$ with eigenvalue $\lambda$, 

it satisfies
\begin{equation}
\bA^{T}_\theta\bA_\theta \be_\theta = \lambda \be_\theta\;.
\end{equation}
Since eigenvectors have unit-norm, i.e., $\be_\theta^T\be_\theta = 1$, multiplying both sides of this equation from the left by $\be_\theta^T$ yields
\begin{equation}
\be_\theta^T\bA^{T}_\theta\bA_\theta \be_\theta = \lambda\;.
\end{equation}

In this paper, we consider zero eigenvalue problems, that is, $\lambda=0$. Since $\bA^{T}_\theta\bA_\theta$ is positive semi-definite, we have that $\be^T\bA^{T}_\theta\bA_\theta \be \geq 0$ for any $\be$. Given the ground-truth eigenvector $\bte$ that we seek to predict, this lets us define the loss function
\begin{equation}
L_{eig}(\theta) = \bte^T \bA^{T}_\theta\bA_\theta \bte\;.
\end{equation}
Intuitively, this loss aims to find the parameters $\theta$ such that $\bte$ is an eigenvector of the resulting matrix $\bA^{T}_\theta \bA_\theta$ with minimum eigenvalue, that is, zero in our case, assuming that we can truly reach the global minimum of our loss. However, this loss alone has multiple, globally-optimal solutions, including the trivial one $\bA_\theta = 0$.

To address this, we note that this trivial solution has not only one zero eigenvalue corresponding to eigenvector $\bte$, but that all its eigenvalues are zero. Since, in practice, we typically search for matrices that have a single zero eigenvalue, we propose to maximize the projection of the data along the directions orthogonal to $\bte$. Such a projection can be achieved by making use of the orthogonal complement to $\bte$, given by $(\bI - \bte\bte^T)$, where $\bI$ is the identity matrix. By defining $\bbA_\theta=\bA_\theta(\bI - \bte\bte^{T})$, we can then re-write our loss function as
\begin{equation}
\tilde{L}(\theta) = \bte^T \bA^{T}_\theta \bA_\theta \bte -  \alpha\text{tr}\left(\bbA^{T}_{\theta} \bbA_\theta\right)\;,
\end{equation}
where $\text{tr}(\cdot)$ computes the trace of a matrix and $\alpha$ sets the relative influence of the two terms. Note that we can apply the same strategy to cases where multiple eigenvalues are zero, by reducing the orthogonal space to only the directions corresponding to non-zero eigenvalues, and introducing the first term for all eigenvectors whose eigenvalues we want to be zero.

For numerical stability, we further propose to bound the second term in the range $[0,1]$. To do so, we therefore re-write our loss as
\begin{equation}
L(\theta) = \bte^T \bA^{T}_\theta \bA_\theta \bte + \alpha\exp\left(-\beta\text{tr}\left(\bbA^{T}_{\theta} \bbA_\theta\right)\right)\;,
\label{eq:general}
\end{equation}
where $\beta$ is a scalar. This loss is fully differentiable, and can thus be used to learn the parameters $\theta$ of a deep network. Since it does not explicitly depend on performing an eigendecomposition at every iteration of the optimization, it suffers from neither the eigenvector switching problem, nor the non-unique eigenvalue problem.

\subsection{Learning to Predict Weights}
\label{sec:method}

In practice, the problem of interest is often more constrained than training a network to directly output a matrix $\bA_\theta$. In particular, in this paper, we consider problems where the goal is to predict a weight $w_i$ for each element of the input. This typically leads to formulations where $\bA_\theta^T\bA_\theta$ has the form $\bX^T\bW\bX$, with $\bX$ a data matrix and $\bW$ a diagonal matrix whose elements are the $w_i$s. Below, we introduce the formulation for each of the applications that we consider in our experiments.

\subsubsection{Outlier Rejection with 3D Points.}

To show that we can indeed back-propagate nicely through the proposed loss formulation where directly using the analytical gradient fails, we first briefly revisit the toy outlier rejection problem used to motivate our approach in Section~\ref{sec:intro}. For this experiment, we do not train a Deep Network, or perform any learning procedure. Instead, given $N$ 3D points $\bx_i$, including inliers and outliers, we directly optimize the weight $w_i$ of each point. At every step of optimization, given the current weight values, we compute the weighted mean of the points $\mu = \frac{1}{\sum_{i = 1}^{N}w_{i}}\sum_{i=1}^{N}w_{i}\bx_{i}$. Let $\bX$ be the $3\times N$ matrix of mean-subtracted 3D points. We then compute the weighted covariance matrix $\mathbf{C} = \bX^T \bW \bX$, where $\bW$ is a diagonal matrix whose elements are the $w_i$s. The smallest eigenvector of $\mathbf{C}$ then defines the direction of noise.

Given the ground-truth such eigenvector $\bte$, let $\bbX=\bI - \bte\bte^{T}$. We adapt the general formulation of Eq.~\ref{eq:general} and formulate the outlier rejection problem as
\begin{equation}
\minimize_{\bw}~~ \bte^T \bX^T \bW \bX \bte\ +
\alpha \exp\left(-\beta tr(\bbX^{T}\bW\bbX)\right)
\;.
\label{eq:loss_denoise}
\end{equation}
Note that this translates directly to Eq.~\ref{eq:general} by defining $\bA_\theta=\bW^{\frac{1}{2}}\bbX$, where $\bW^{\frac{1}{2}}$ is a diagonal matrix with elements $\sqrt{w_i}$.

\subsubsection{Keypoint Matching with the Essential Matrix.}

For this task, to isolate the effect of the loss function only, we followed the same setup as in~\cite{Yi18a}. Specifically, we used the same network architecture as in~\cite{Yi18a}, which takes $C$ correspondences between two 2D points as input and outputs a $C$-dimensional vector of weights, that is, one weight for each correspondence. 

Formally, let
\begin{equation}
\mathbf{q}_{i} = \left[ u_{i}, v_{i}, u'_{i}, v'_{i}\right]^T\;,
\end{equation}
encode the coordinates of correspondence $i$ in the two images. Following the 8 points algorithm~\cite{Longuet-Higgins81}, we construct as matrix $\bX \in \mathbb{R}^{C\times9}$, each row of which is computed from one correspondence vector $\bq_i$ as
\begin{equation}
\bX^{(i)} = [u_iu_i', u_iv_i, u_i, v_iu_i', v_iv_i', v_i, u_i', v_i', 1]\;,
\end{equation}
where $\bX^{(i)}$ denotes row $i$ of $\bX$. A weighted version of the 8 points algorithm~\cite{Zhang98} then computes the essential matrix as the smallest eigenvector of $\bX^T\bW \bX$, with $\bW$ the diagonal matrix of weights.

Let $\bbX = \bX(\bI - \bte\bte^{T})$, where $\bte$ is the ground-truth eigenvector representing the true essential matrix. We can then write an eigendecomposition-free essential loss as
\begin{equation}
L(\bW) = \bte^T \bX^{T}\bW \bX \bte + \alpha\exp\left(-\beta\text{tr}\left(\bbX^{T} \bW \bbX\right)\right)\;.
\label{eq:loss_weights}
\end{equation}
Given a set of training samples, consisting of $N$ image pairs with ground-truth essential matrices, we can then use this loss, instead of the classification loss or essential loss of~\cite{Yi18a}, to train a network to predict the weights.

Note that, as suggested by~\cite{Hartley00} and done in~\cite{Yi18a}, we use the 2D coordinates normalized to $[-1,1]$ using the camera intrinsics as input to the network.

When calculating the loss, as suggested by~\cite{Hartley97}, we move the centroid of the reference points to the origin of the coordinate system and scale the points so that their RMS distance to the origin is equal to $\sqrt{2}$. This means that we also have to scale and translate $\bte$ accordingly.

\subsubsection{3D-to-2D Correspondences for Pose Estimation.}

The goal of this problem, also known as the Perspective-n-Point (PnP) problem~\cite{Lepetit09}, is to determine the absolute pose (rotation and translation) of a calibrated camera, given known 3D points and corresponding 2D image points.

For this task, as we are still dealing with sparse correspondences, we use the same network architecture as before for 2D-to-2D correspondences, except that we now have one additional input dimension, since we have {\it 3D}-to-2D correspondences. 

This network takes $C$ correspondences between 3D and 2D points as input and outputs a $C$-dimensional vector of weights, still one weight for each correspondence. 

Mathematically, we can denote the input correspondences as
\begin{equation}
\mathbf{q}_{i} = [x_{i}, y_{i}, z_{i}, u_{i}, v_{i}]^{T}\;,
\end{equation}
where $x_{i}, y_{i}, z_{i}$ are the coordinates of a 3D point, and $u_i$, $v_i$ denote the corresponding image location.
According to~\cite{Hartley00}, we have
\begin{equation}
f_{scale}\begin{bmatrix}
u_{i}\\
v_{i}\\
1
\end{bmatrix} = 
\begin{bmatrix}
\mathbf{R},\mathbf{t}
\end{bmatrix}
\begin{bmatrix}
x_{i}\\
y_{i}\\
z_{i}\\
1
\end{bmatrix}=
\begin{bmatrix}
 p_{1} &p_{2} &p_{3} &p_{4} \\ 
 p_{5} &p_{6} &p_{7} &p_{8} \\ 
 p_{9} &p_{10} &p_{11} &p_{12} \\
\end{bmatrix}
\begin{bmatrix}
x_{i}\\
y_{i}\\
z_{i}\\
1
\end{bmatrix}\;.
\end{equation}
To recover the pose, we then follow the Direct Linear Transform (DLT) method~\cite{Hartley00}. This consists of constructing the matrix $\bX \in \mathbb{R}^{2C\times12}$, every two rows of which are computed from one correspondence $\bq_{i}$ as
\setcounter{MaxMatrixCols}{12}
\begin{equation}
\begin{bmatrix}
\mathbf X^{(2i-1)}\\
\mathbf X^{(2i)}
\end{bmatrix} = 
\begin{bmatrix}
 x_{i}& y_{i}& z_{i}& 1& 0& 0& 0& 0& -u_{i}x_{i}& -u_{i}y_{i}& -u_{i}z_{i}& -u_{i} \\
 0& 0& 0& 0& x_{i}& y_{i}& z_{i}& 1& -v_{i}x_{i}& -v_{i}y_{i}& -v_{i}z_{i}& -v_{i}
\end{bmatrix}
\;,
\end{equation}
where $\bX^{(i)}$ denotes row $i$ of $\bX$. Then, the solution of the weighted PnP problem can be obtained as the eigenvector of $\bX^{T}\bW\bX$ corresponding to the smallest eigenvalue. Therefore, we can define a PnP loss similar to the one of Eq.~\ref{eq:loss_weights} for 2D-to-2D correspondences, but with $\bX$ defined as discussed above, and, given $N$ training samples, each consisting of a set of 3D-to-2D correspondences with corresponding ground-truth eigenvector encoding the pose, train a network to predict weights such that we obtain the correct pose via DLT. As in the 2D-to-2D case, we use the normalized coordinate system for the 2D coordinates.

Note that the characteristics of the rotation matrix, that is, orthogonality and determinant 1, are not preserved by the DLT solution. Therefore, to make the result a valid rotation matrix, we refine the DLT results by the generalized Procrustes algorithm~\cite{Garro12,Schonemann66}, which is a common post-processing technique for PnP algorithms. Note that this step is not involved during training, but only in the validation process to select the best model and at test time.


%% file: tex/experiments.tex
\section{Experiments}

We now present our results for the three tasks discussed above, that is, plane fitting as in Section~\ref{sec:motivation}, distinguishing good keypoint correspondences from bad ones, and solving the Perspective-n-Point (PnP) problem. We rely on a  TensorFlow implementation using the Adam~\cite{Kingma15} optimizer, with a learning rate of $10^{-4}$, unless stated otherwise, and default parameters. When training a network for keypoint matching and PnP, we used mini-batches of 32 samples and, in the plane fitting case, we also tested vanilla gradient descent in addition to Adam. 
 
\subsection{Plane Fitting}

The setup is the one discussed in Section~\ref{sec:motivation}. We randomly sampled 100 3D points on the $z=1$ plane. Specifically, we uniformly sampled $x\in[0,40]$ and $y\in[0,2]$. We then added zero-mean Gaussian noise  with standard deviation $0.001$ in the $z$ dimension. We also generate outliers in a similar way, where $x$ and $y$ is uniformly samples in the same range, and $z$ is sampled from a Gaussian distribution with mean 50 and standard deviation of 5. For the baselines that directly use the analytical gradients of SVD and ED, we take the objective function to be $\min\left\|\be_{min}(\bw) \pm \be_{gt}\right\|_{2}$, where $\be_{min}(\bw)$ is the minimum eigenvector of $\bX^\top\bW\bX$ in Eq.~\ref{eq:loss_denoise} and $\be_{gt}$ is the ground-truth noise direction, which is also the plane normal and is the vector $[0,0,1]$ in this case. Note that we consider both $+\be_{gt}$ and $-\be_{gt}$ and take the minimum distance, denoted by the $\pm$ and the $\min$ in the loss function. For this problem, both solutions are correct due to the sign ambiguity of eigendecomposition, and we need to take this into account. 

We consider two ways of computing analytical gradients, one using the SVD and the other the self-adjoint eigendecomposition (Eigh), which both yield mathematically valid solutions. To implement our approach, we rely on Eq.~\ref{eq:loss_denoise}.

\input{fig/denoise_loss}
\input{fig/in_svd_eig}
Fig.~\ref{fig:denoise_loss} shows the evolution of the loss as the optimization proceeds when using either Adam or vanilla gradient descent, when a single outlier is present. Note that SVD and Eigh have  exactly the same behavior because they constitute two equivalent ways of solving the same problem. Using Adam in conjunction with either one  initially yields a very slow decrease in the loss function, until it suddenly drops to zero when the switch of the eigenvector with the smallest eigenvalue occurs. By contrast, our approach produces a much more gradual decrease in the loss with no overly large gradients ever being generated. The difference in behavior is even more drastic when vanilla gradient descent is used instead of Adam: SVD and Eigh take millions of iterations to converge. 
We tried multiple learning rates within the range $[10^{-5}, 1]$, none of them has led to faster convergence. We provide the results with different learning rates in the supplementary material. 

We also evaluate the behavior of our method and the baselines in the presence of more outliers. As shown in Fig.~\ref{fig:plane_out}, while both our method and the baseline still present the same convergence patterns as before, our approach correctly recovers the inliers and outliers, while the SVD baseline discards many outliers and even accepts outliers.

Note that, while in this plane-fitting example the SVD- or Eigh-based methods converge, in the more complex cases below, this is not always true. 

\subsection{Keypoint Matching}

To evaluate our method on a real-world problem, we use the SUN3D dataset~\cite{Xiao13}. For a fair comparison, we trained our network on the same data as~\cite{Yi18a}, that is, the ``brown-bm-3-05'' sequence, and evaluate it on the test sequences used for testing in~\cite{Ummenhofer17,Yi18a} . Additionally, to show that our method is not overfitting, we also test on a completely different dataset, the ``fountain-P11'' and ``Herz-Jesus-P8'' sequences of~\cite{Strecha08b}.

We follow the evaluation protocol of~\cite{Yi18a}, which constitutes the state-of-the-art in keypoint matching, and only change the loss function to our own loss of Eq.~\ref{eq:loss_weights}. We use $\alpha=10$ and $\beta = 10^{-3}$, which we empirically found to work well for 2D-to-2D keypoint matching. We compare our method against that of~\cite{Yi18a}, both in its original implementation that involves minimizing a classification loss first and  then without that initial step, which we denote as ``Essential\_Only''. The latter is designed to show how critical the initial classification-based minimization of~\cite{Yi18a} is.  In addition, we also compare against 
standard RANSAC~\cite{Cantzler}, LMeds~\cite{Simpson97}, MLESAC~\cite{Torr00}, and GMS~\cite{Bian17} to provide additional reference points.  We do this in terms of the performance metric used in~\cite{Yi18a} and referred to as mean Average Precision (mAP). This metric is computed by observing the ratio of accurately recovered poses given a certain maximum threshold, and taking the area under the curve of this graph.

\input{fig/general}
\input{fig/fountain_brown}

We summarize the results in Fig.~\ref{fig:general} and provide numbers for individual datasets in the supplementary material. Our approach performs roughly on par with~\cite{Yi18a}, the state-of-the-art method on keypoint matching, and outperforms all the other baselines. Importantly, ``Essential\_Only'' severely underperforms and even often fails completely. In short, instead of having to find a workaround to the eigenvector switching problem as in~\cite{Yi18a}, we can directly optimize our objective function, which is far more generally applicable.  Furthermore, the workaround in~\cite{Yi18a} would converge to a sub-optimal solution, as it the classification loss depends on a user-selected decision boundary, that is, a heuristic definition of inliers. By contrast, our method can simply discover the inliers automatically while training, thanks to the second term in Eq.~\ref{eq:general}.

In Fig.~\ref{fig:fountain_brown}, we compare the correspondences classified as inlier by our method to those of RANSAC on image pairs from the dataset of~\cite{Strecha08b} and SUN3D, respectively. Note that even the correspondences that are misclassified as inliers are very close to being inliers. By contrast, RANSAC yields much larger errors.

\subsection{PnP}

Following standard practice for evaluating PnP algorithms~\cite{Lepetit09,Ferraz14}, we used the procedure of~\cite{Ferraz14} to generate a synthetic dataset composed of 3D-to-2D correspondences with noise and outliers added. Each training example comprises two thousand 3D points and we set the ground truth translation of the camera pose $\bt_{gt}$ to be their centroid.

We then create a random ground-truth rotation $\bR_{gt}$, and project the 3D points to the image plane of our virtual camera. As in REPPnP~\cite{Ferraz14}, we apply Gaussian noise with a standard deviation of $5$ to these projections. For outliers, we include random outliers by assigning 3D points to arbitrary valid 2D image positions. 

We train a neural network with the same architecture as in the keypoint matching case, except that it now takes 3D-to-2D correspondences as input. We empirically found that $\alpha=1$ and $\beta = 5\times10^{-3}$ works well for this task. During training, to learn to be robust to outliers, we randomly select between 100 and 1000 of the two thousand matches and turn them into outliers. In other words, the two thousand training matches will contain a random number of outliers that our network will learn to filter out.

We compare our method against modern PnP methods, EPnP~\cite{Lepetit09}, OPnP~\cite{Zheng13}, PPnP~\cite{Garro12}, RPnP~\cite{Li12c} and REPPnP~\cite{Ferraz14}. We also evaluate the DLT~\cite{Hartley00}, since our loss formulation is based on it. Among these methods, REPPnP is the one most specifically designed to handle outliers.
We also report the performance of two commonly used baselines that leverage RANSAC~\cite{Fischler81}, P3P~\cite{Kneip11}+RANSAC and EPnP+RANSAC. For other methods, RANSAC did not bring noticeable improvements, and we omitted them in the graph for better visual clarity.

To compare all these methods, we use standard rotation and translation error metrics~\cite{Crivellaro17}. Specifically, we report the closest arc distance in radians for the rotation matrix measured using quaternions, and the distance between the translation vectors normalized by the ground truth. To demonstrate the effect of outliers at test time, we fix the number of matches to be 200 and vary the number of outliers from $10$ to $150$. We perform each experiment 100 times and report the average. 

\input{fig/pnp}

\input{fig/eigen_loss}

Fig.~\ref{fig:pnp} summarizes the results. We outperform all other methods significantly, especially when the number of outliers increases. REPPnP is the one competing method that seems least affected. As long as the number of outliers is small, it is on a par with us but passed a certain point----when there are more than 40 outliers, that is, 20\% of the total---its performance, particularly in terms of rotation error, decreases quickly whereas ours does not. 

As in the keypoint matching case, we have tried to compute the results of a network relying explicitly on eigendecomposition and minimizing the $\ell_2$ norm of the difference between the ground-truth eigenvector and the predicted one. However, we found that such a network was unable to converge,

as depicted in Fig.~\ref{fig:eigen_loss}, where we compare the loss evolution of this approach to that of ours. This again clearly shows the benefits of our eigendecomposition-free approach.  


%% file: fig/denoise_loss.tex
\begin{figure}[t]
\centering
  \begin{subfigure}{.40\textwidth}
    \centering
    \includegraphics[width=1.\linewidth]{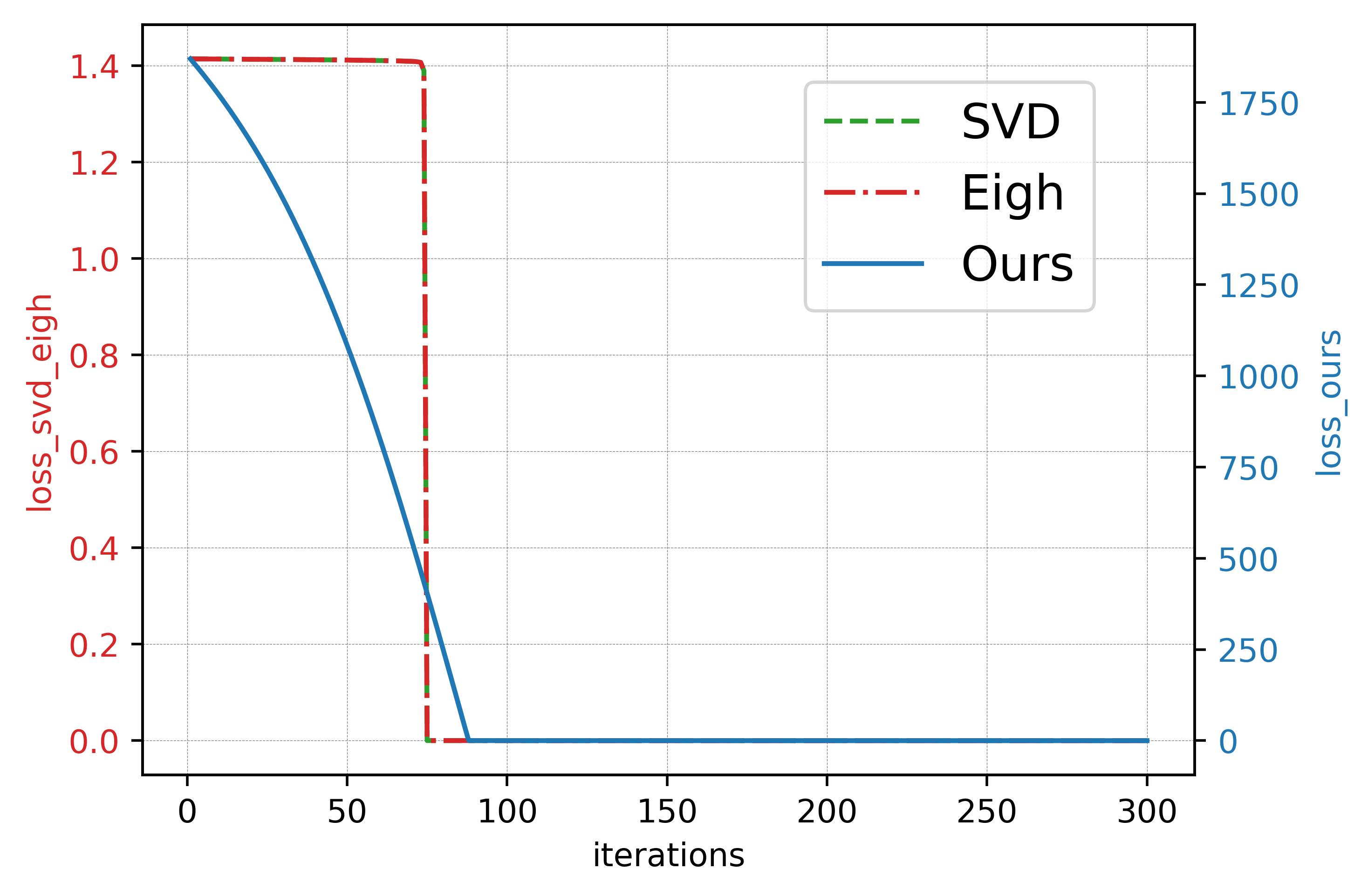}
    \caption{Adam}
    \label{fig:adam}
  \end{subfigure}
  \hspace{2em}
  \begin{subfigure}{.40\textwidth}
    \centering
    \includegraphics[width=1.\linewidth]{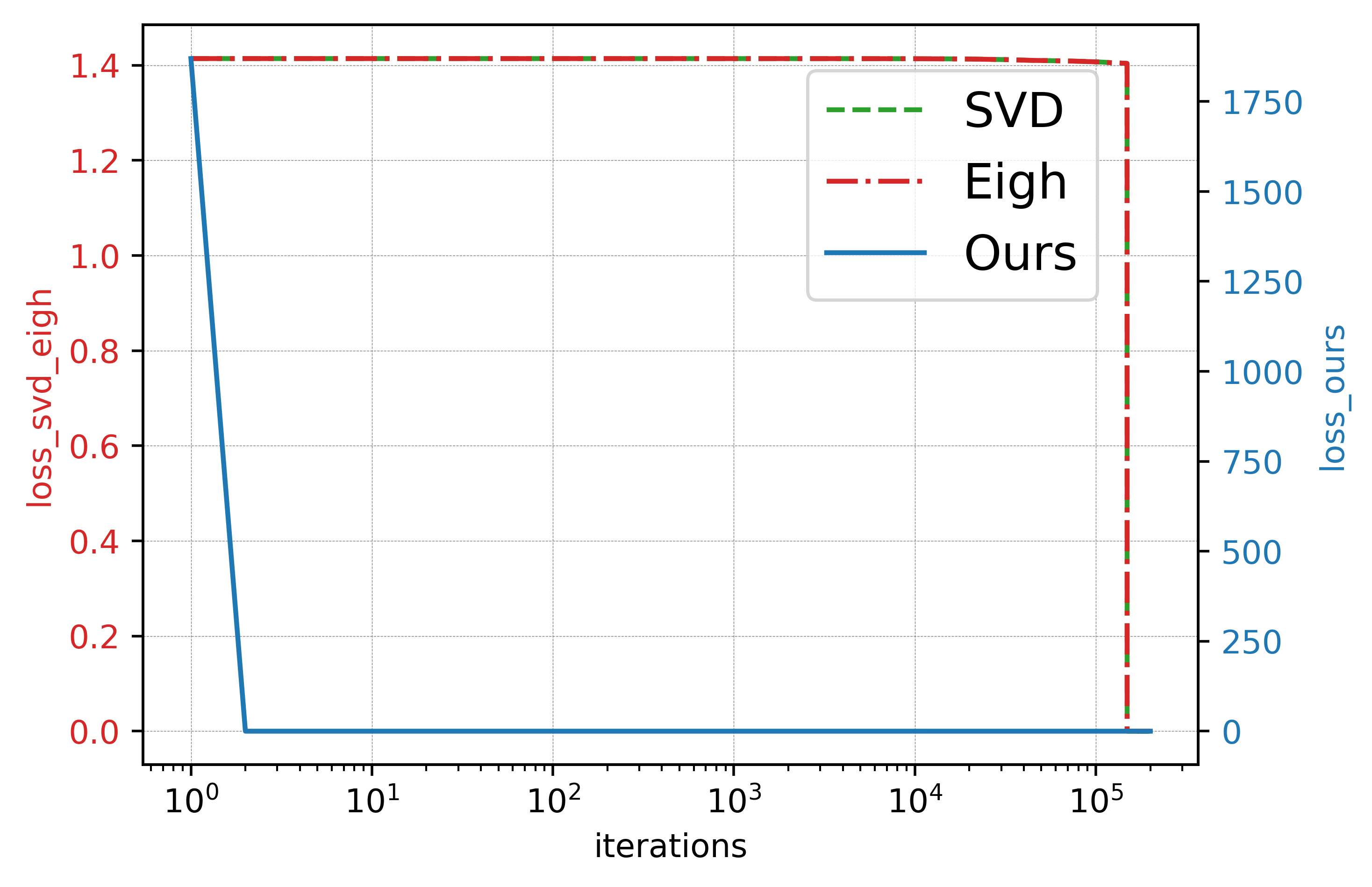}
    \caption{Gradient descent}
    \label{fig:gd}
  \end{subfigure}
  \caption{{\bf Loss evolution graph of the simple toy example.} {\bf (a)} When the Adam optimizer is used, and {\bf (b)} when vanilla gradient descent (GD) is applied. We report results for Singular Value Decomposition (SVD), self-adjoint Eigendecomposition(Eigh), and for our loss function. For each loss, we tried multiple learning rates within the range $[10^{-5},1]$ and report the best results in terms of convergence. In both cases, our loss formulation converges nicely, whereas SVD and Eigh do not. For SVD and Eigh, they do not optimize well until they reach a point where eigenvector swap happens, where only then they start to converge. This happens in an extreme case when GD is used in (b), where it takes millions of iterations to converge even for this very simple example.
    }
  \label{fig:denoise_loss}
  \vspace{-1.5em}
\end{figure}


%% file: fig/in_svd_eig.tex
\begin{figure}[t]
	\centering
  \begin{subfigure}{.32\textwidth}
    \centering
    \includegraphics[width=1.\linewidth]{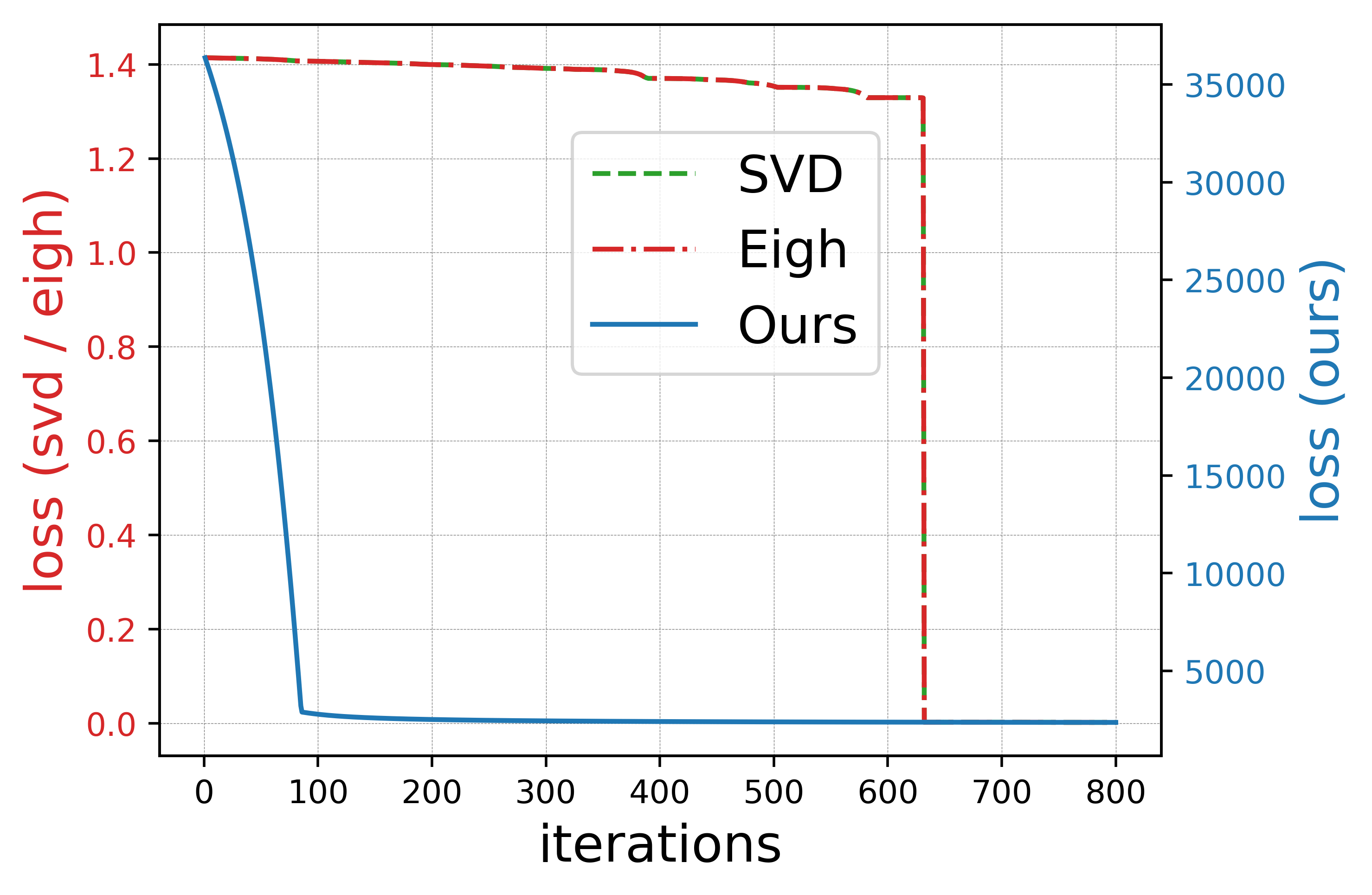}
    \caption{Loss evolution with SVD in Hard case.}
    \label{fig:hard}
  \end{subfigure}
  \begin{subfigure}{.32\textwidth}
    \centering
    \includegraphics[width=1.\linewidth]{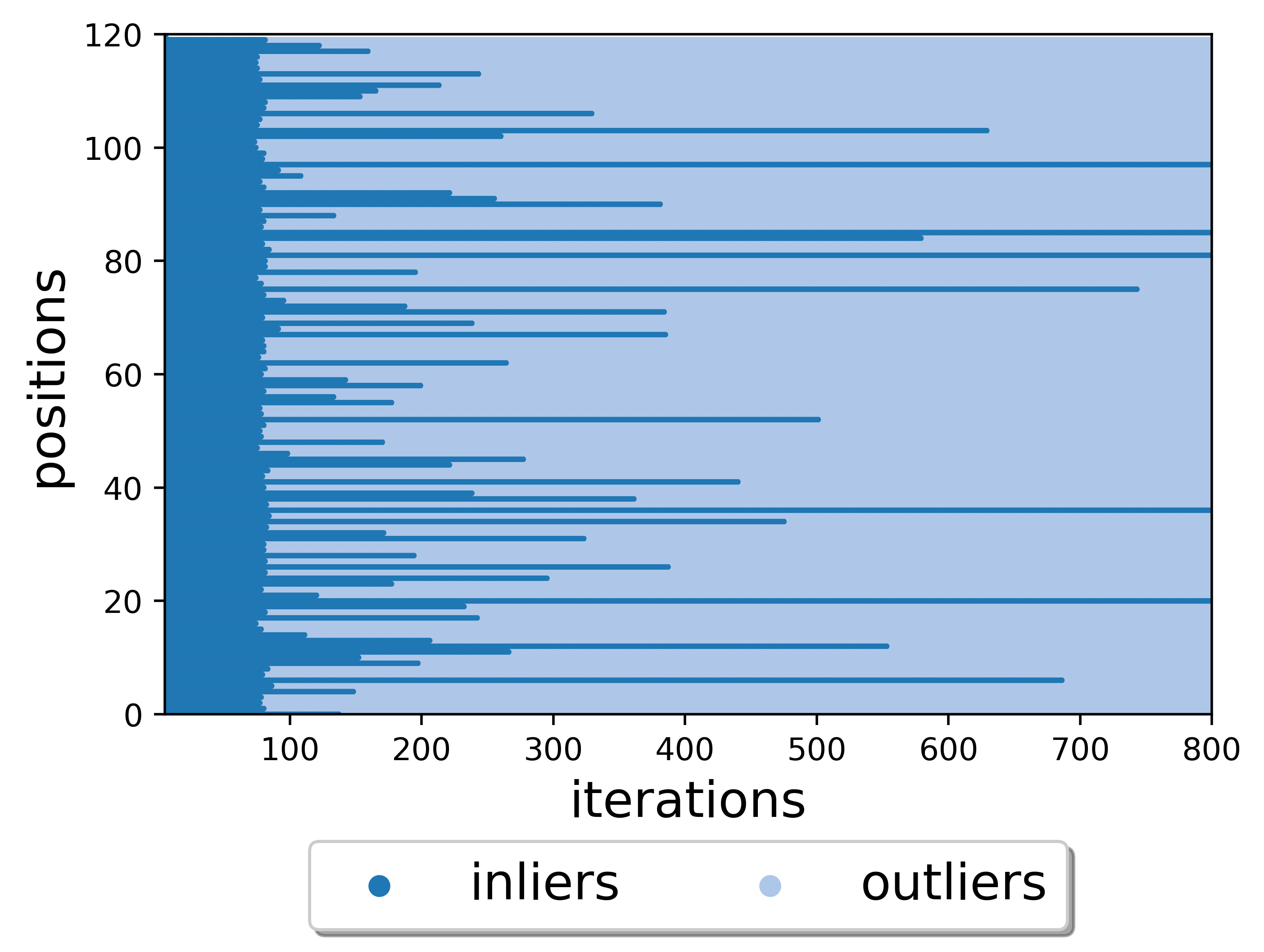}
    \caption{Inliers with SVD}
    \label{fig:in_our}
  \end{subfigure}
  \begin{subfigure}{.32\textwidth}
    \centering
    \includegraphics[width=1.\linewidth]{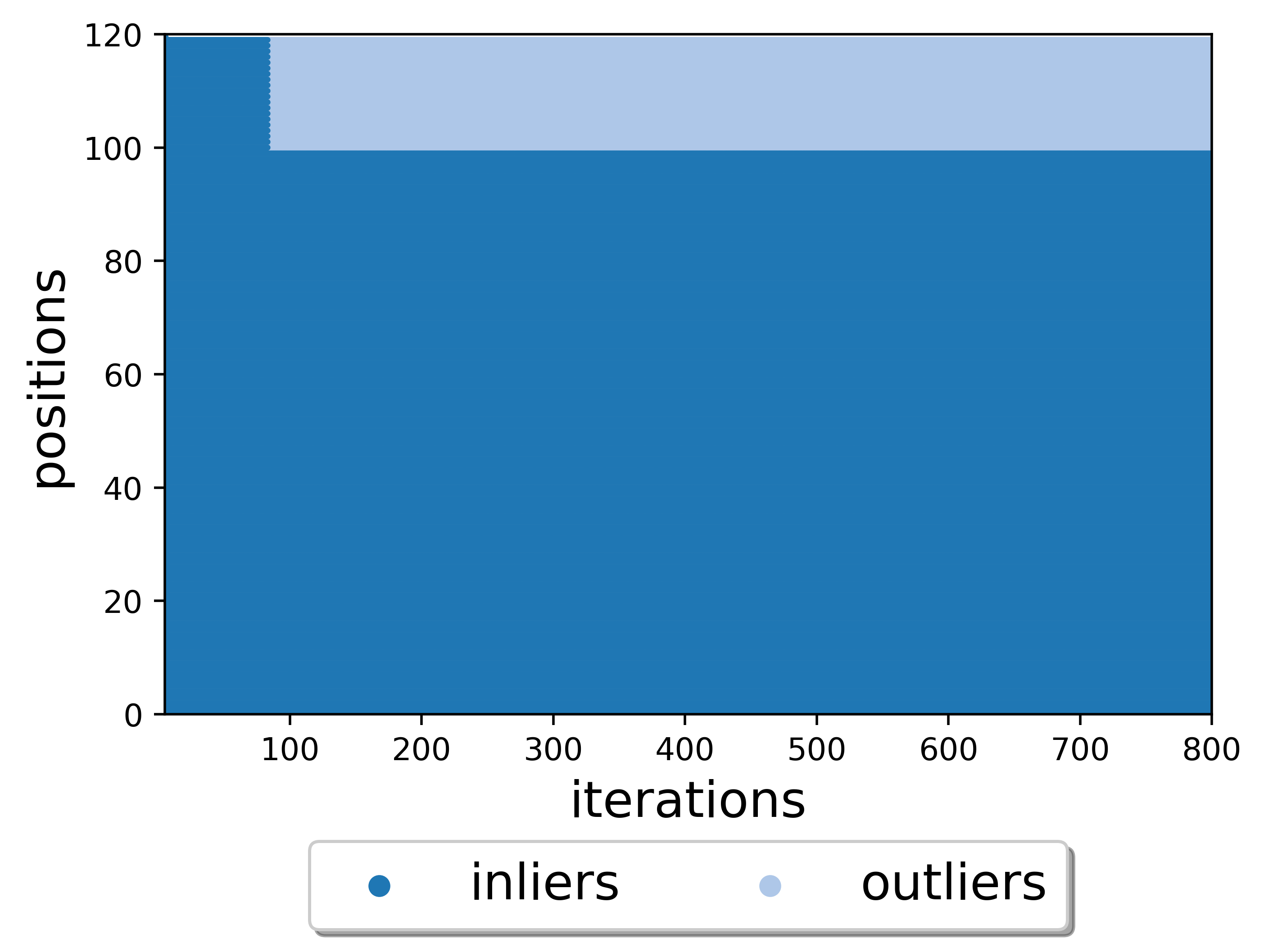}
    \caption{Inliers with Ours}
    \label{fig:in_eig}
  \end{subfigure}
  \caption{{\bf Plane fitting in the presence of multiple outliers.} With multiple outliers, both our approach and the SVD/Eigh baselines still converge (a). However, as illustrated in (b) where we plot the weight of each input point during optimization, the SVD baseline discards many inliers (Positions 1 to 100 are true inliers), while accepting outliers. By contrast, as shown in (c), our approach correctly rejects the outliers and accepts the inliers.}
  \label{fig:plane_out}
  \vspace{-1em}
\end{figure}

%% file: fig/general.tex
\begin{figure}[t]
\begin{subfigure}{0.5\textwidth}
    \centering
    \includegraphics[width=1.\linewidth]{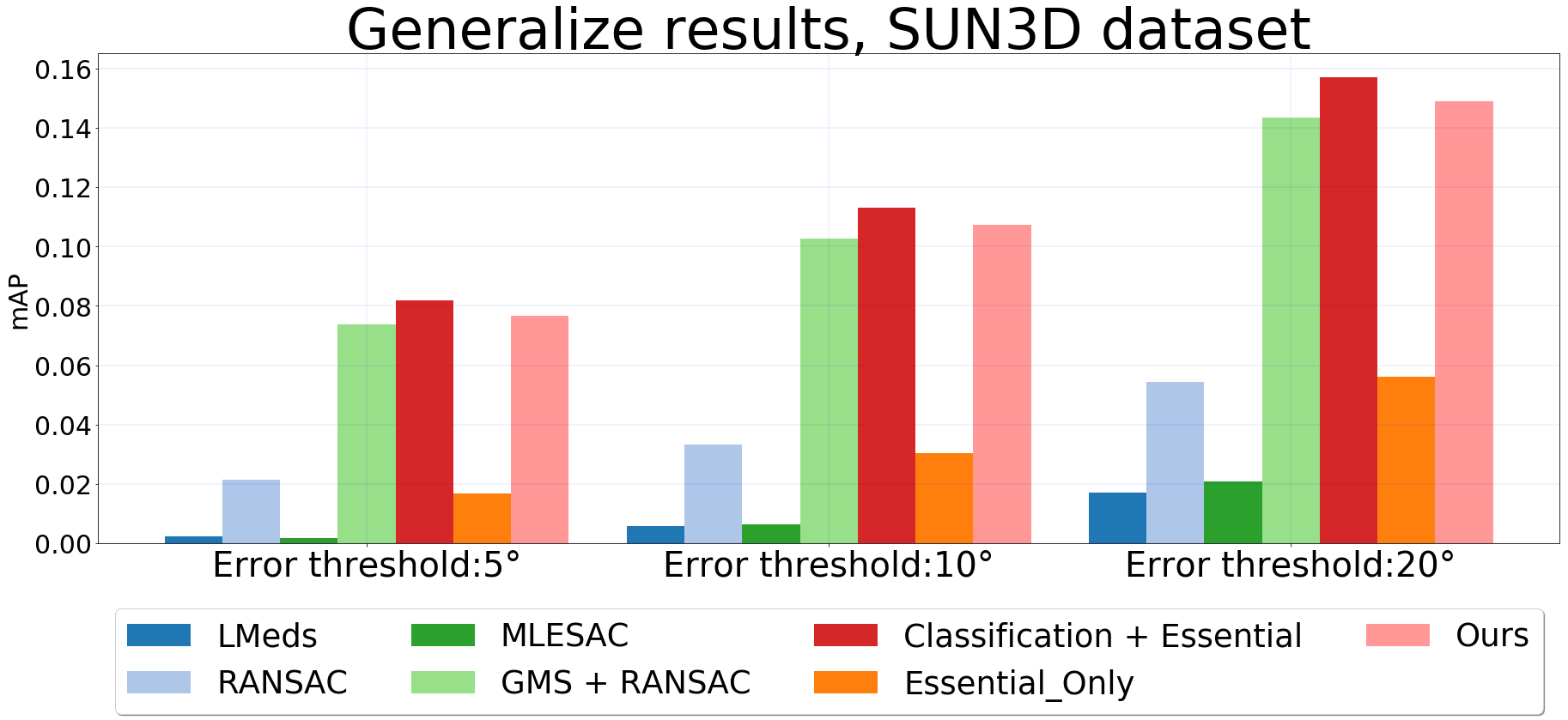}
    \caption{Results on the SUN3D dataset.}
    \label{fig:arc_sun}
\end{subfigure}
\begin{subfigure}{0.5\textwidth}
    \centering
    \includegraphics[width=1.\linewidth]{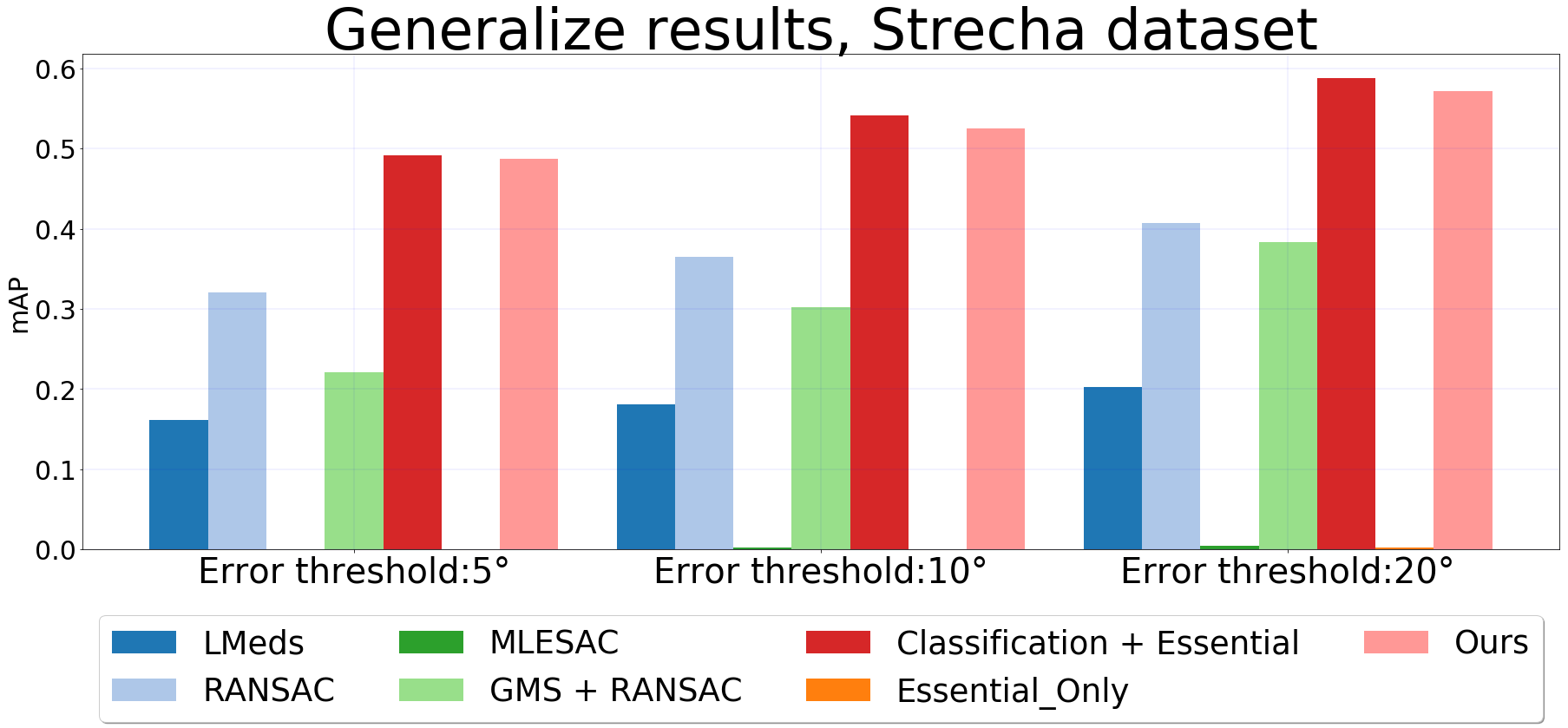}
    \caption{Results on the dataset of~\cite{Strecha08b}}
    \label{fig:arc_strecha}
\end{subfigure}
\caption{{\bf Keypoint matching results.} We report the accuracy of the estimated
  relative pose in terms of the mean Average Precision (mAP) measure of~\cite{Yi18a}. 
  (a) Results for the SUN3D dataset. (b) Results for the dataset of \cite{Strecha08b}. Our method performs on par with the
  state-of-the-art method of~\cite{Yi18a}, denoted as ``Classification +
  Essential'', without the need of any pre-training. Note the significant
  performance gap between ``Essential\_Only'', which utilizes
  eigendecomposition directly, and our method which is
  eigendecomposition-free.}
\label{fig:general}
  \vspace{-1em}
\end{figure}

%% file: fig/fountain_brown.tex
\begin{figure}[t]
  \begin{subfigure}{.25\textwidth}
    \centering
    \includegraphics[width=1.\linewidth]{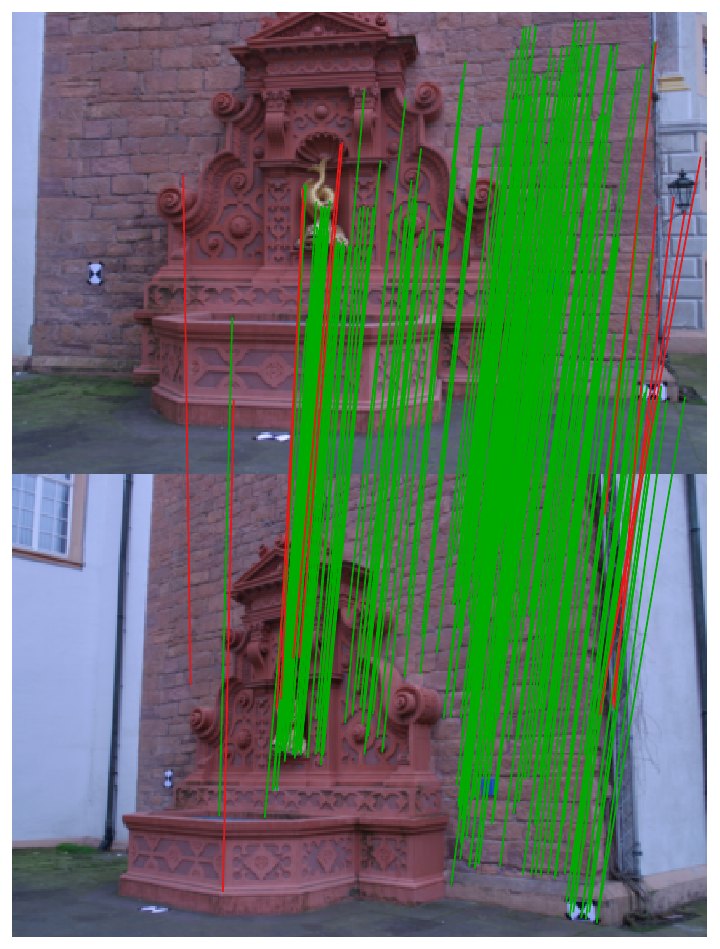}
    \caption{Ours}
    \label{fig:f_ours}
  \end{subfigure}
  \begin{subfigure}{.25\textwidth}
    \centering
    \includegraphics[width=1.\linewidth]{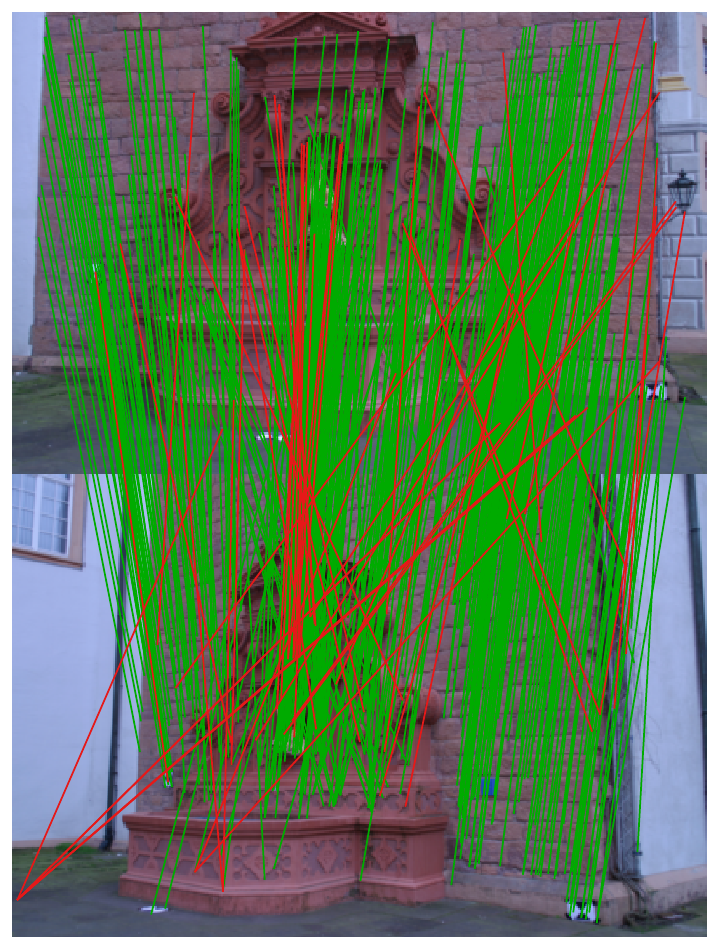}
    \caption{RANSAC}
    \label{fig:f_ransac}
  \end{subfigure}
  \begin{subfigure}{.223\textwidth}
    \centering
    \includegraphics[width=1.\linewidth]{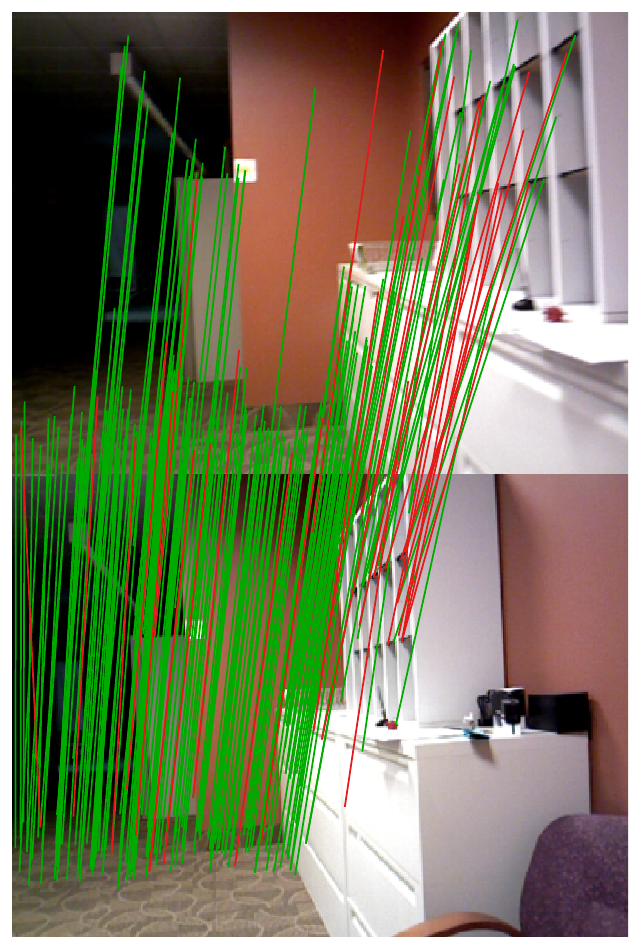}
    \caption{Ours}
    \label{fig:b_ransac}
  \end{subfigure}
  \begin{subfigure}{.223\textwidth}
    \centering
    \includegraphics[width=1.\linewidth]{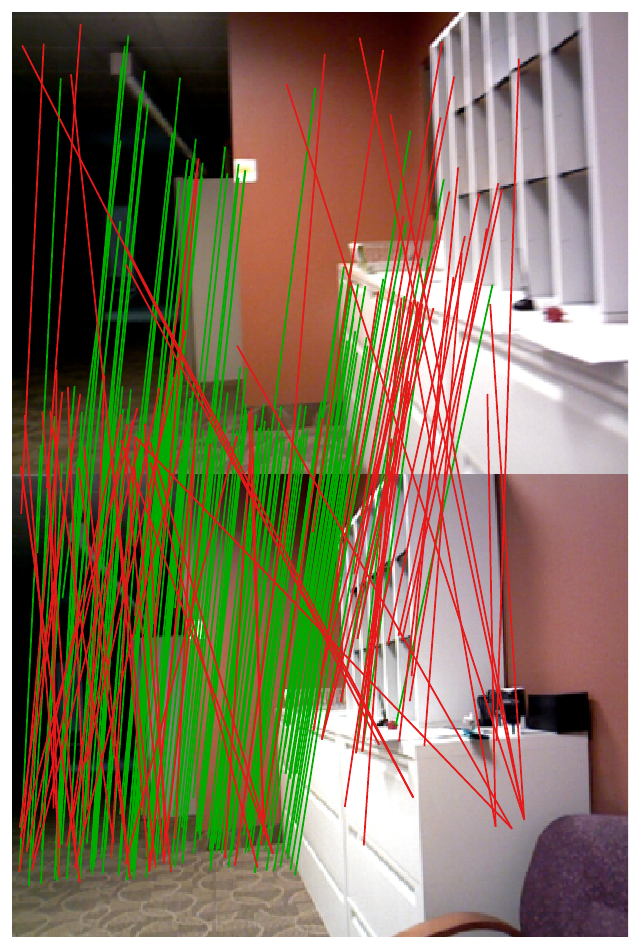}
    \caption{RANSAC}
    \label{fig:b_ours}
  \end{subfigure}
  \caption{{\bf Qualitative comparison of our results with those of RANSAC.}  {\bf (a)} Our results and {\bf (b)} RANSAC results on the ``fountain-P11'' of~\cite{Strecha08b}, {\bf (c)} Our results and {\bf (d)} RANSAC results on the ``brown-bm-3-05'' of SUN3D. We display the correspondences that the algorithms labeled as inliers. True positives are shown in green and the false ones in red. The false positives of our approach are still close to being correct, while those of RANSAC are truly wrong.}

  \label{fig:fountain_brown}
  \vspace{-1.5em}
\end{figure}

%% file: fig/pnp.tex
\begin{figure}[t]
\centering
\begin{subfigure}{.48\textwidth}
    \centering
    \includegraphics[width=1.\linewidth, trim = -10 0 -10 0, clip]{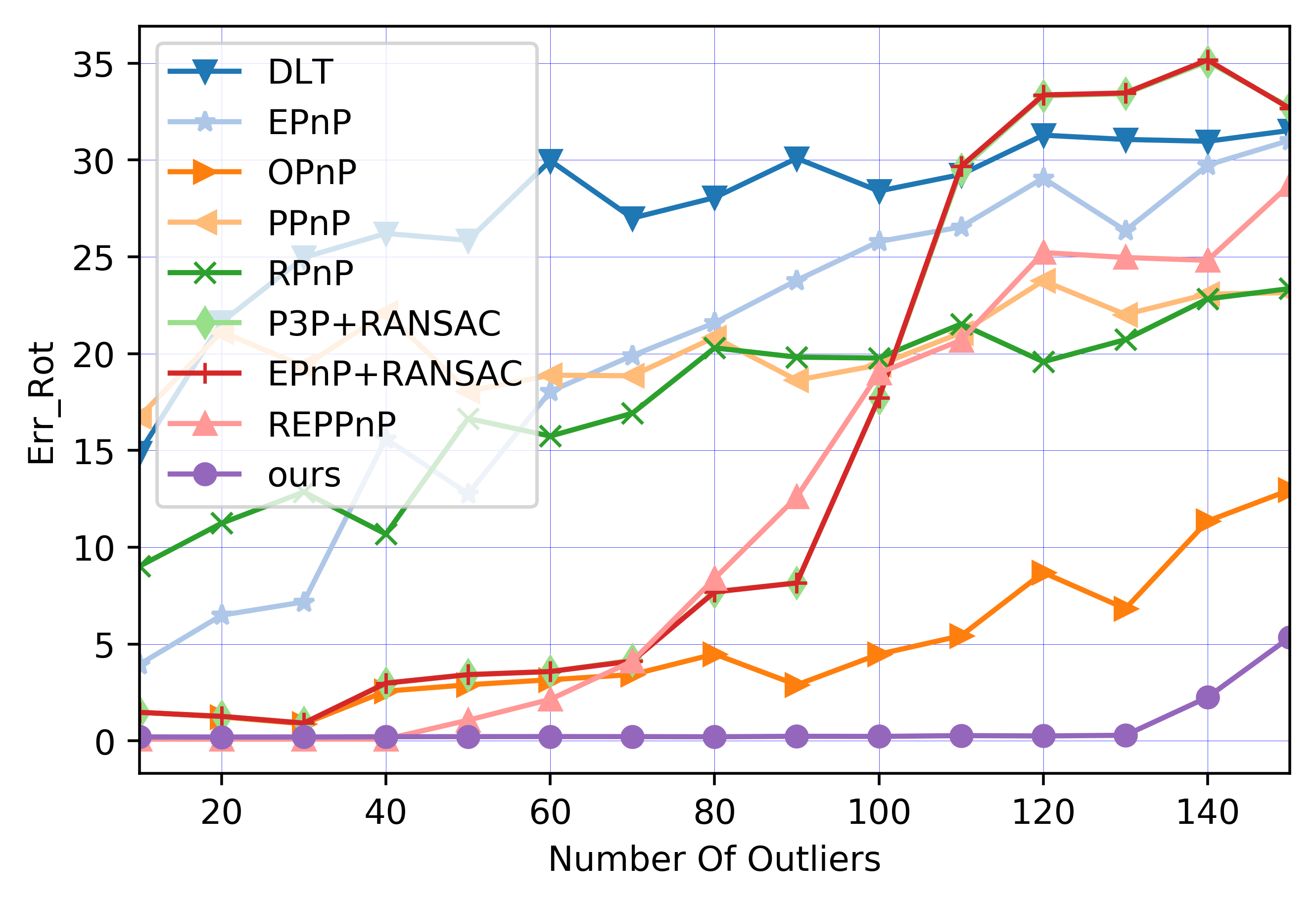}
    \caption{Rotation error (degrees)}
    \label{fig:pnp_rot}
 \end{subfigure}
 \begin{subfigure}{.48\textwidth}
    \centering
    \includegraphics[width=1.\linewidth, trim = -10 0 -10 0, clip]{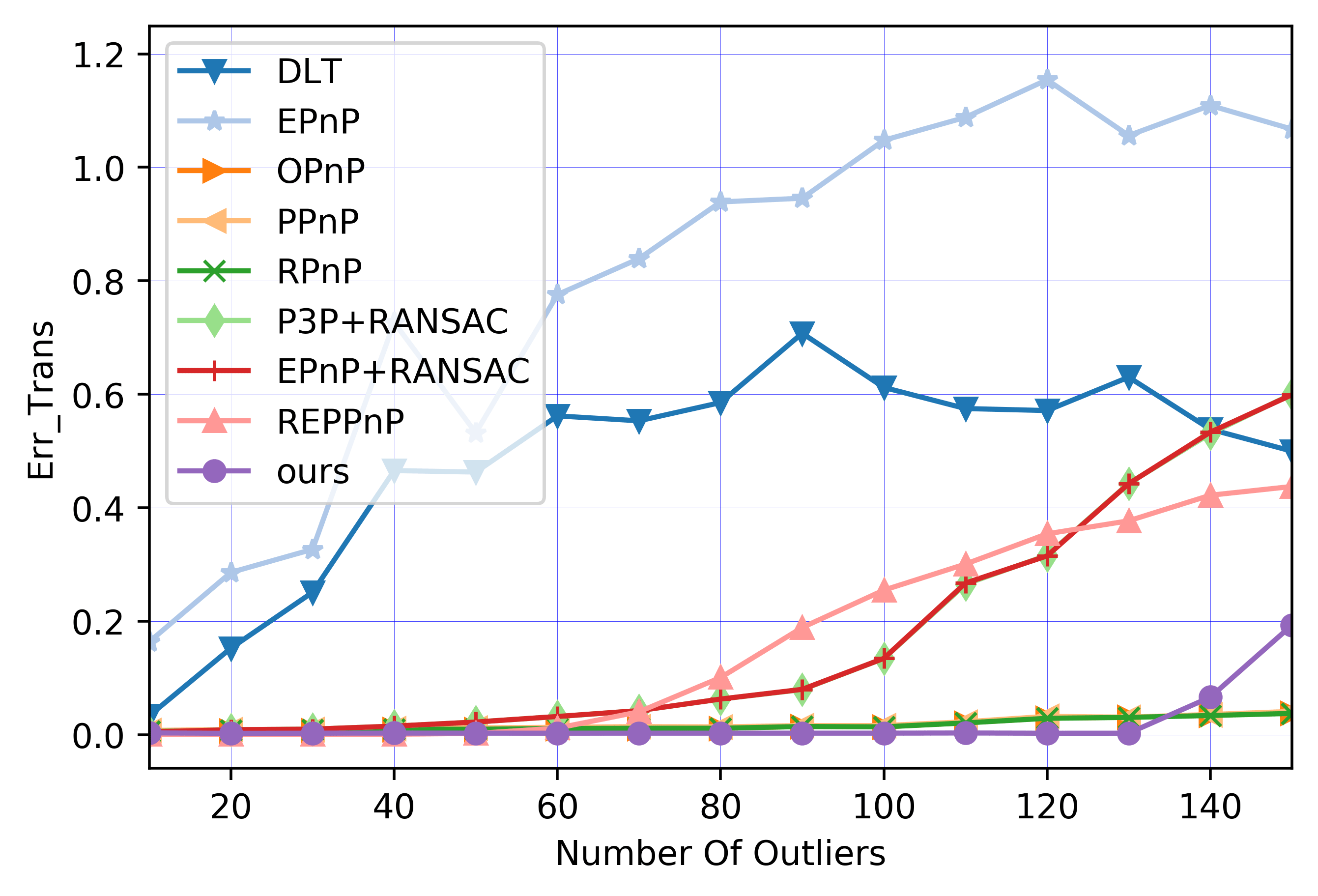}
    \caption{Translation error (normalized)}
    \label{fig:pnp_trans}
 \end{subfigure}
 \caption{{\bf PnP results.} Our method gives extremely stable results despite the abundance of outliers, whereas all compared methods perform significantly worse as the number of outliers increase. Even when these method do well for either rotation or translation, they do not perform well on both. Ours, on the other hand gives near zero error for both measures up to 130 outliers (i.e., 65\%).}
 \label{fig:pnp}
 \vspace{-0.2em}
 \end{figure}


%% file: fig/eigen_loss.tex
\begin{figure}[t]
\centering
  \begin{subfigure}{.48\textwidth}
    \centering
    \includegraphics[width=1.\linewidth, trim = -10 0 -10 0, clip]{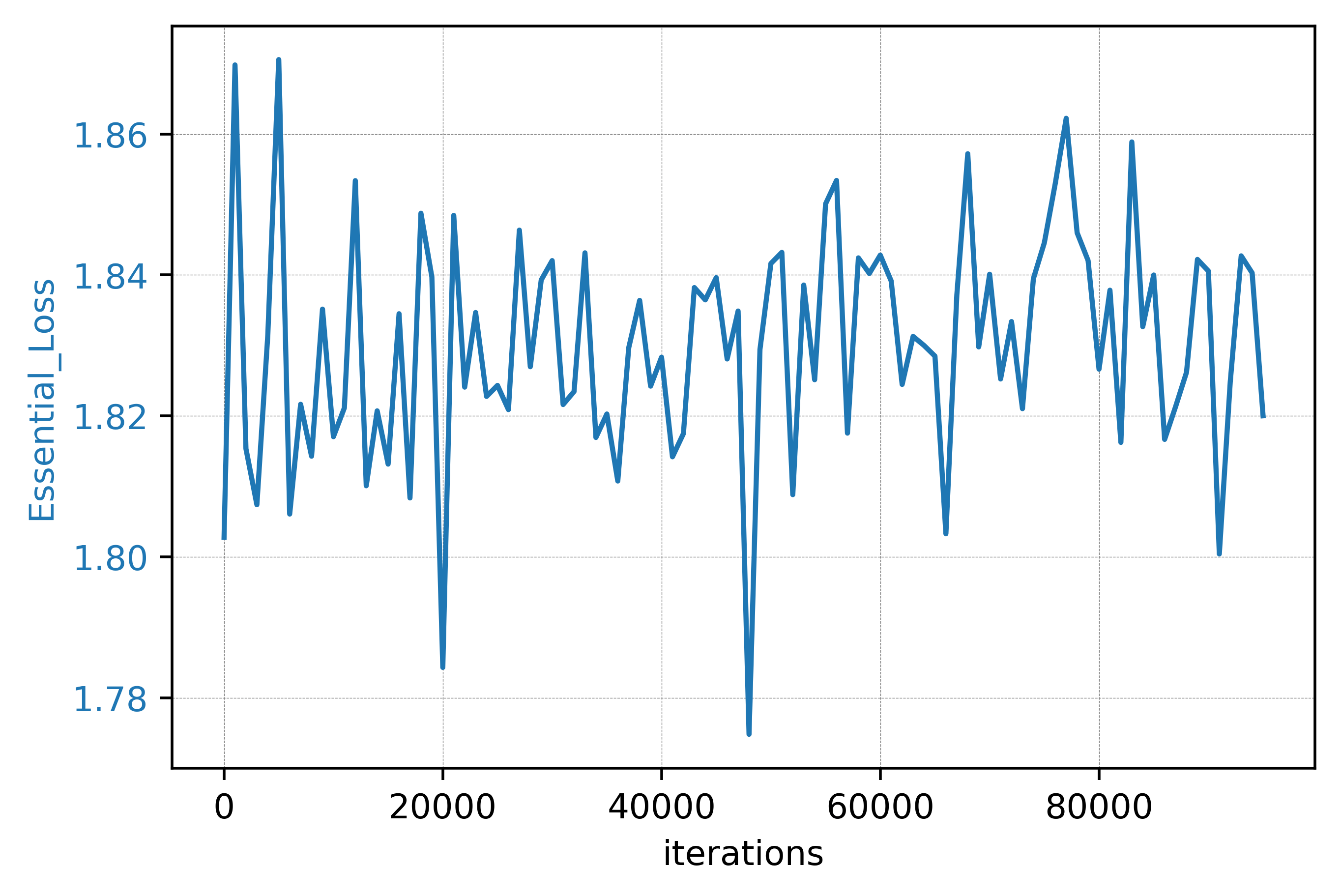}
    \caption{Loss with eigendecomposition}
  \end{subfigure}
  \begin{subfigure}{.48\textwidth}
    \centering
    \includegraphics[width=1.\linewidth, trim = -10 0 -10 0, clip]{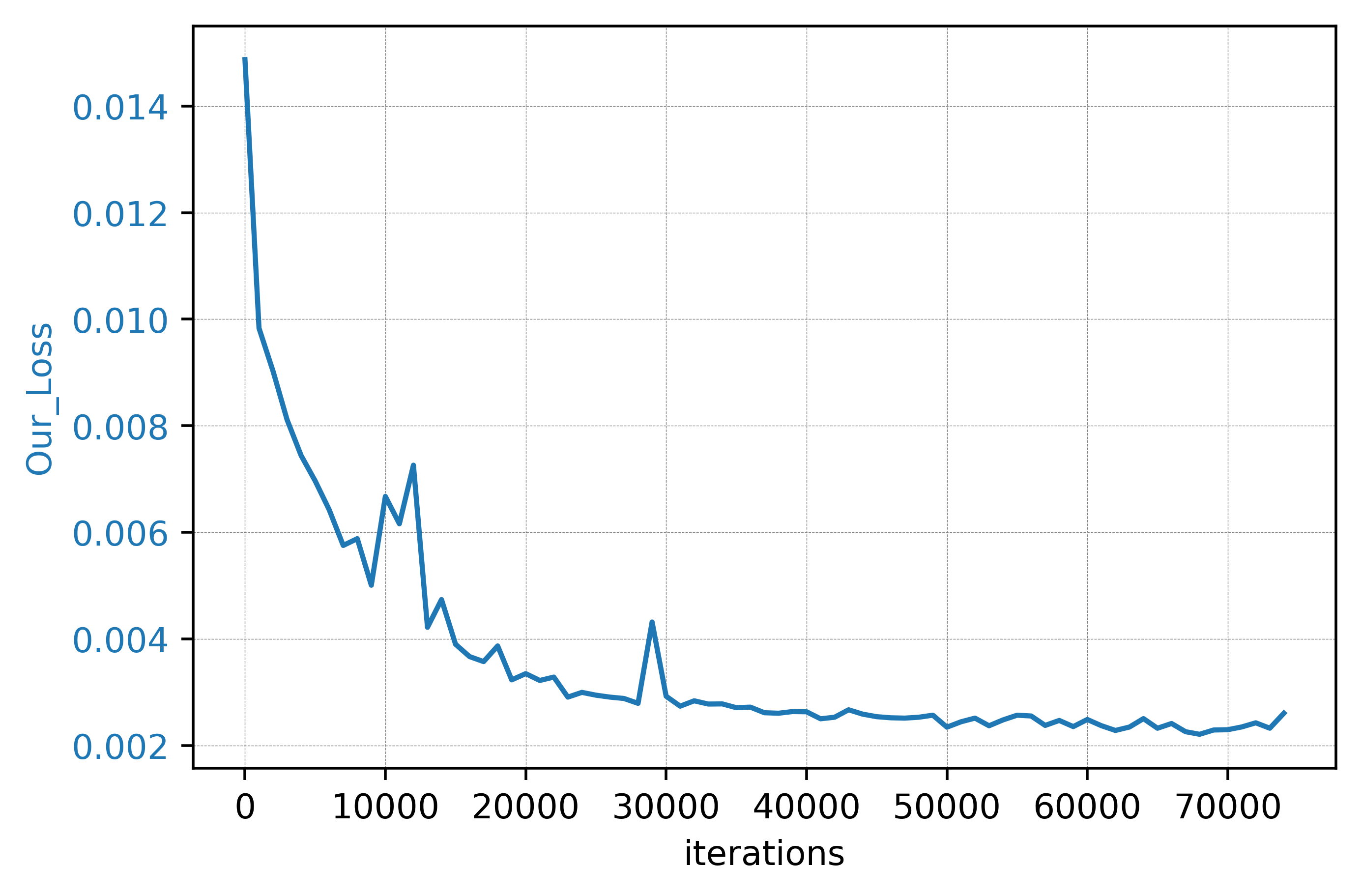}
    \caption{Loss with our approach}
  \end{subfigure}
  \caption{{\bf Loss evolution for the PnP problem.} We compare the loss based on explicit eigendecomposition with our loss.  Despite our best efforts, we were not able to make the eigendecomposition-based loss converge into anything meaningful, whereas our loss function converges nicely.}
  \label{fig:eigen_loss}
 \vspace{-1.5em}
\end{figure}

%% file: tex/conclusion.tex
\section{Conclusion}

We have introduced a novel approach to training deep networks that rely on
  losses computed from an eigenvector corresponding to a zero eigenvalue of a matrix defined by the network's output. Our loss does not suffer from the numerical
  instabilities of analytical differentiation of eigendecomposition, and converges to
  the correct solution much faster. We have demonstrated the
  effectiveness of our method  on the tasks of keypoint matching in real images and outlier rejection for the PnP problem. In both cases, our new loss has allowed us to achieve state-of-the-art results.

  Since many Computer Vision tasks rely on least-square solutions to linear systems, we will investigate the use of our approach for other ones, such as homography estimation. Furthermore, we hope that our work will contribute to imbuing Deep Learning techniques with traditional
  Computer Vision knowledge, thus avoiding discarding decades of valuable research, and leading to more principled frameworks.


%% file: tex/acknowledgement.tex
\section{Acknowledgements}
This research was supported in part by the National Natural Science Foundation of China under Grant 61603291 and the program of introducing talents of discipline to university B13043. This work was also supported in part by the Swiss Commission for Technology and Innovation. This work is performed when Zheng Dang was visiting the CVLab, EPFL, Switzerland.

%% file: tex/supplementary.tex
\begin{subappendices}
\renewcommand{\thesection}{\arabic{section}}
\section{Appendix}
We provide additional details about the results presented in Section~5 of the main paper.

\subsection{Plane Fitting}

As mentioned in Section~5.1 of the main paper, we tested different learning rates in the range $[10^{-5}, 1]$. Figs.~\ref{fig:adam} and~\ref{fig:gd} depict the learning curves for both our loss and the standard SVD/Eigh-based loss for several different ones, using either Adam or vanilla gradient descent as optimizer. As can be seen from the different plots, our approach always converges and correctly finds the inliers, as indicated by the bar plots on the right. While SVD/Eigh do converge when using Adam, this requires an eigenvector switch and learning fails when using GD. Furthermore, some inliers are systematically classified as outliers and vice-versa.

\input{fig/adam}
\input{fig/gd}

\subsection{Keypoint Matching}

Here, we provide the detailed keypoint matching results on the SUN3D and Strecha~\cite{Strecha08b} datasets. Specifically, we compare the mAP of the baselines and of our model on the individual sequences of these datasets in Figs.~\ref{fig:arc5},~\ref{fig:arc10} and~\ref{fig:arc20} for error thresholds of 5, 10 and 20, respectively. Note that the general trend is the same as the average  one reported in the main paper, with our method essentially performing on par with the state-of-the-art method of~\cite{Yi18a}, but without the need for pre-training with a different loss.

\input{fig/arc5}
\input{fig/arc10}
\input{fig/arc20}

\subsection{PnP}

We evaluated our PnP approach on real data, using the dataset of~\cite{Heinly15}. Specifically, the 3D points in this dataset were obtained using the Structure-from-Motion algorithm of~\cite{Wu13}, which also provides a rotation matrix and translation vector for each image. We treat these rotations and translations as ground-truth to compare different PnP algorithms. Given a pair of images, we extract SIFT features at the reprojection of the 3D points in one image, and match these features to SIFT keypoints detected in the other image. This procedure produces erroneous correspondences, which a robust PnP algorithm should discard. 

In this example, we used the model trained on the synthetic data described in Section~5.3 of the main paper. Note that we apply the model {\bf without any fine-tuning}, that is, the model is only trained with purely synthetic data. We report the quantitative results of this experiments, performed on four image pairs in Tables~\ref{tab1} and~\ref{tab2}. Note that our approach yields much lower errors than all the baselines. In Fig.~\ref{fig:pnp}, we compare the reprojection of the 3D points on the input image after applying the rotation and translation obtained with our model and with EPnP+RANSAC. Note that the points obtained with our approach reproject much more closely to the ground-truth image locations than those of this baseline. Note that EPnP+RANSAC constitutes the best-performing baseline, on par with OPnP and P3P+RANSAC. For the other baselines, we are unable to provide similar figures because the errors reported in Tables~\ref{tab1} and~\ref{tab2} translate to points reprojecting outside the input image. This underscores the strength of our approach, which, despite being trained on synthetic data nevertheless works on real images.

\input{fig/table.tex}

\input{fig/pnp_supply.tex}

\end{subappendices}
 

%% file: fig/adam.tex
\begin{figure}[t]
\centering
\begin{subfigure}{.28\textwidth}
    \centering
    \includegraphics[width=1.\linewidth]{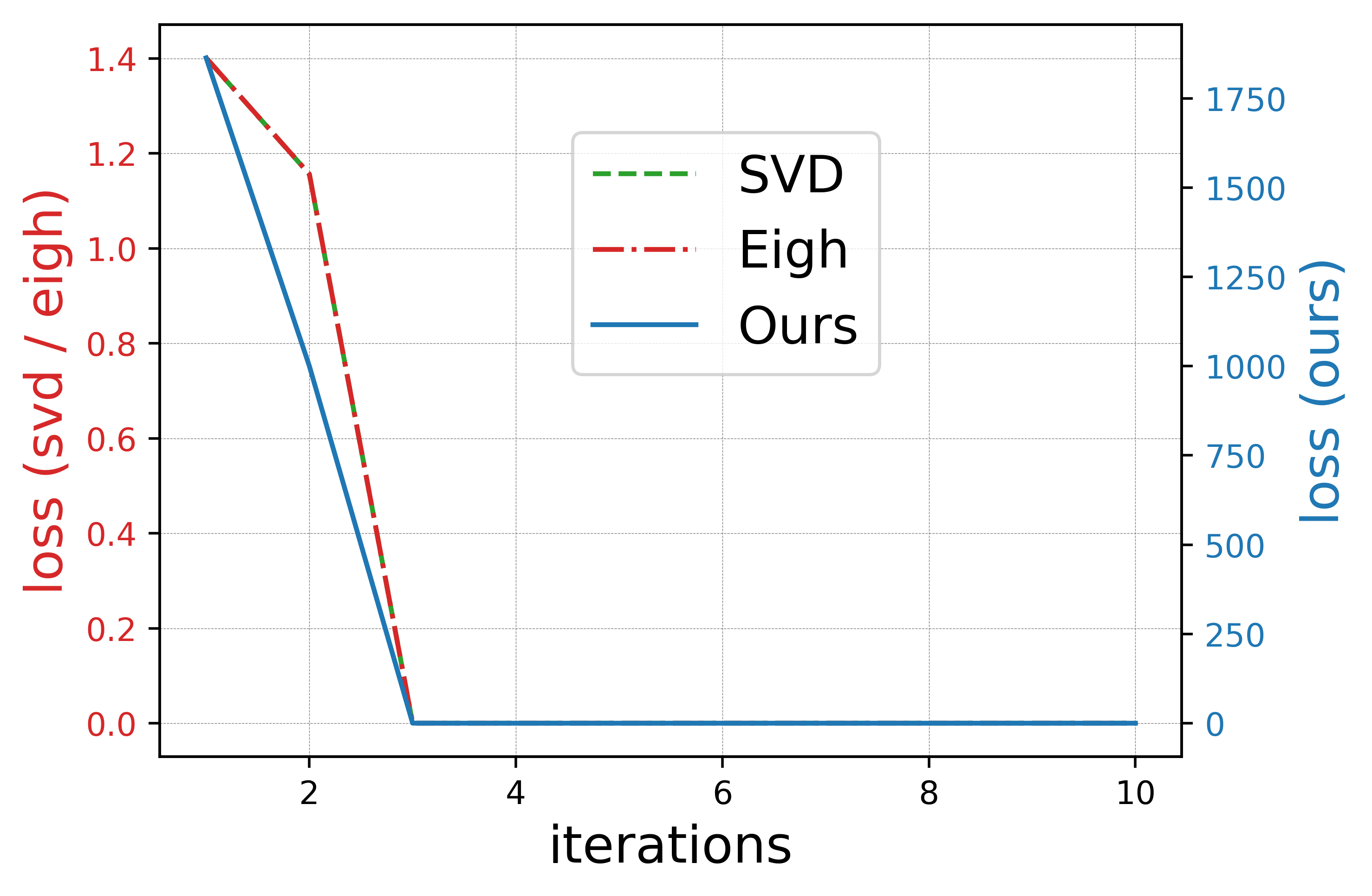}
\end{subfigure}
\begin{subfigure}{.28\textwidth}
	\centering
	\includegraphics[width=1.\linewidth]{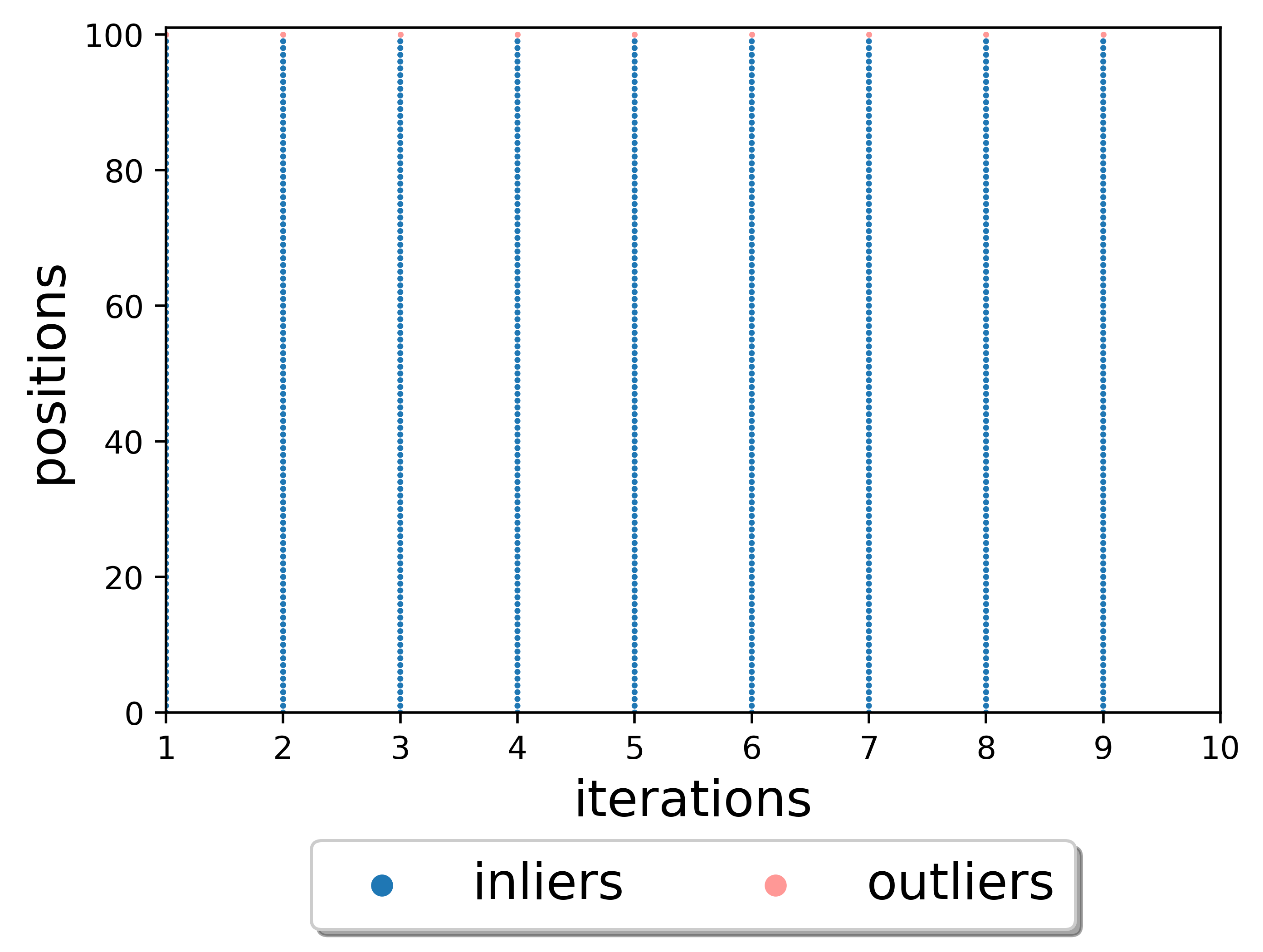}
\end{subfigure}
\begin{subfigure}{.28\textwidth}
	\centering
	\includegraphics[width=1.\linewidth]{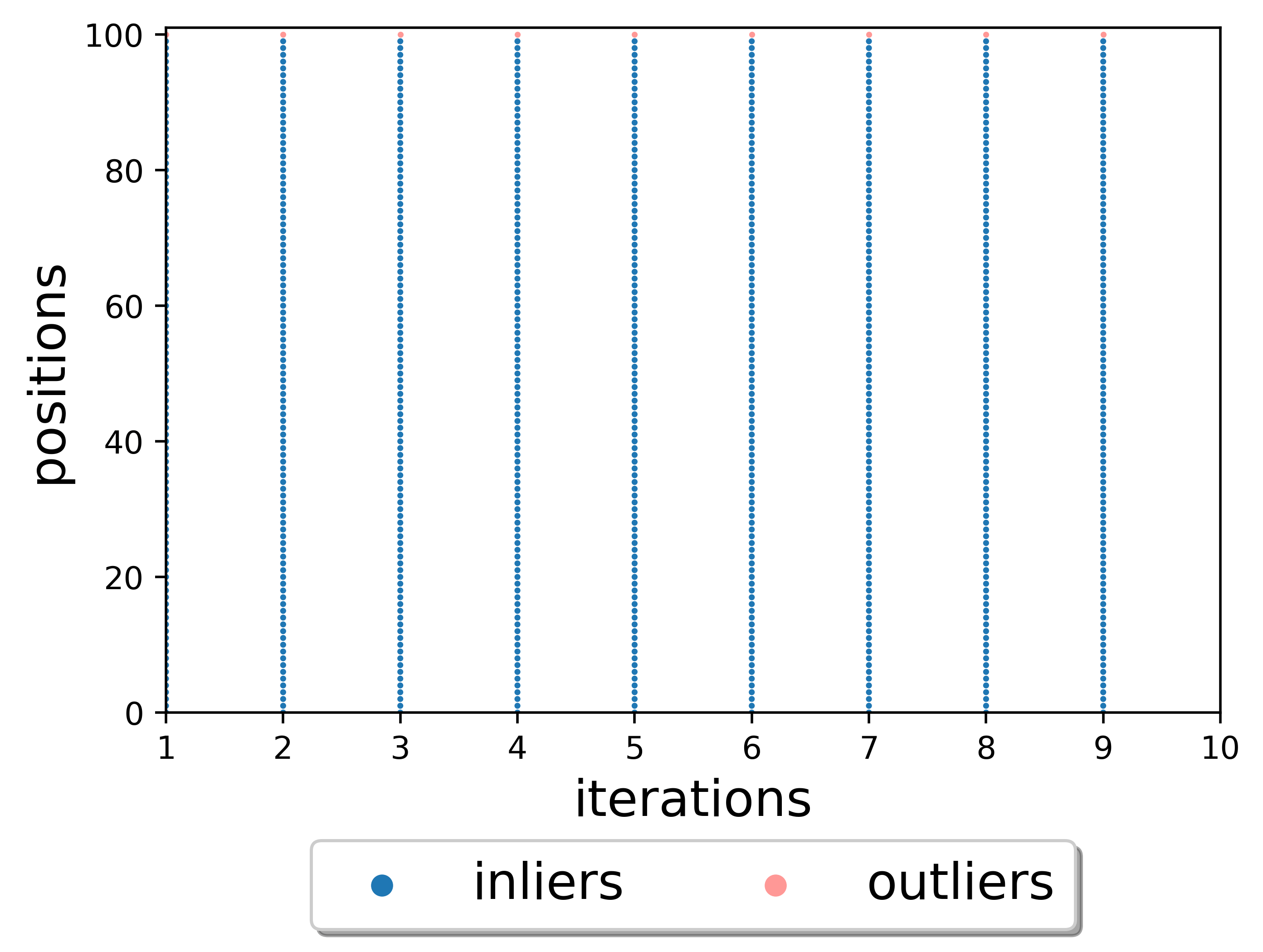}
\end{subfigure}

\begin{subfigure}{.28\textwidth}
    \centering
    \includegraphics[width=1.\linewidth]{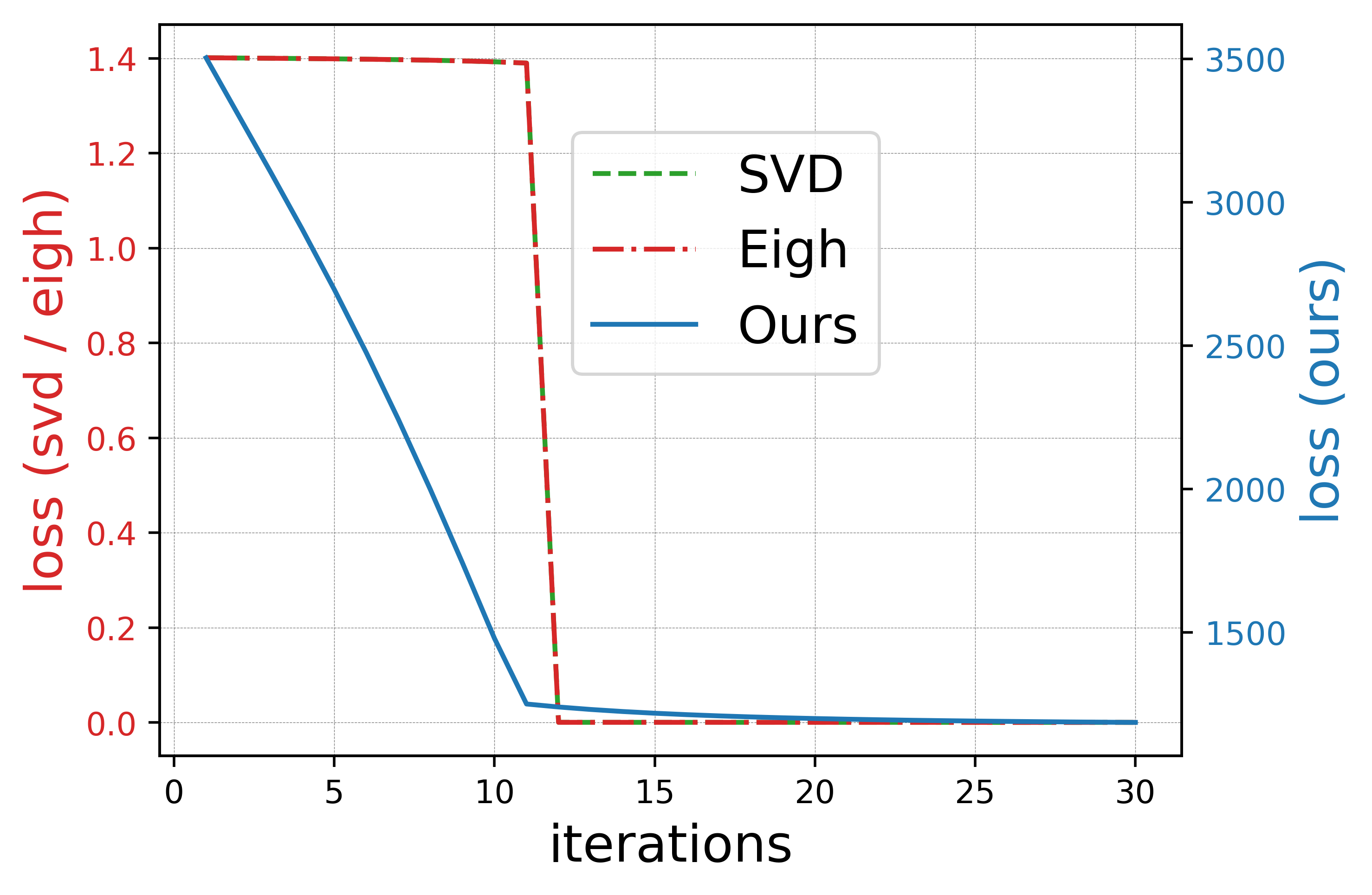}
\end{subfigure}
\begin{subfigure}{.28\textwidth}
	\centering
	\includegraphics[width=1.\linewidth]{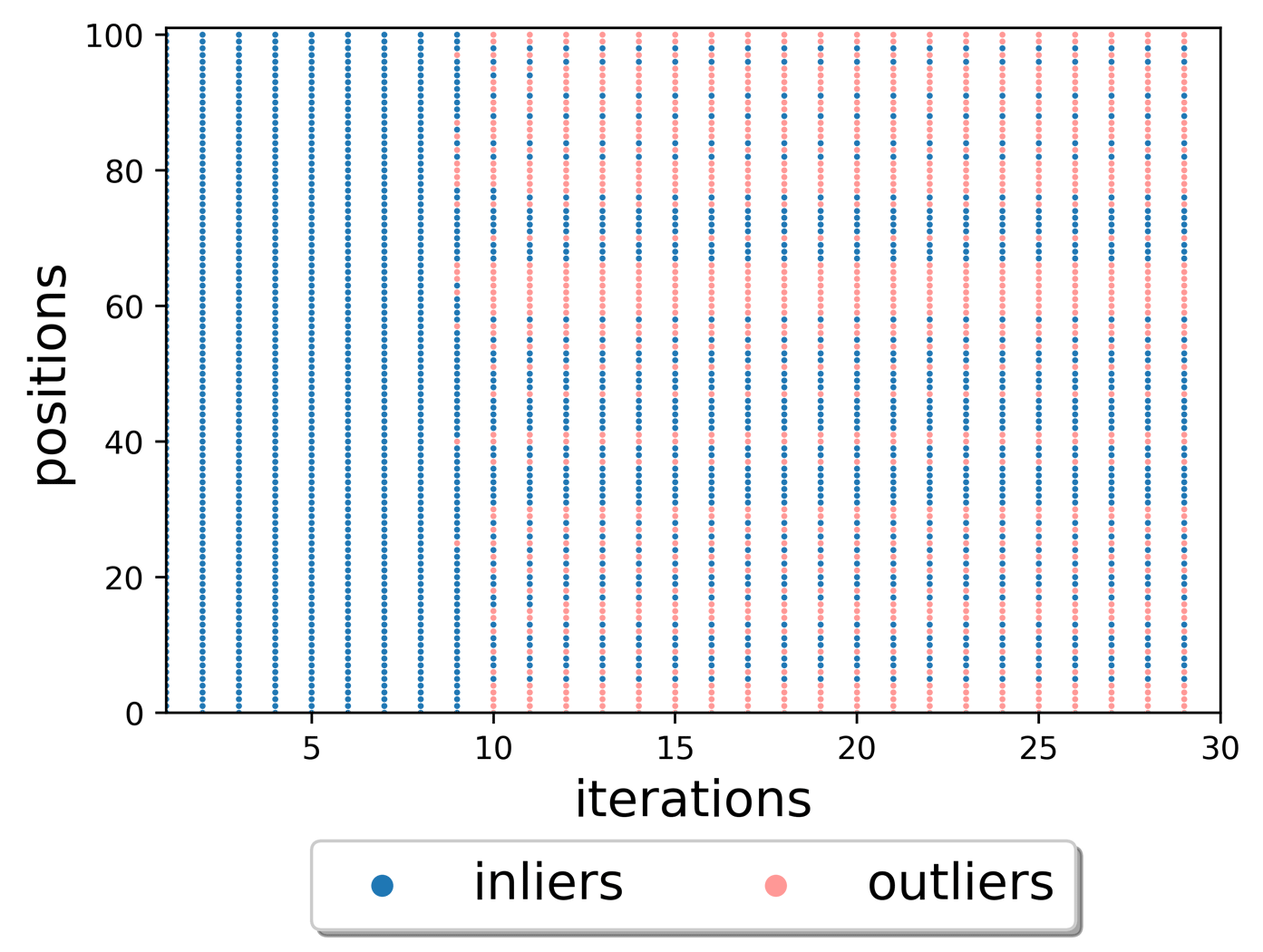}
\end{subfigure}
\begin{subfigure}{.28\textwidth}
	\centering
	\includegraphics[width=1.\linewidth]{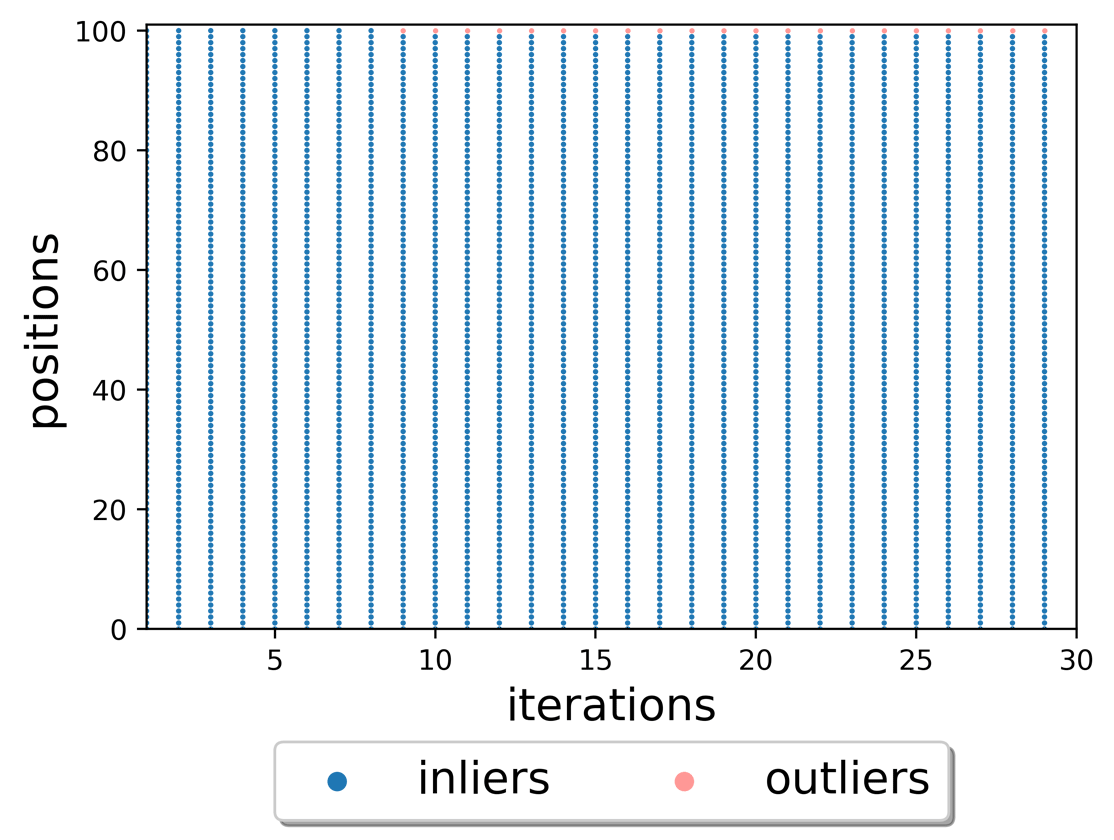}
\end{subfigure}

\begin{subfigure}{.28\textwidth}
    \centering
    \includegraphics[width=1.\linewidth]{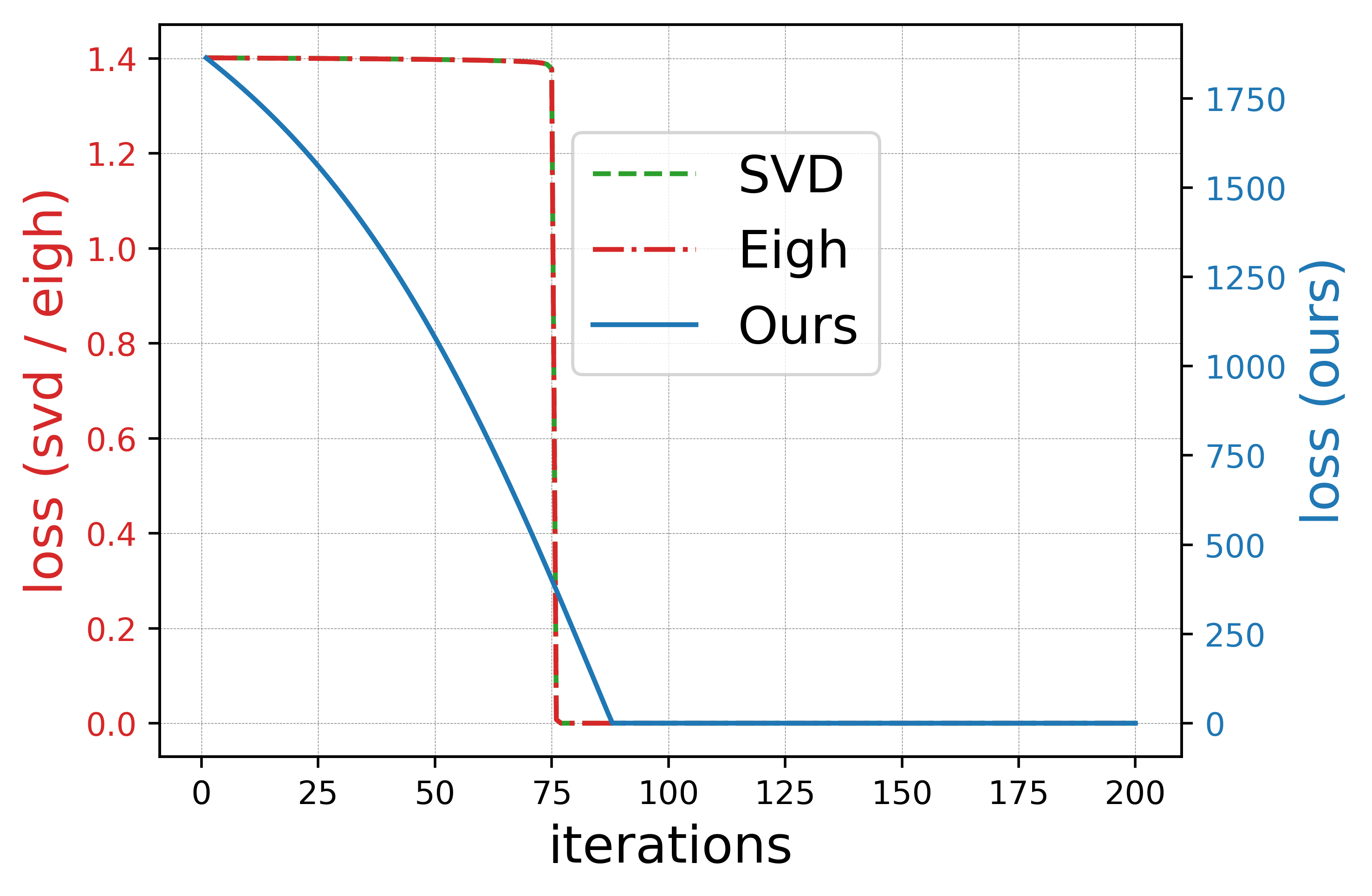}
\end{subfigure}
\begin{subfigure}{.28\textwidth}
	\centering
	\includegraphics[width=1.\linewidth]{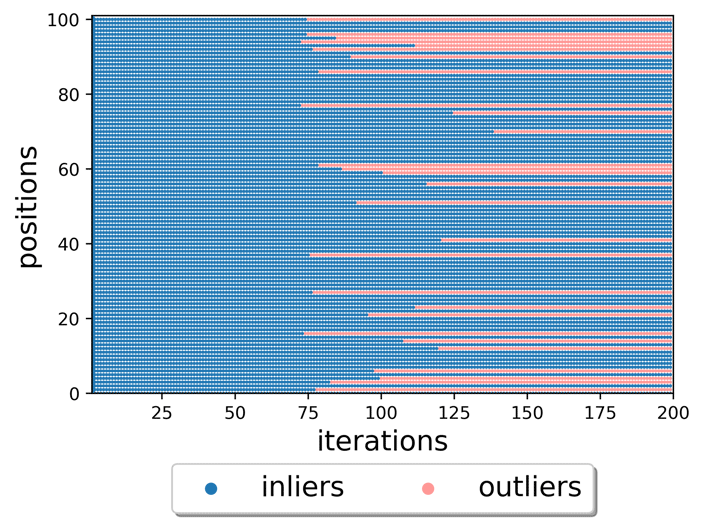}
\end{subfigure}
\begin{subfigure}{.28\textwidth}
	\centering
	\includegraphics[width=1.\linewidth]{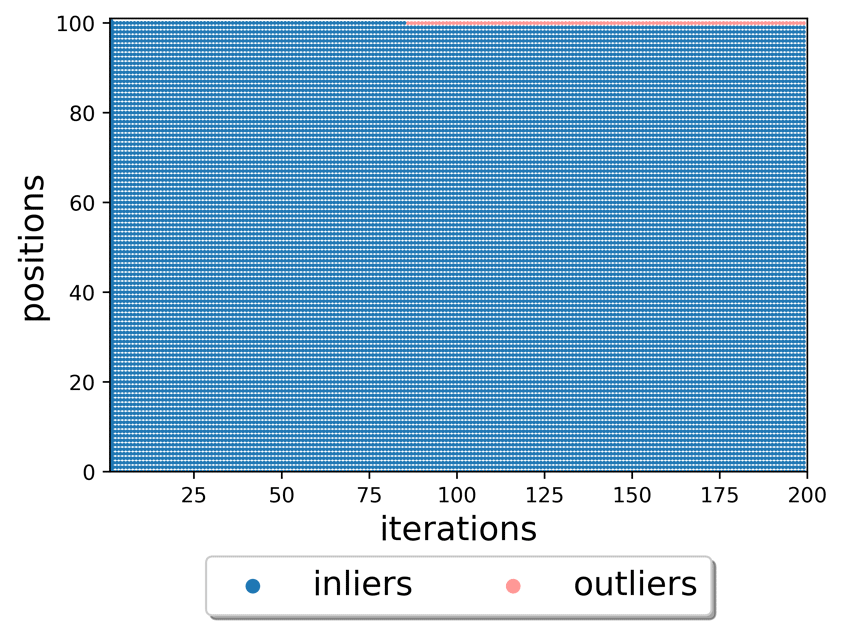}
\end{subfigure}

\begin{subfigure}{.28\textwidth}
    \centering
    \includegraphics[width=1.\linewidth]{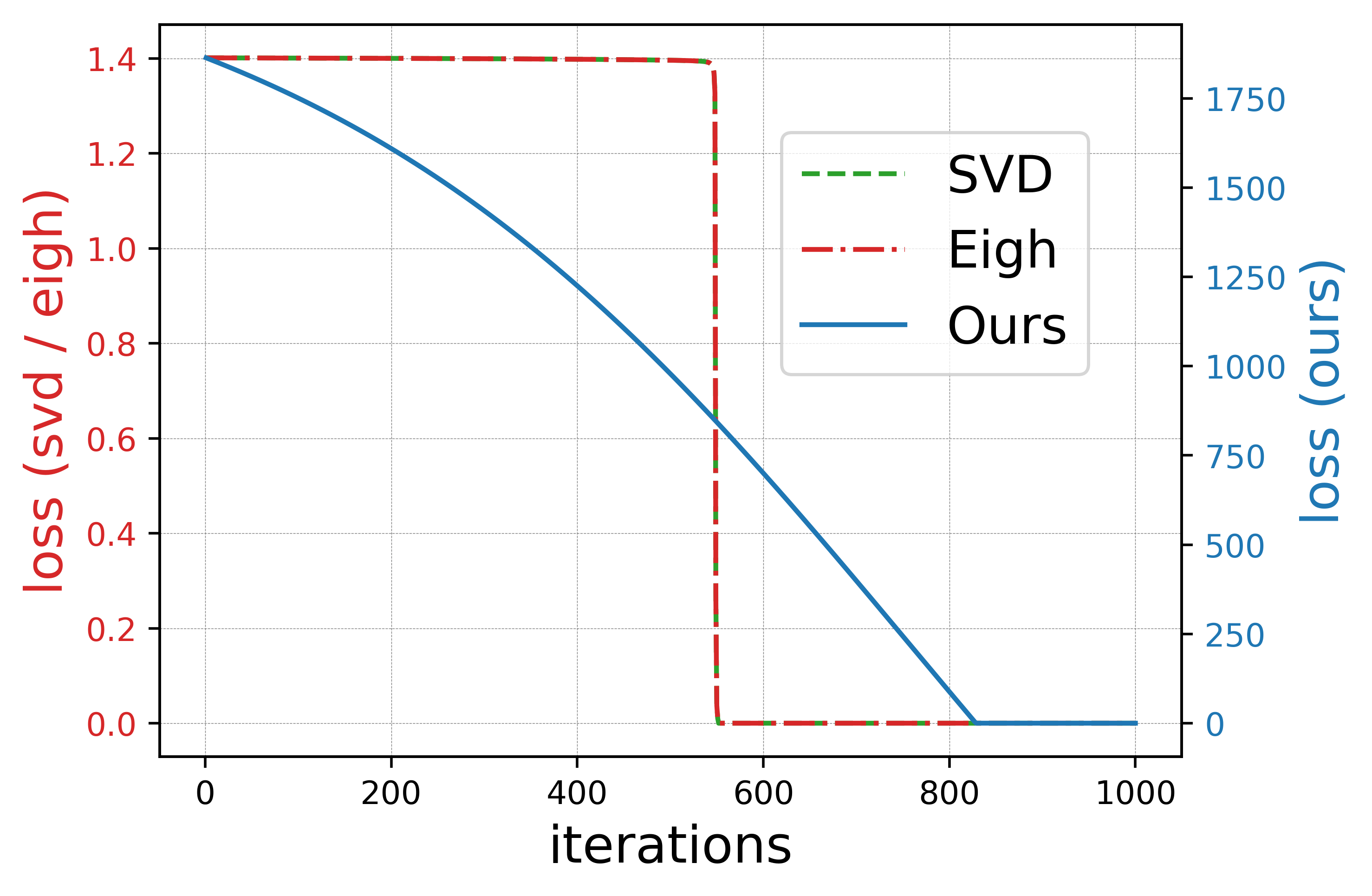}
\end{subfigure}
\begin{subfigure}{.28\textwidth}
	\centering
	\includegraphics[width=1.\linewidth]{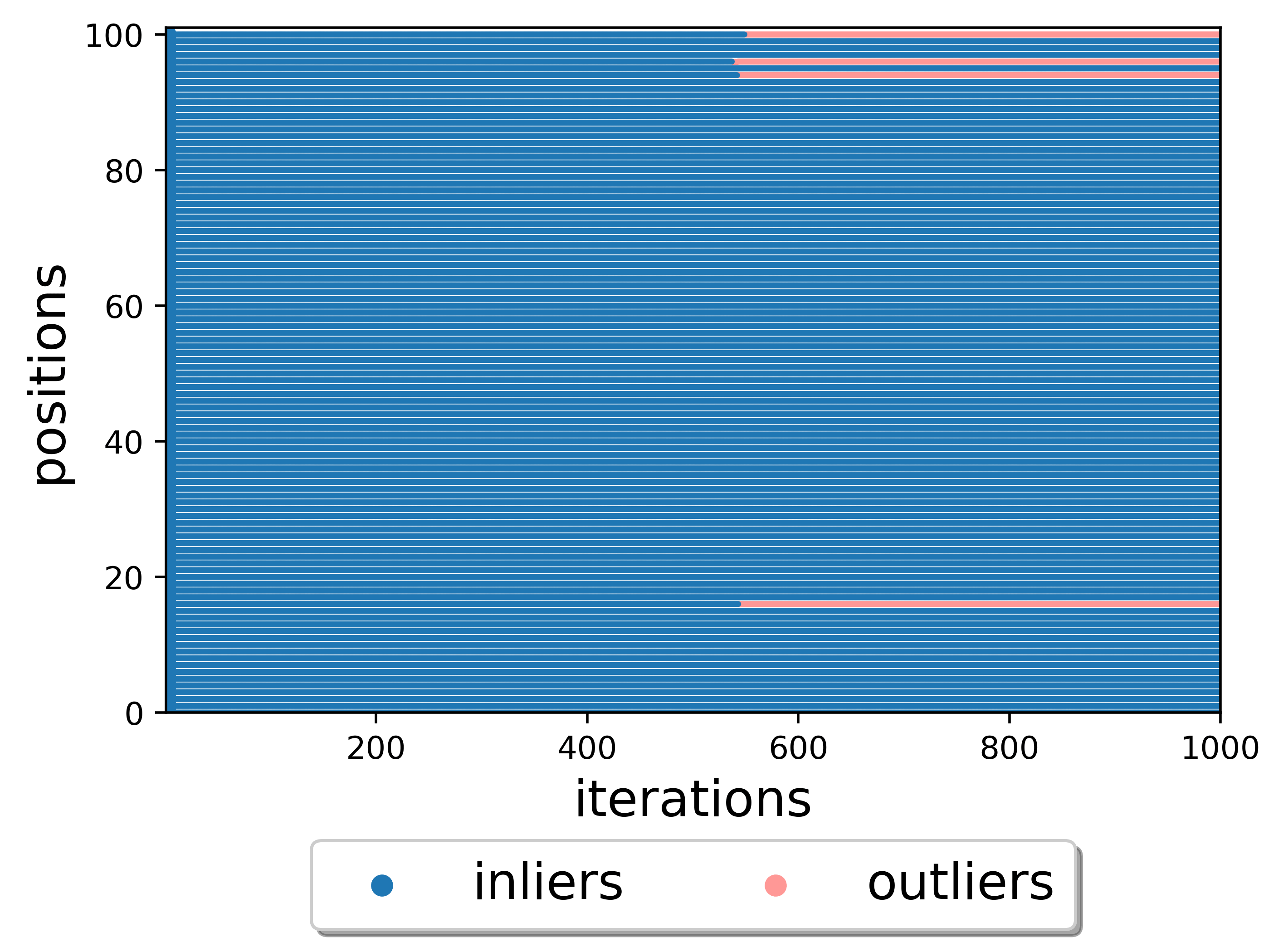}
\end{subfigure}
\begin{subfigure}{.28\textwidth}
	\centering
	\includegraphics[width=1.\linewidth]{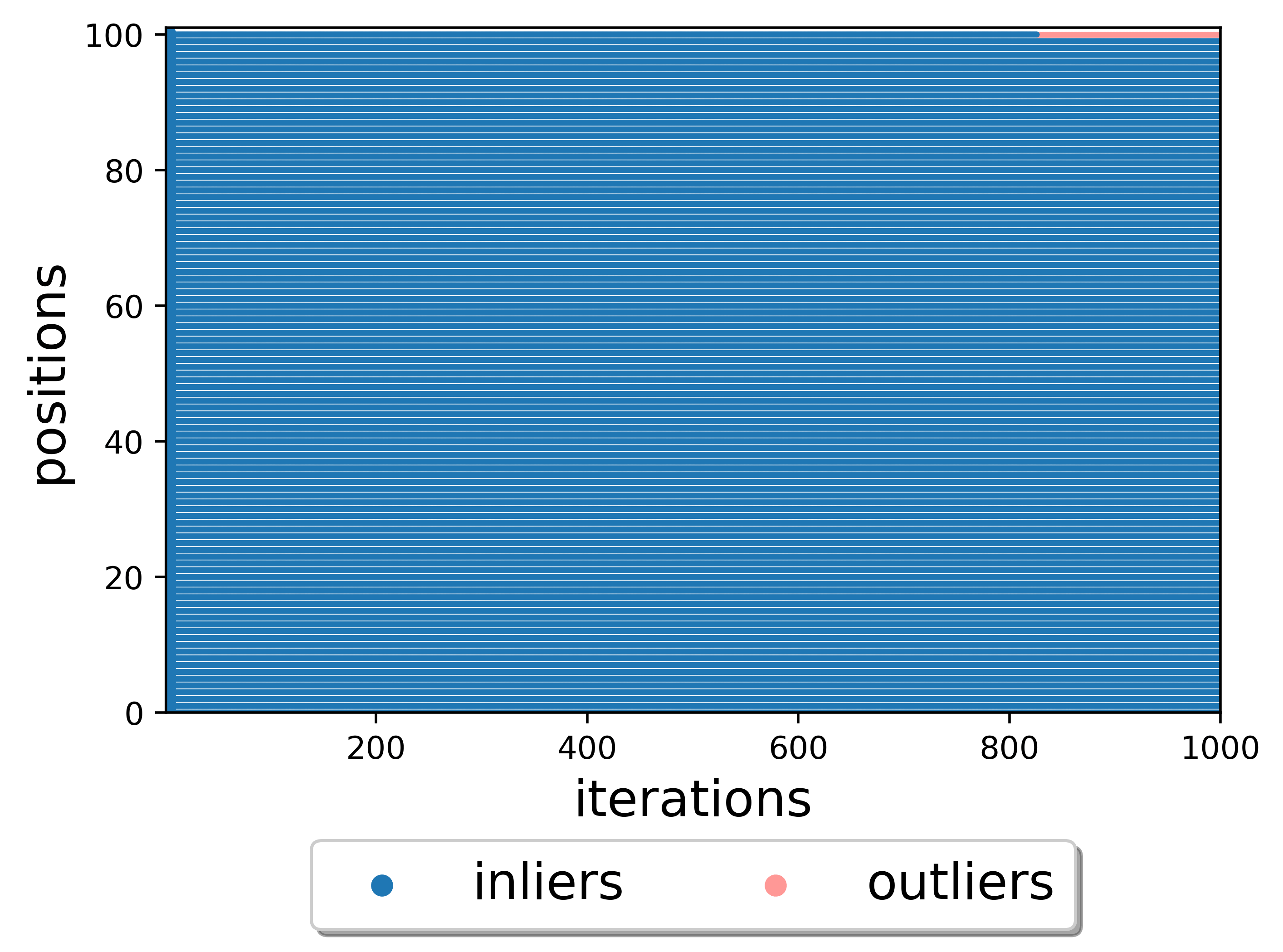}
\end{subfigure}

\begin{subfigure}{.28\textwidth}
    \centering
    \includegraphics[width=1.\linewidth]{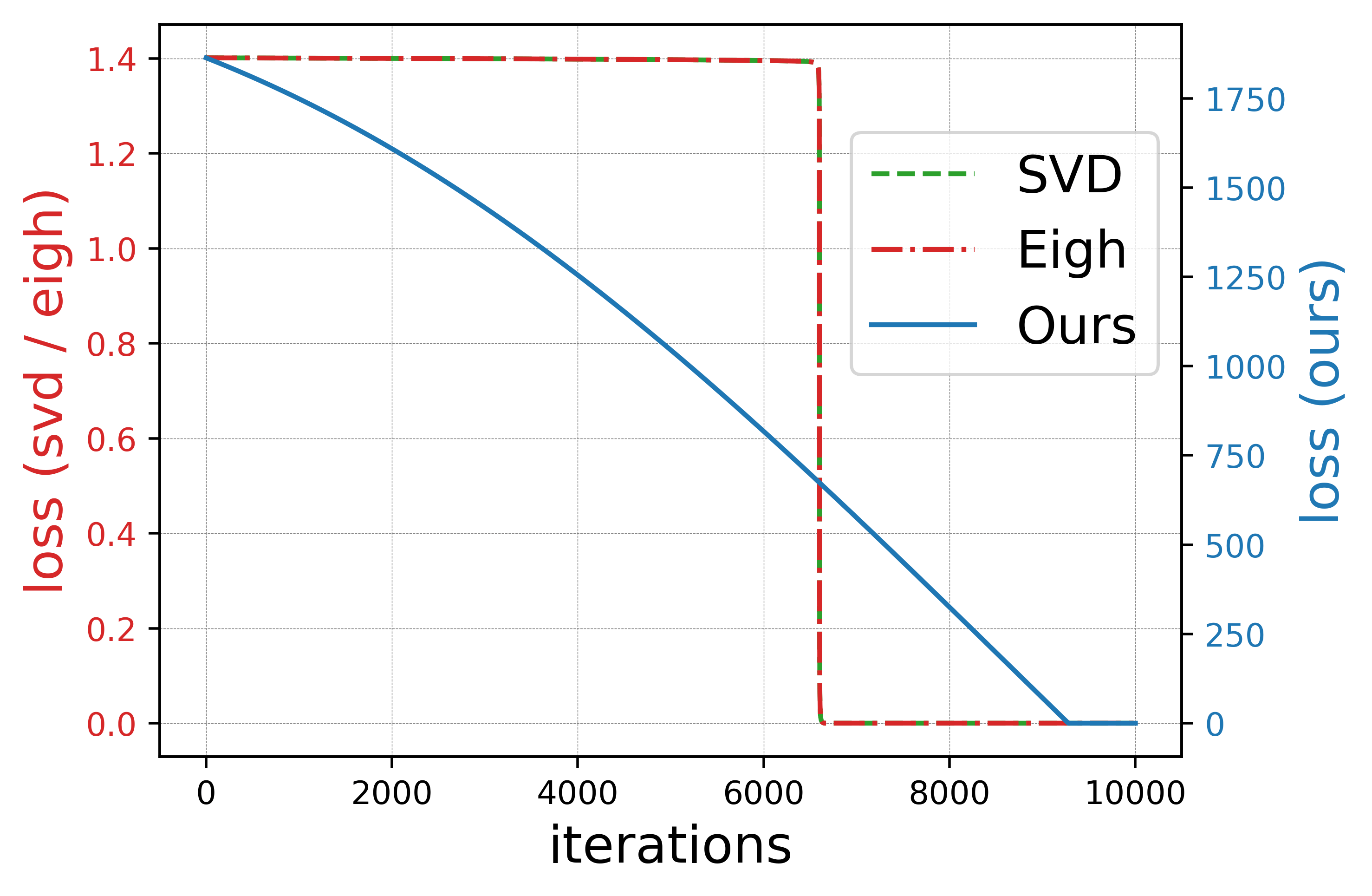}
\end{subfigure}
\begin{subfigure}{.28\textwidth}
	\centering
	\includegraphics[width=1.\linewidth]{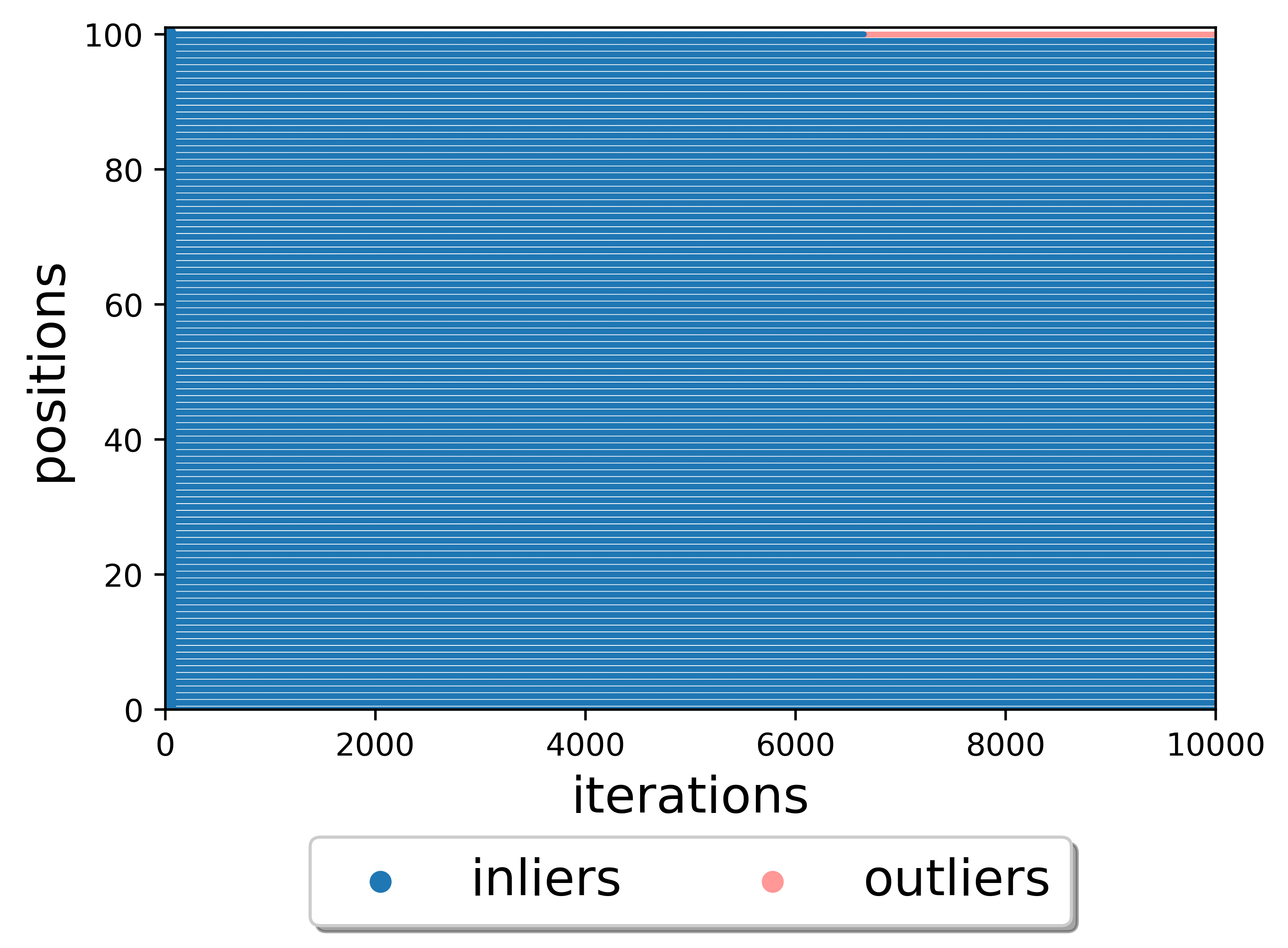}
\end{subfigure}
\begin{subfigure}{.28\textwidth}
	\centering
	\includegraphics[width=1.\linewidth]{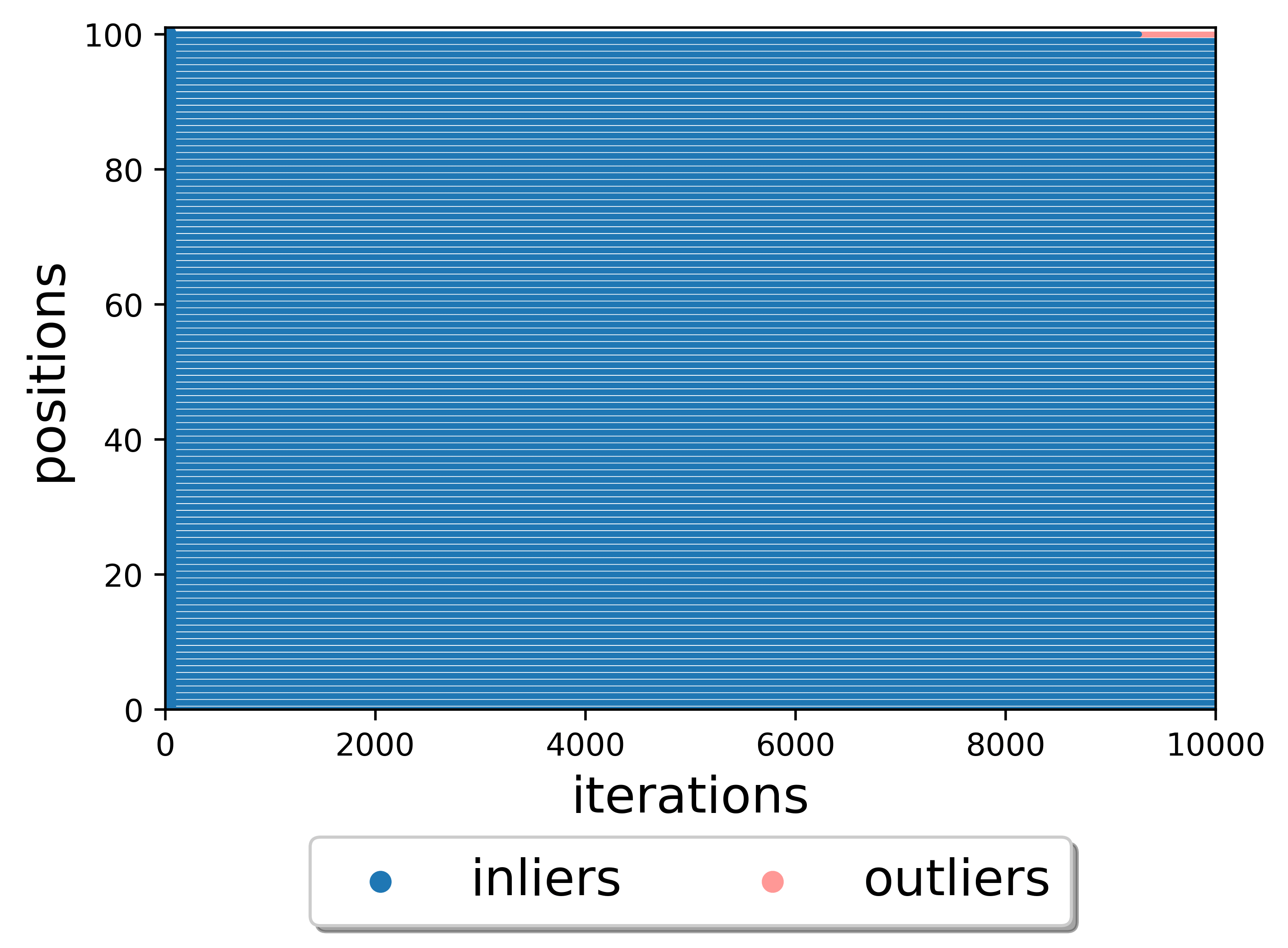}
\end{subfigure}
\begin{subfigure}{.28\textwidth}
    \centering
    \includegraphics[width=1.\linewidth]{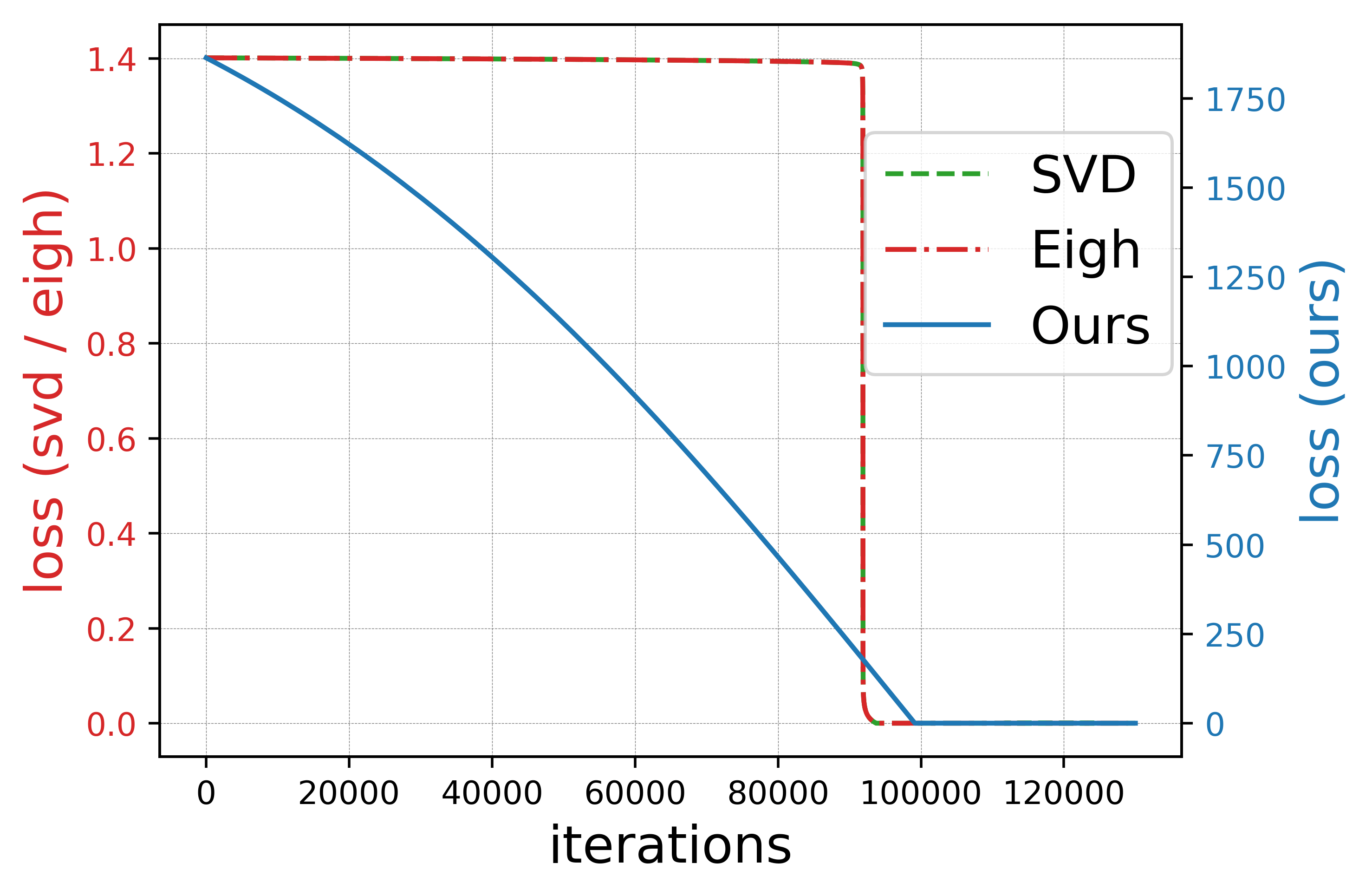}
\end{subfigure}
\begin{subfigure}{.28\textwidth}
	\centering
	\includegraphics[width=1.\linewidth]{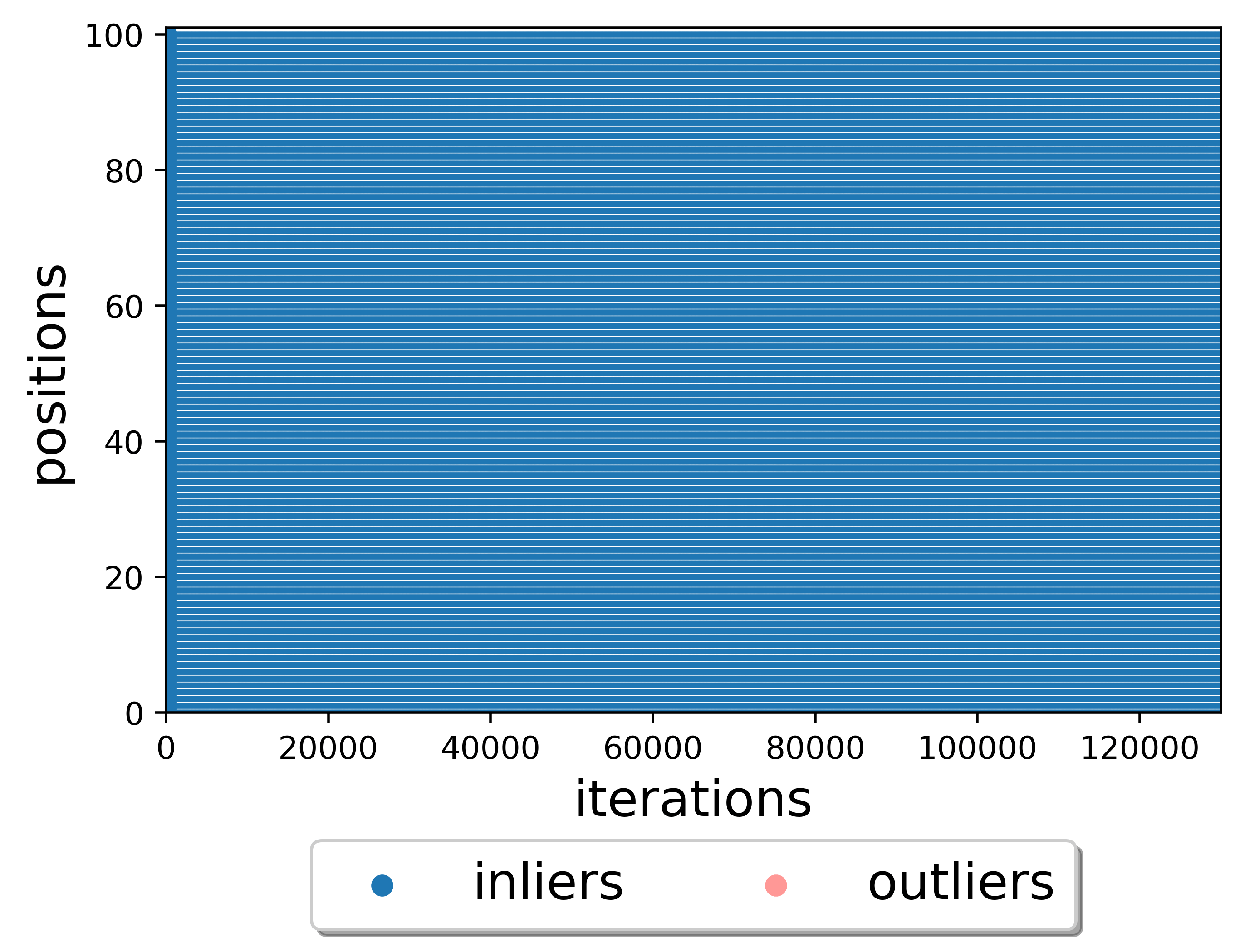}
\end{subfigure}
\begin{subfigure}{.28\textwidth}
	\centering
	\includegraphics[width=1.\linewidth]{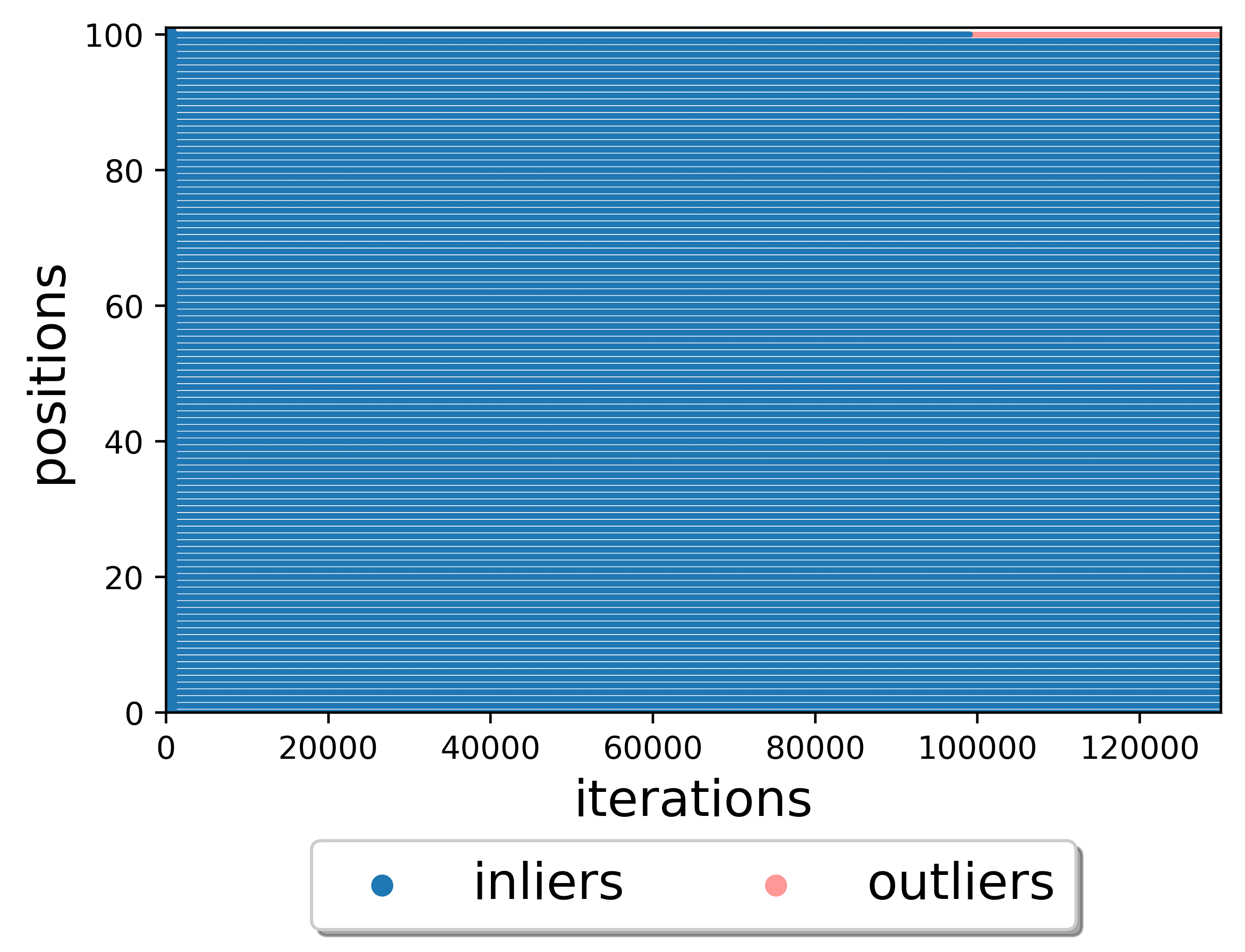}
\end{subfigure}
\caption{{\bf Loss evolution for the fitting plane problem with Adam.} The different rows correspond to different learning rates, from $10^{-5}$ to $1$. On the right, we show the loss evolution for our approach and for SVD/Eigh. On the left and middle, we show bar plots indicating the points that were classified as outliers/inliers by SVD and our approach, respectively. Note that our approach always find the correct inliers (indices 1 to 100), whereas SVD typically misclassifies points.
}
\label{fig:adam}
\end{figure}

%% file: fig/gd.tex
\begin{figure}[t]
\centering
\begin{subfigure}{.32\textwidth}
    \centering
    \includegraphics[width=1.\linewidth]{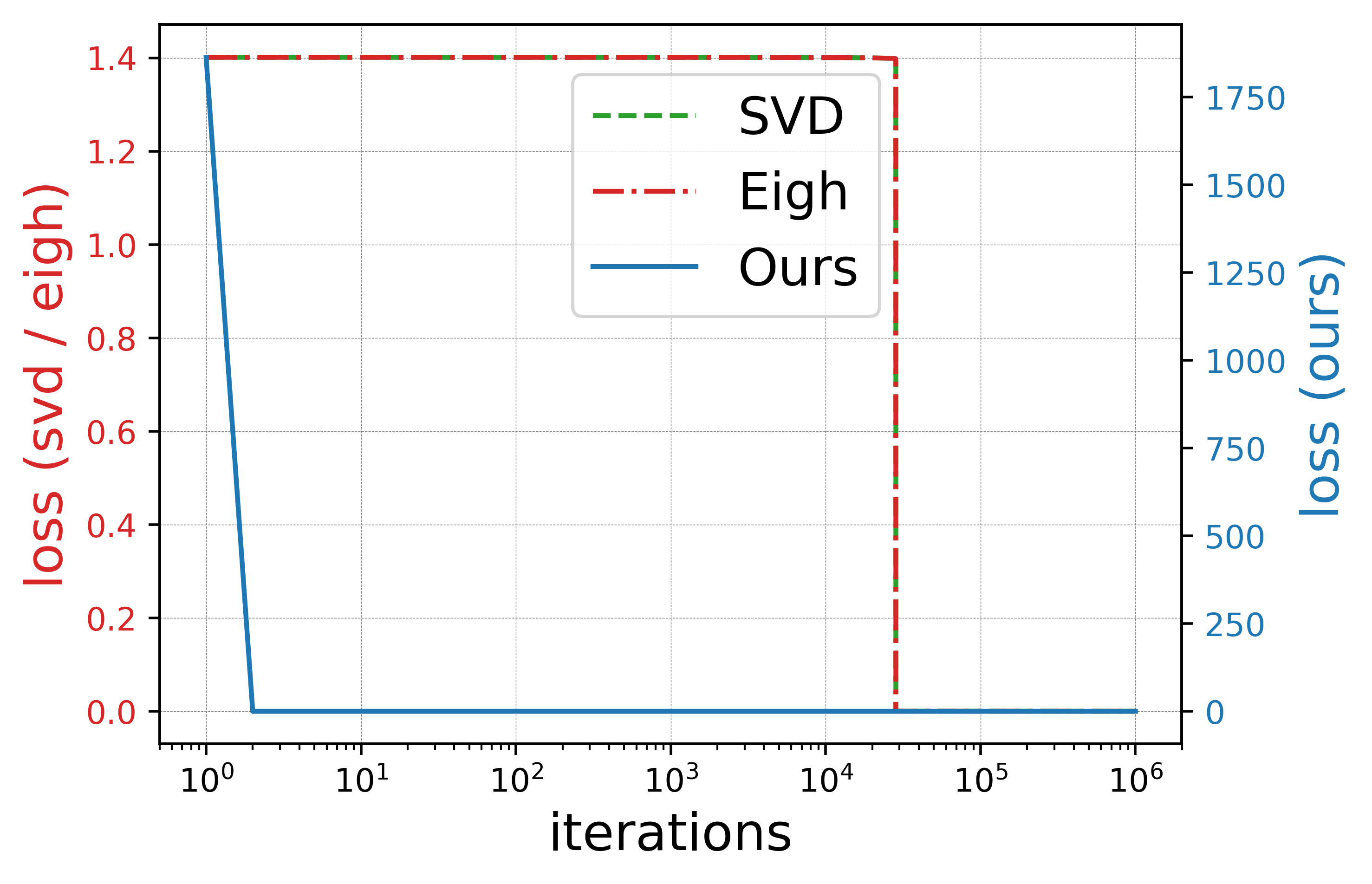}
\end{subfigure}
\begin{subfigure}{.32\textwidth}
	\centering
	\includegraphics[width=1.\linewidth]{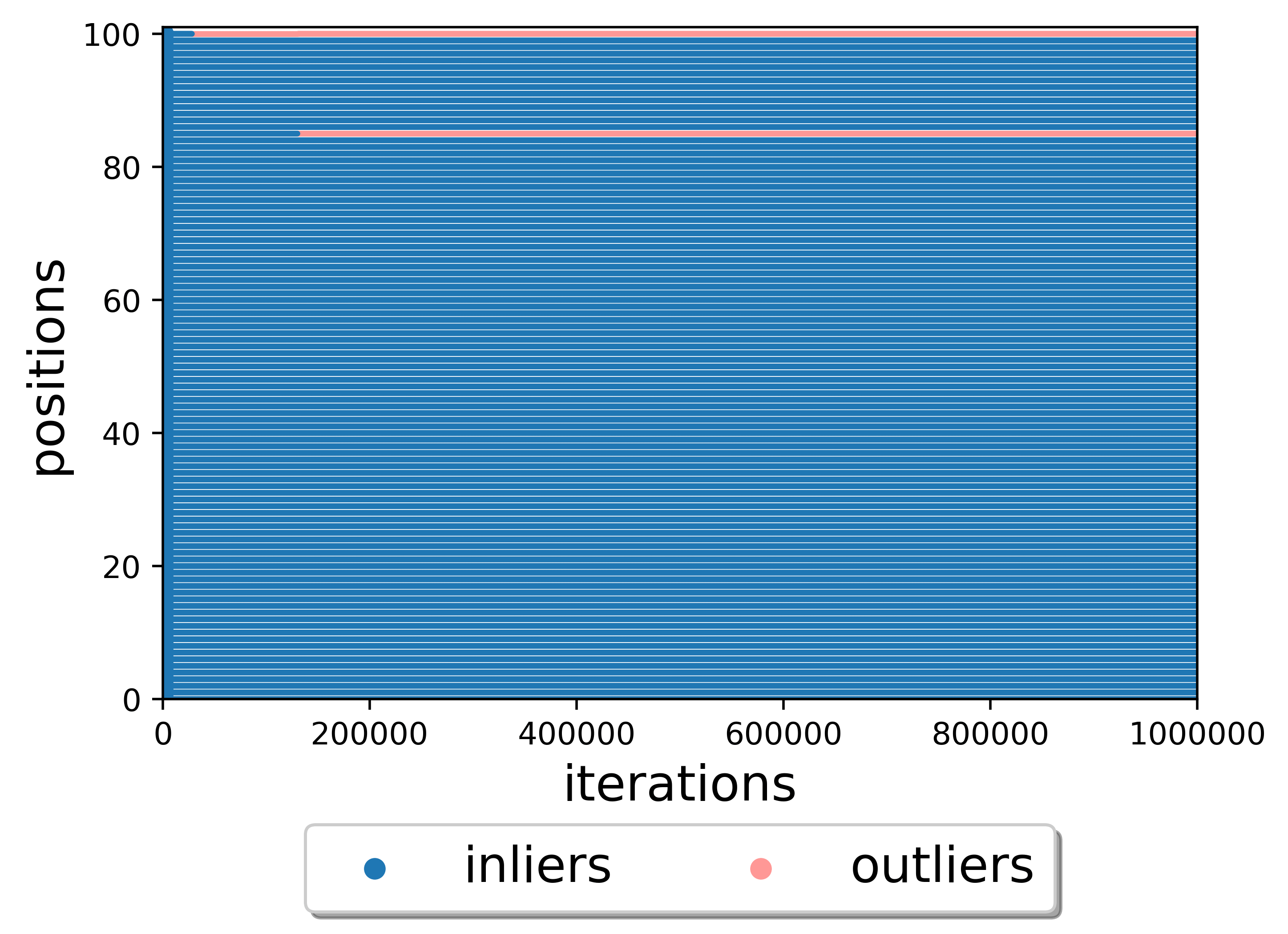}
\end{subfigure}
\begin{subfigure}{.32\textwidth}
	\centering
	\includegraphics[width=1.\linewidth]{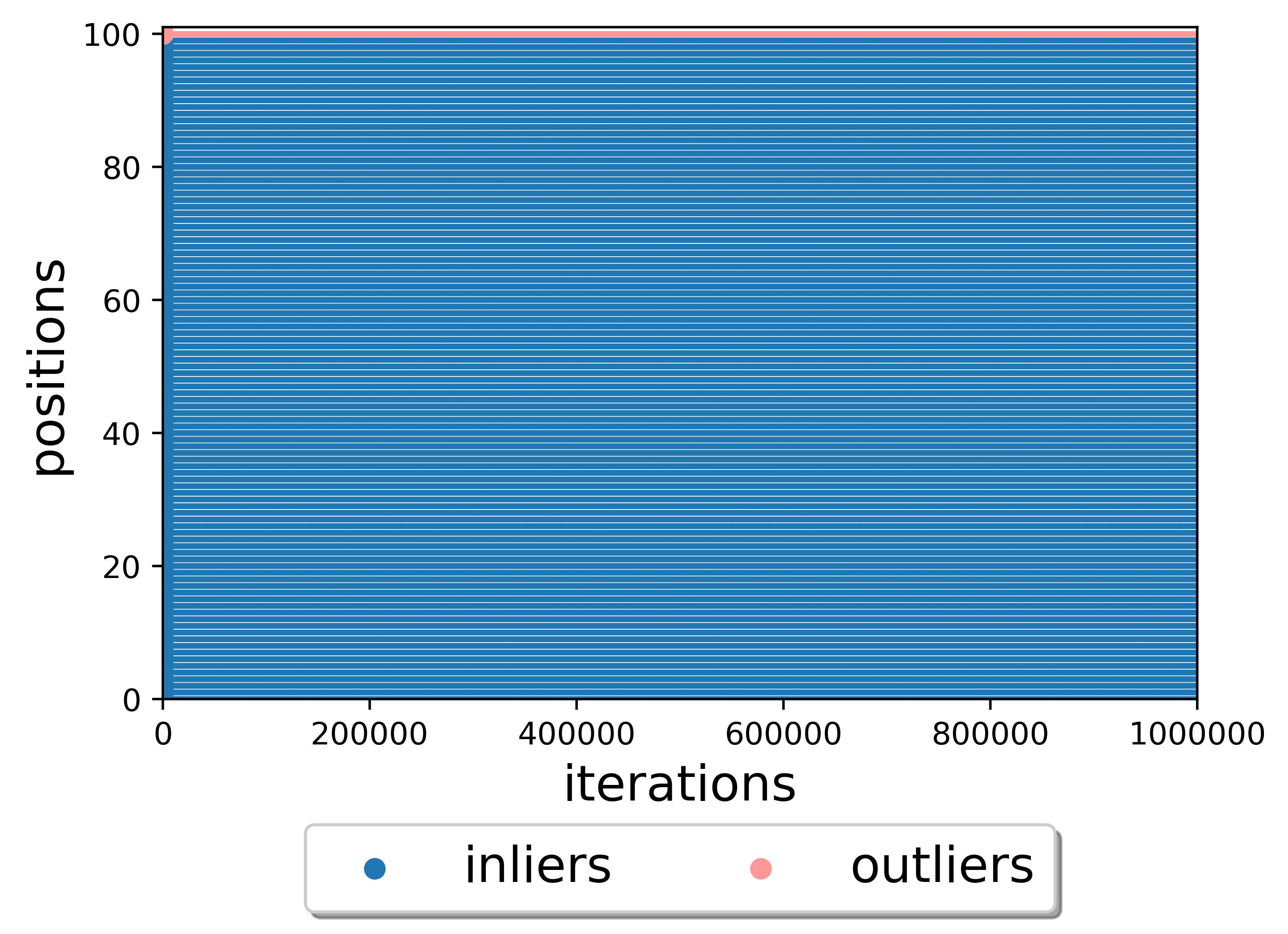}
\end{subfigure}

\begin{subfigure}{.32\textwidth}
    \centering
    \includegraphics[width=1.\linewidth]{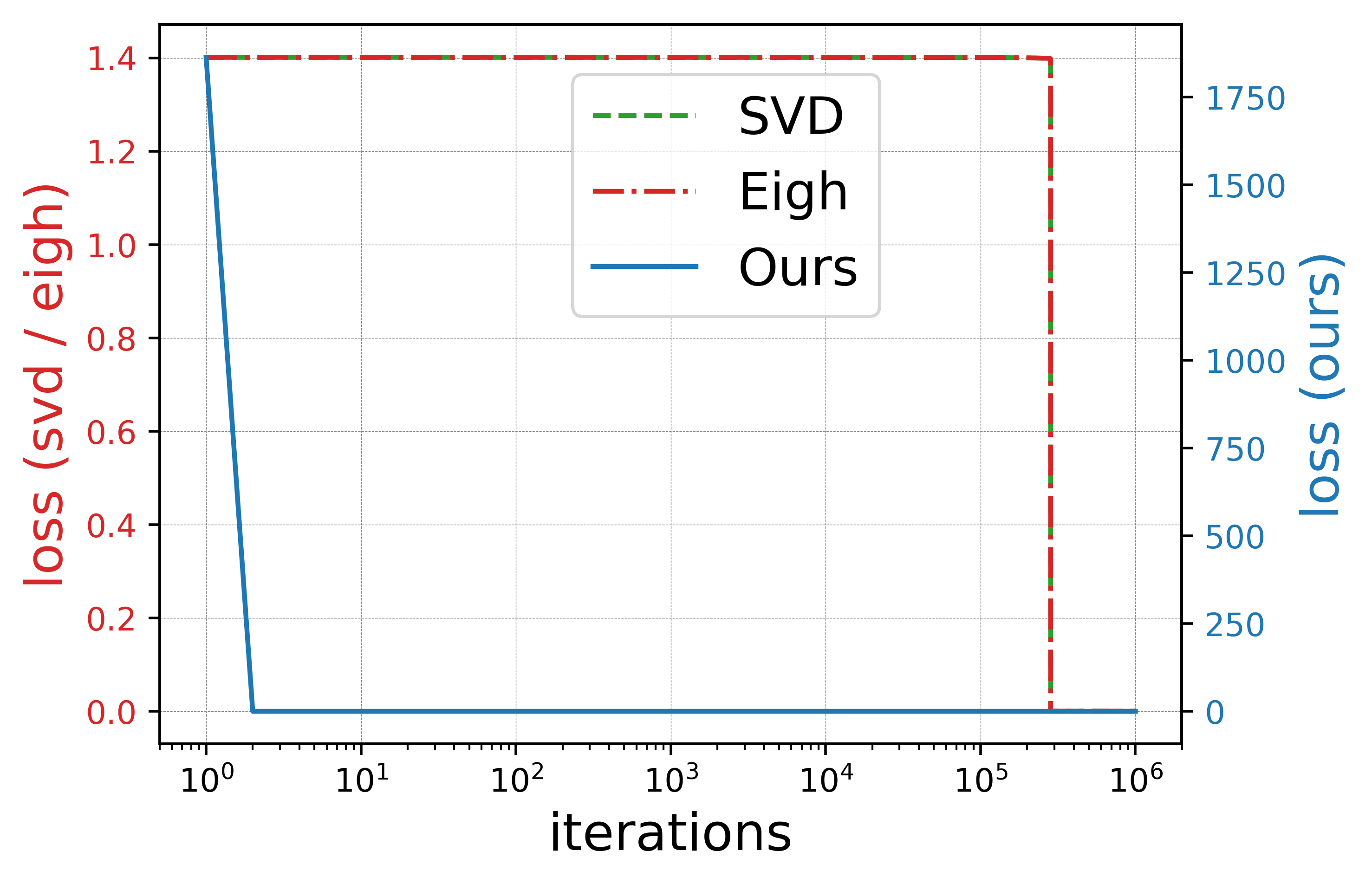}
\end{subfigure}
\begin{subfigure}{.32\textwidth}
	\centering
	\includegraphics[width=1.\linewidth]{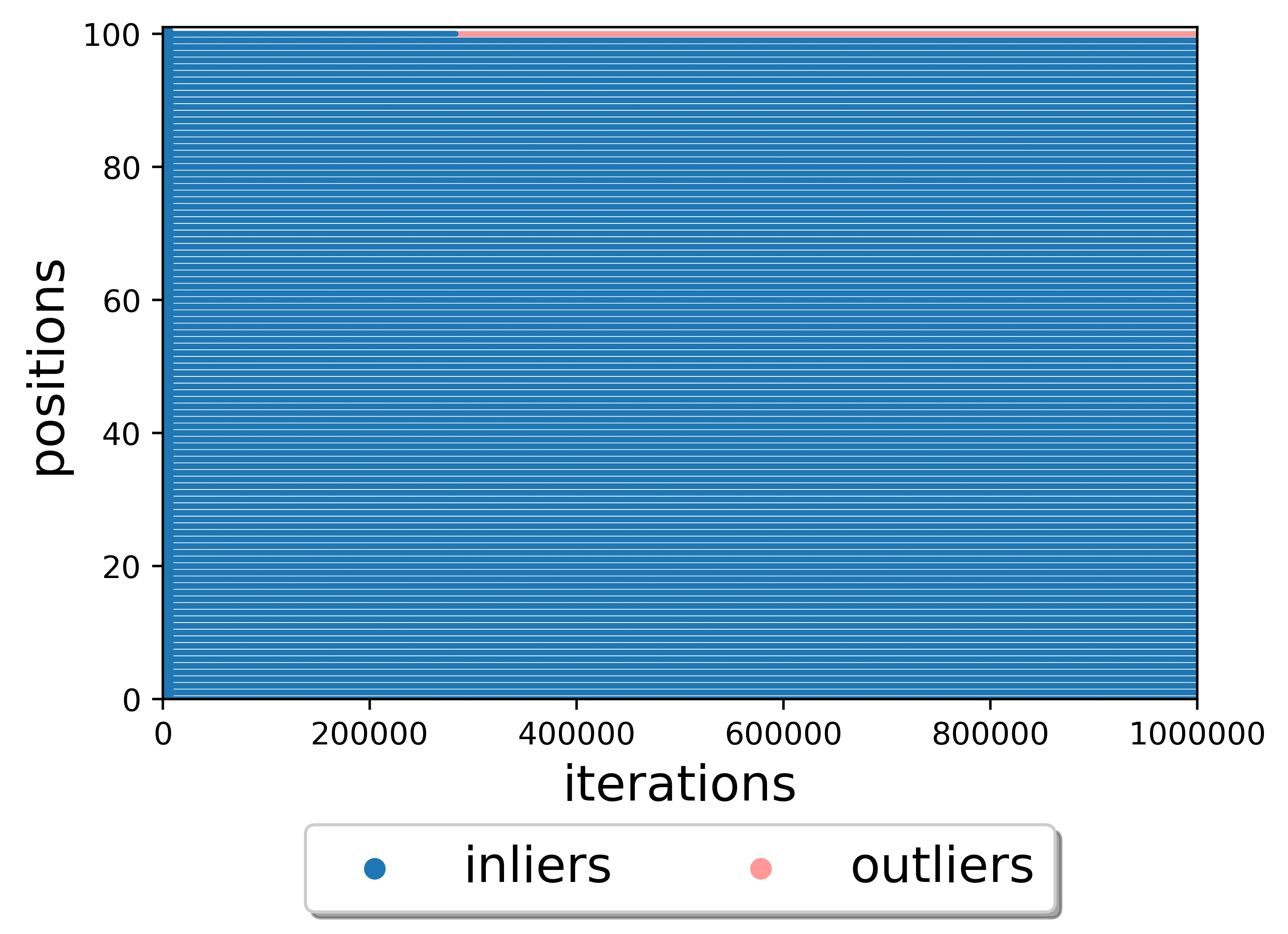}
\end{subfigure}
\begin{subfigure}{.32\textwidth}
	\centering
	\includegraphics[width=1.\linewidth]{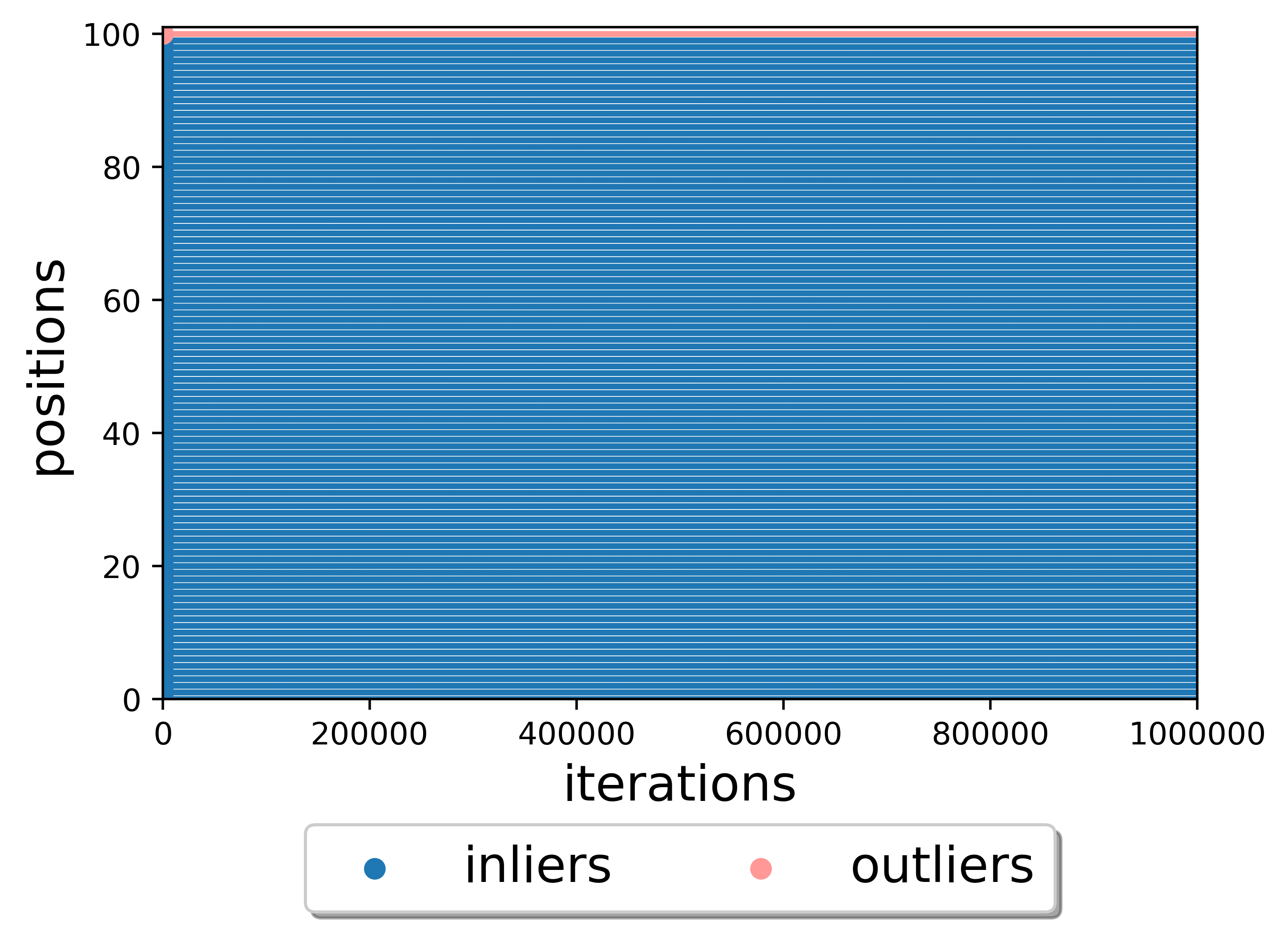}
\end{subfigure}

\begin{subfigure}{.32\textwidth}
    \centering
    \includegraphics[width=1.\linewidth]{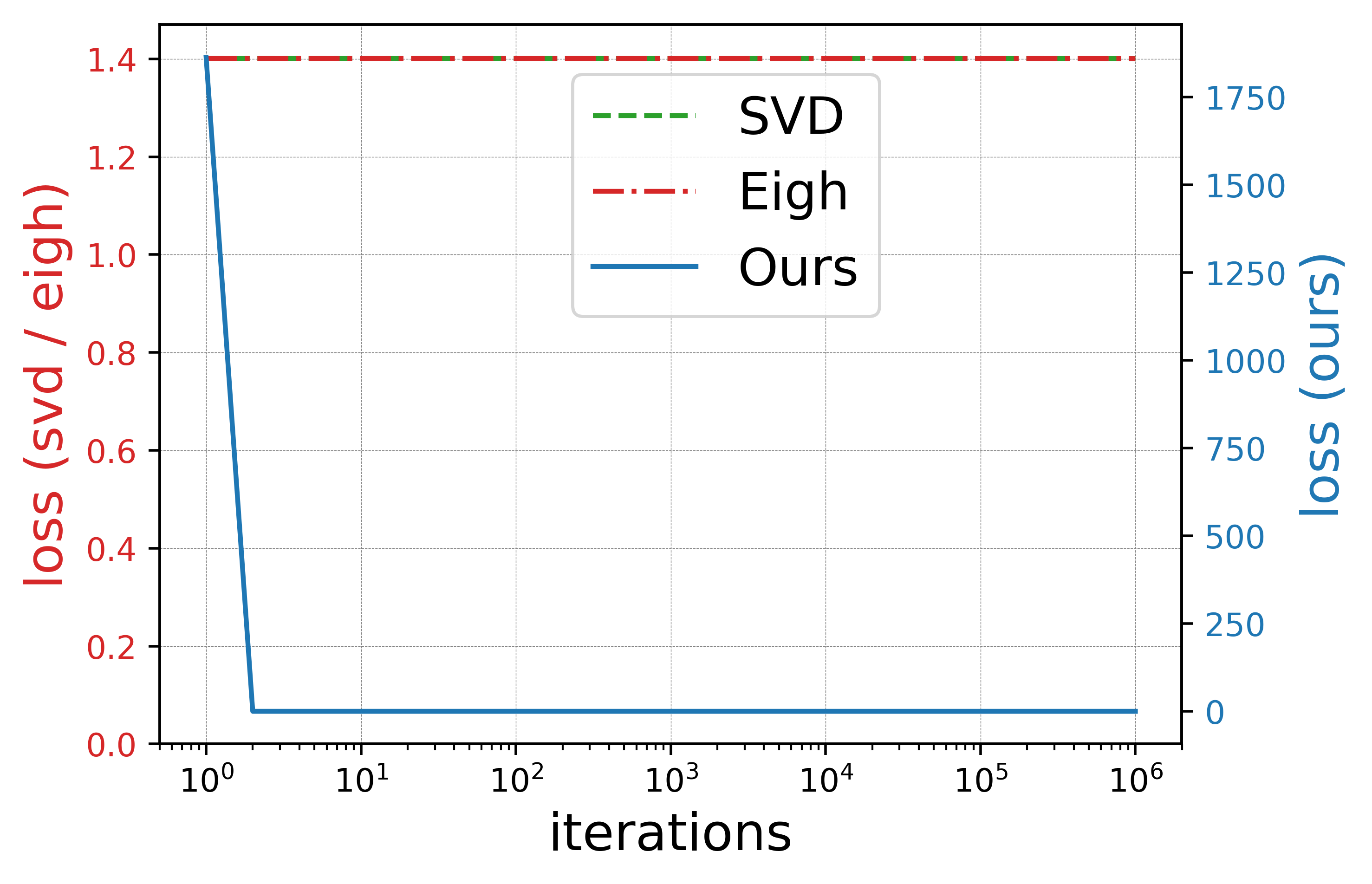}
\end{subfigure}
\begin{subfigure}{.32\textwidth}
	\centering
	\includegraphics[width=1.\linewidth]{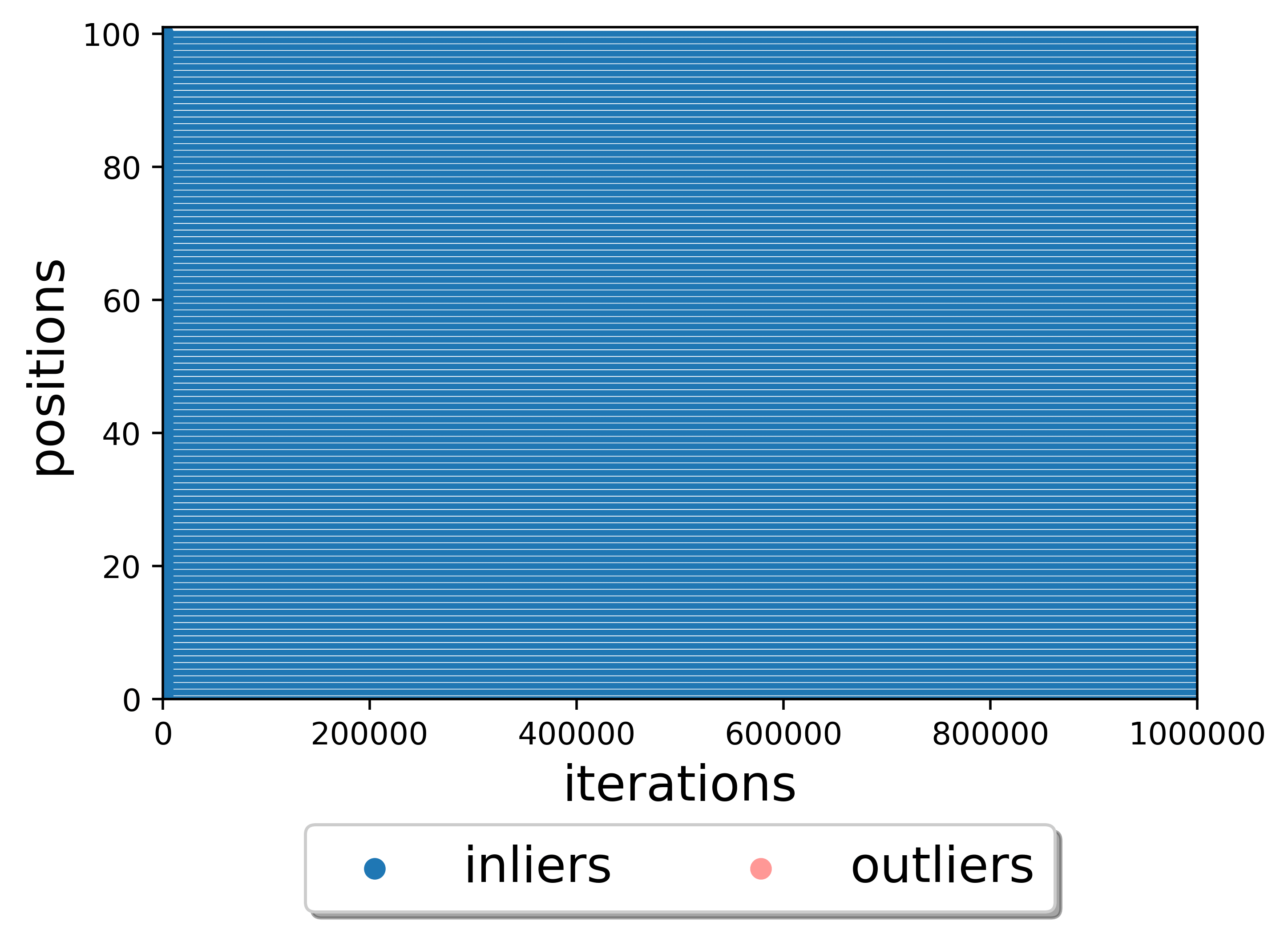}
\end{subfigure}
\begin{subfigure}{.32\textwidth}
	\centering
	\includegraphics[width=1.\linewidth]{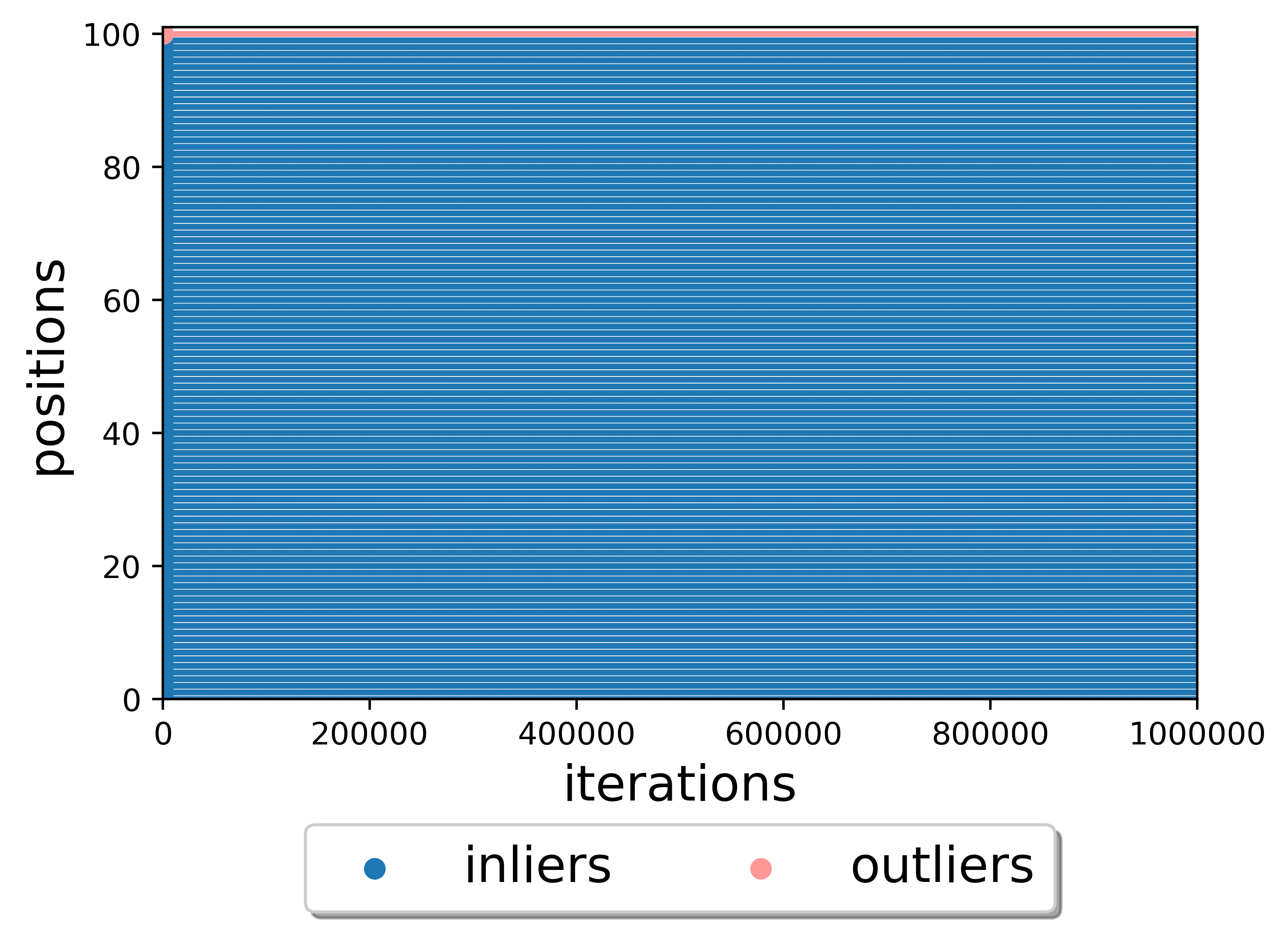}
\end{subfigure}

\begin{subfigure}{.32\textwidth}
    \centering
    \includegraphics[width=1.\linewidth]{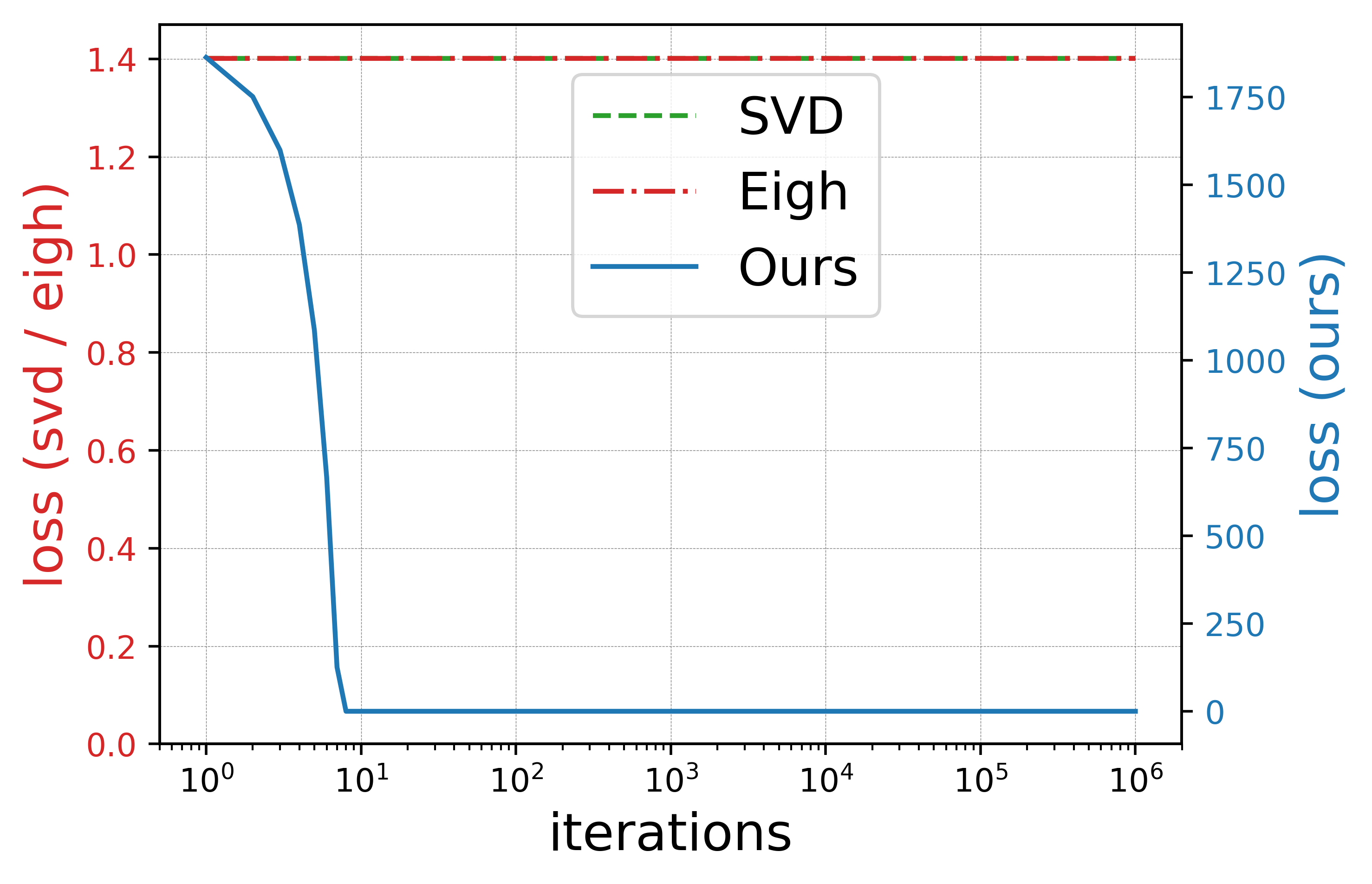}
\end{subfigure}
\begin{subfigure}{.32\textwidth}
	\centering
	\includegraphics[width=1.\linewidth]{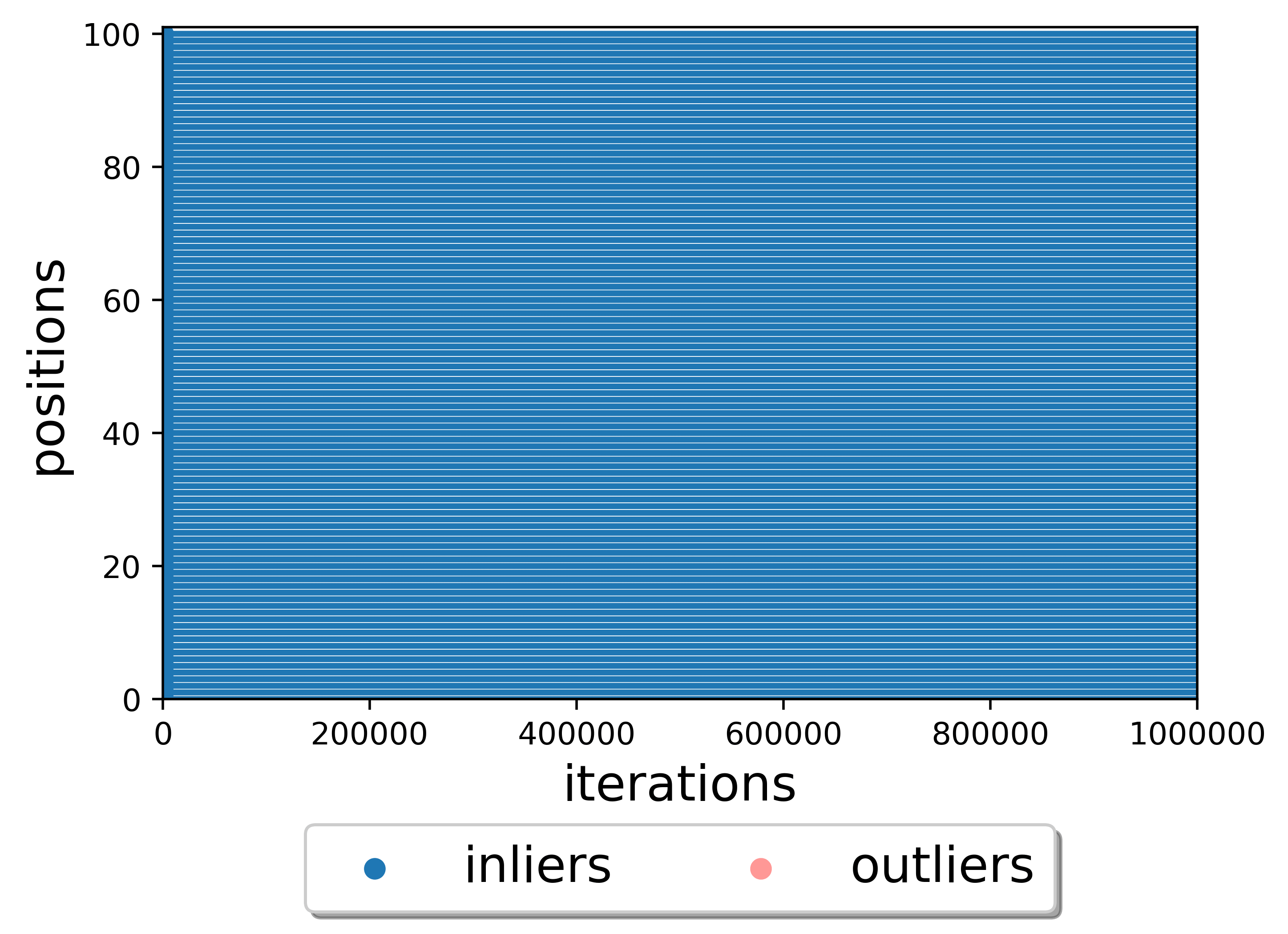}
\end{subfigure}
\begin{subfigure}{.32\textwidth}
	\centering
	\includegraphics[width=1.\linewidth]{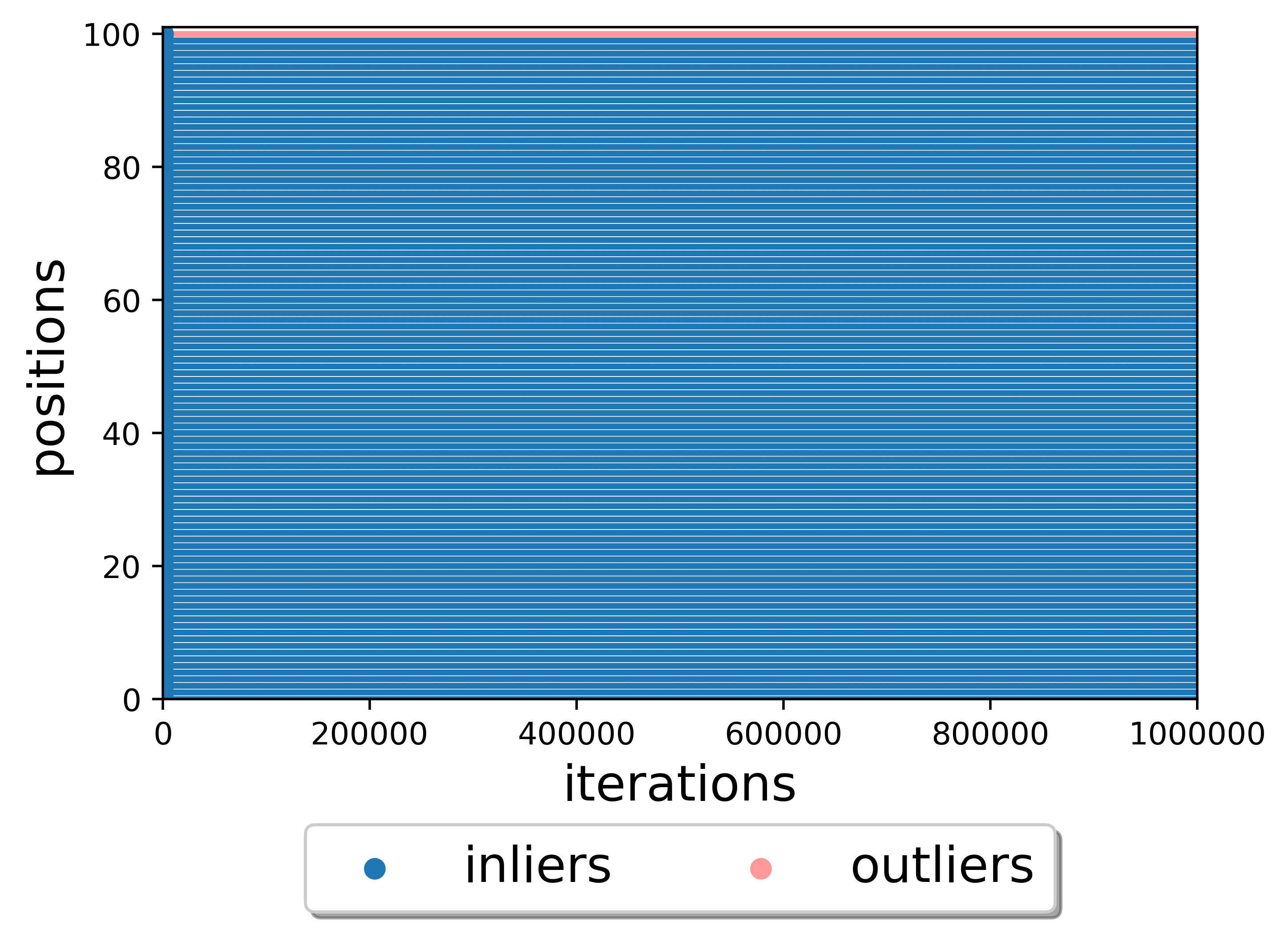}
\end{subfigure}
\begin{subfigure}{.32\textwidth}
    \centering
    \includegraphics[width=1.\linewidth]{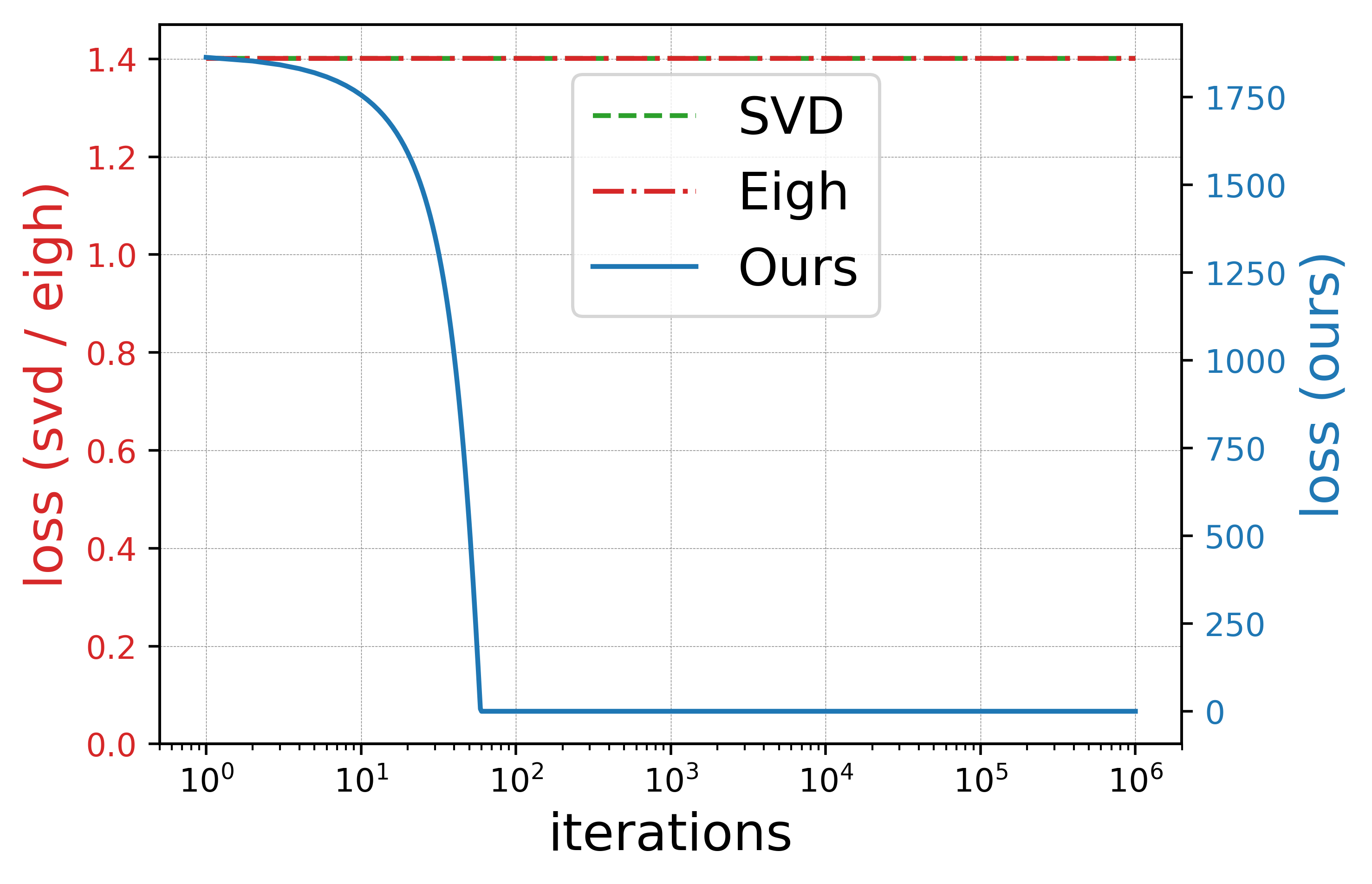}
\end{subfigure}
\begin{subfigure}{.32\textwidth}
	\centering
	\includegraphics[width=1.\linewidth]{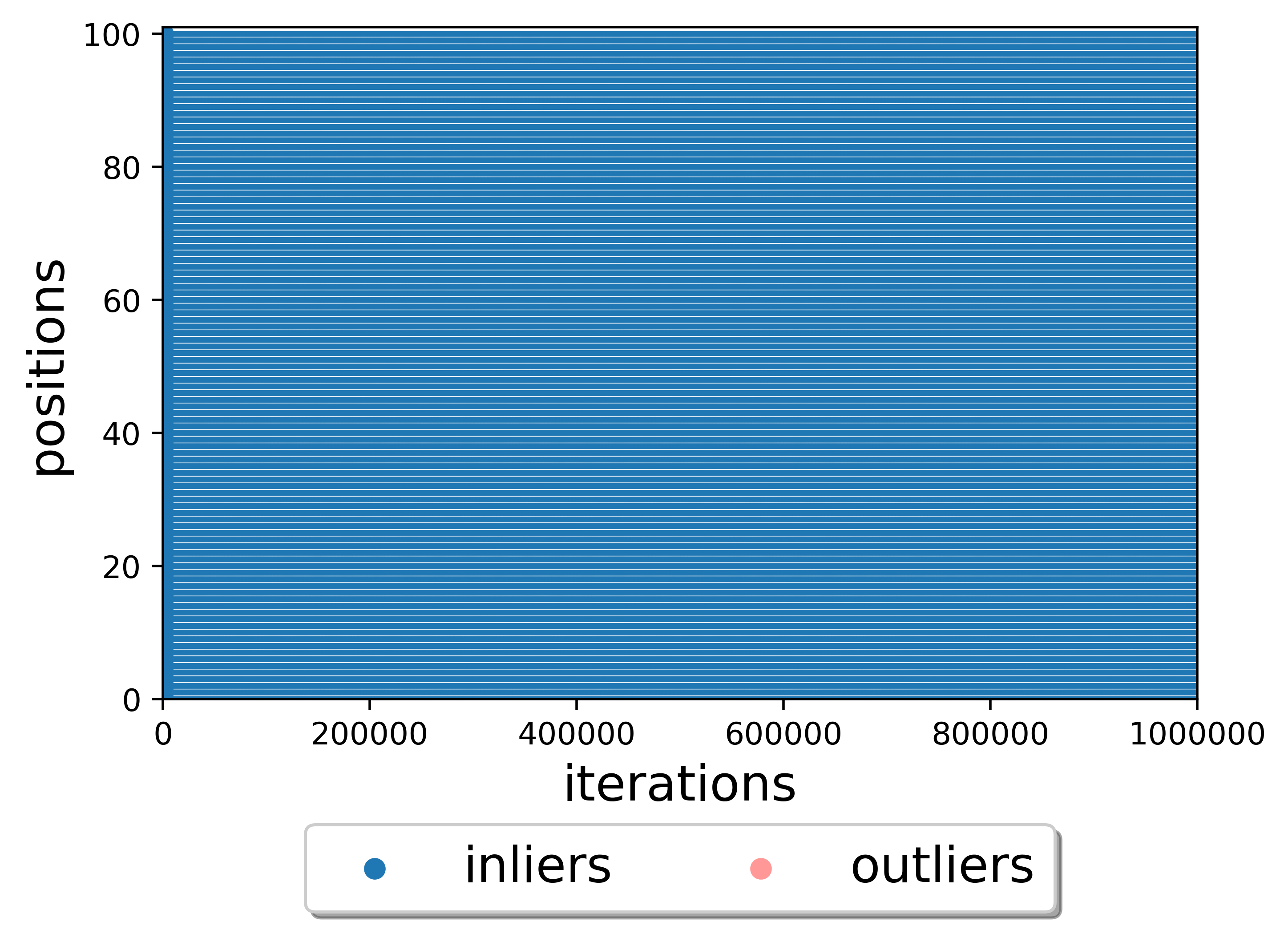}
\end{subfigure}
\begin{subfigure}{.32\textwidth}
	\centering
	\includegraphics[width=1.\linewidth]{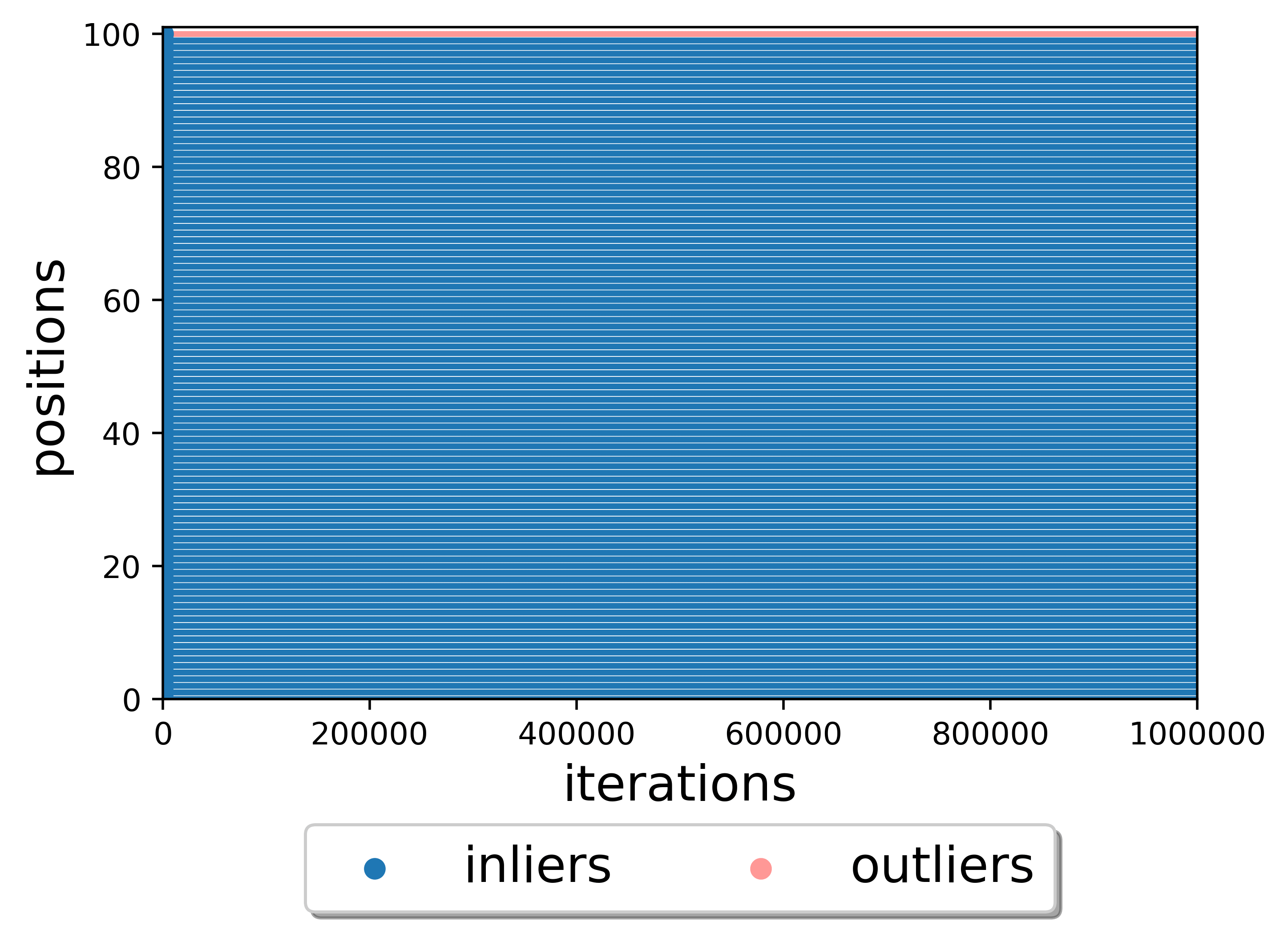}
\end{subfigure}
\caption{
{\bf Loss evolution for the fitting plane problem with GD.} As with Adam in Fig.~\ref{fig:adam}, our approach converges (left) and finds the correct inliers (right). By contrast, SVD/Eigh often do not converge, and when they do, tend to misclassify some points. Note that, for a learning rate of 1, SVD/Eigh returned NaN values during optimization, and we therefore omit this setting here.}
\label{fig:gd}
\end{figure}

%% file: fig/arc5.tex
\begin{figure}[t]
\begin{subfigure}{1.\textwidth}
    \centering
    \includegraphics[width=1.\linewidth]{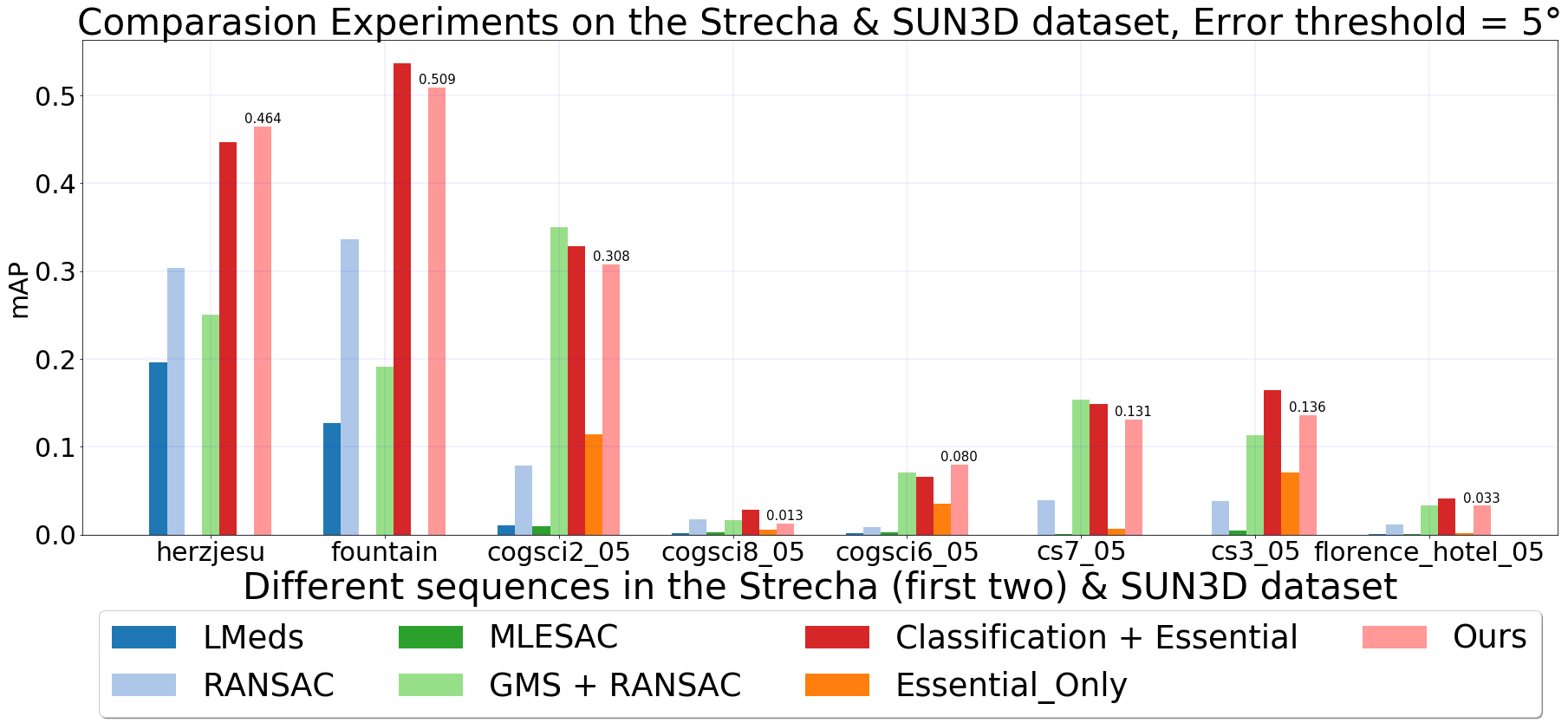}
    \caption{}
    \label{fig:arc_5_8}
\end{subfigure}

\begin{subfigure}{1.\textwidth}
	\centering
	\includegraphics[width=1.\linewidth]{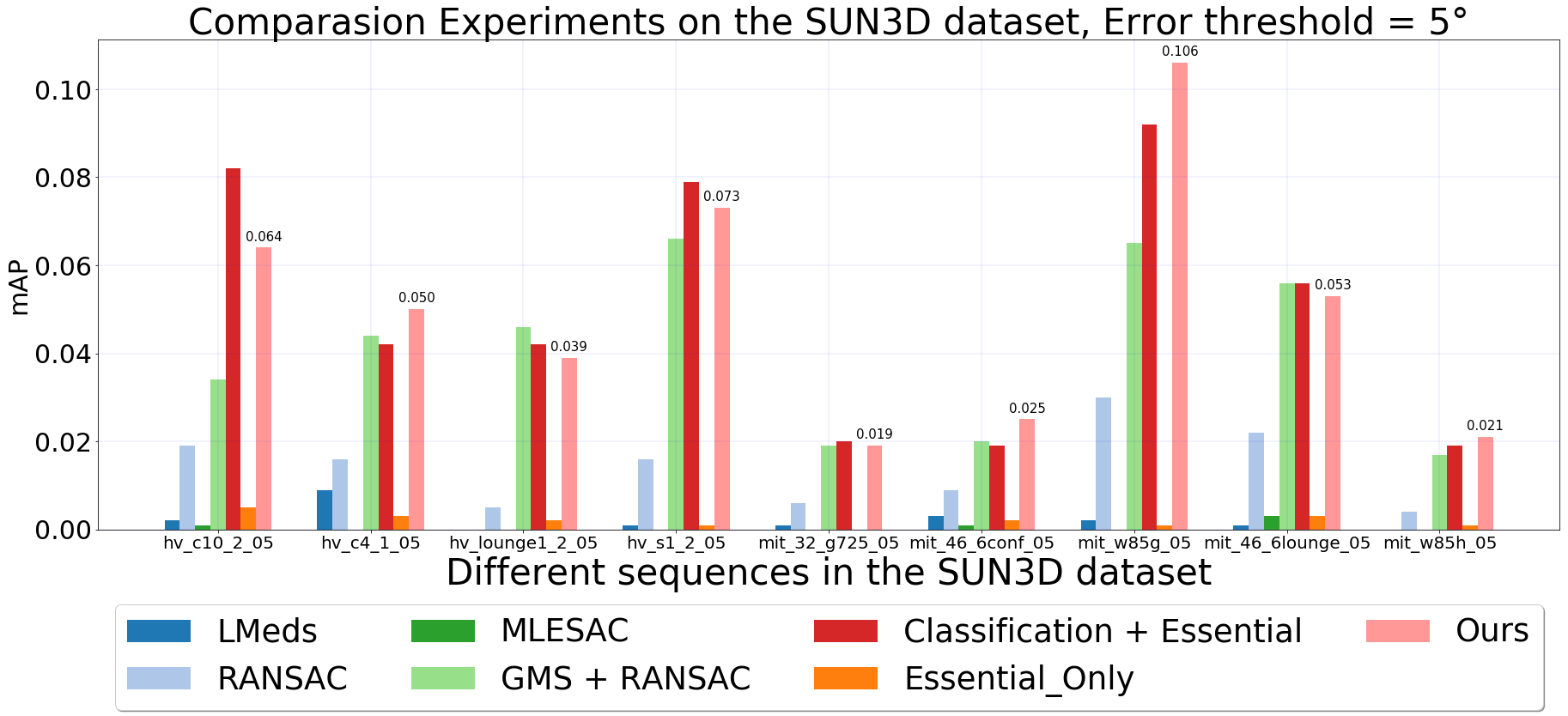}
	\caption{}
	\label{fig:arc_5_9}
\end{subfigure}

\caption{
{\bf Keypoint matching mAP with error threshold 5.} We report the accuracy of the estimated
  relative pose in terms of the mean Average Precision (mAP) measure of~\cite{Yi18a} for the SUN3D dataset and the dataset of~\cite{Strecha08b}.
}
\label{fig:arc5}
\end{figure}

%% file: fig/arc10.tex
\begin{figure}
\begin{subfigure}{1.\textwidth}
    \centering
    \includegraphics[width=1.\linewidth]{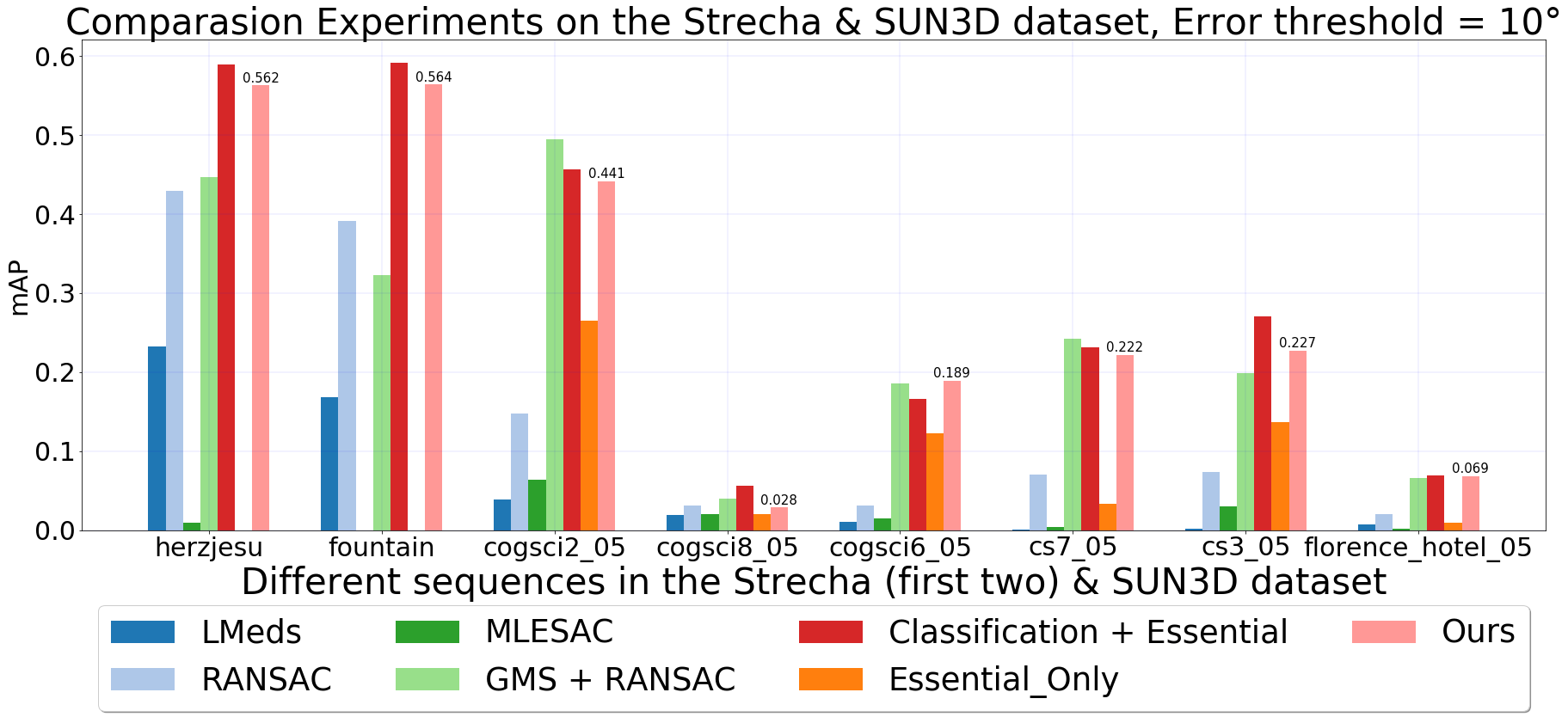}
    \caption{}
    \label{fig:arc_10_8}
\end{subfigure}

\begin{subfigure}{1.\textwidth}
	\centering
	\includegraphics[width=1.\linewidth]{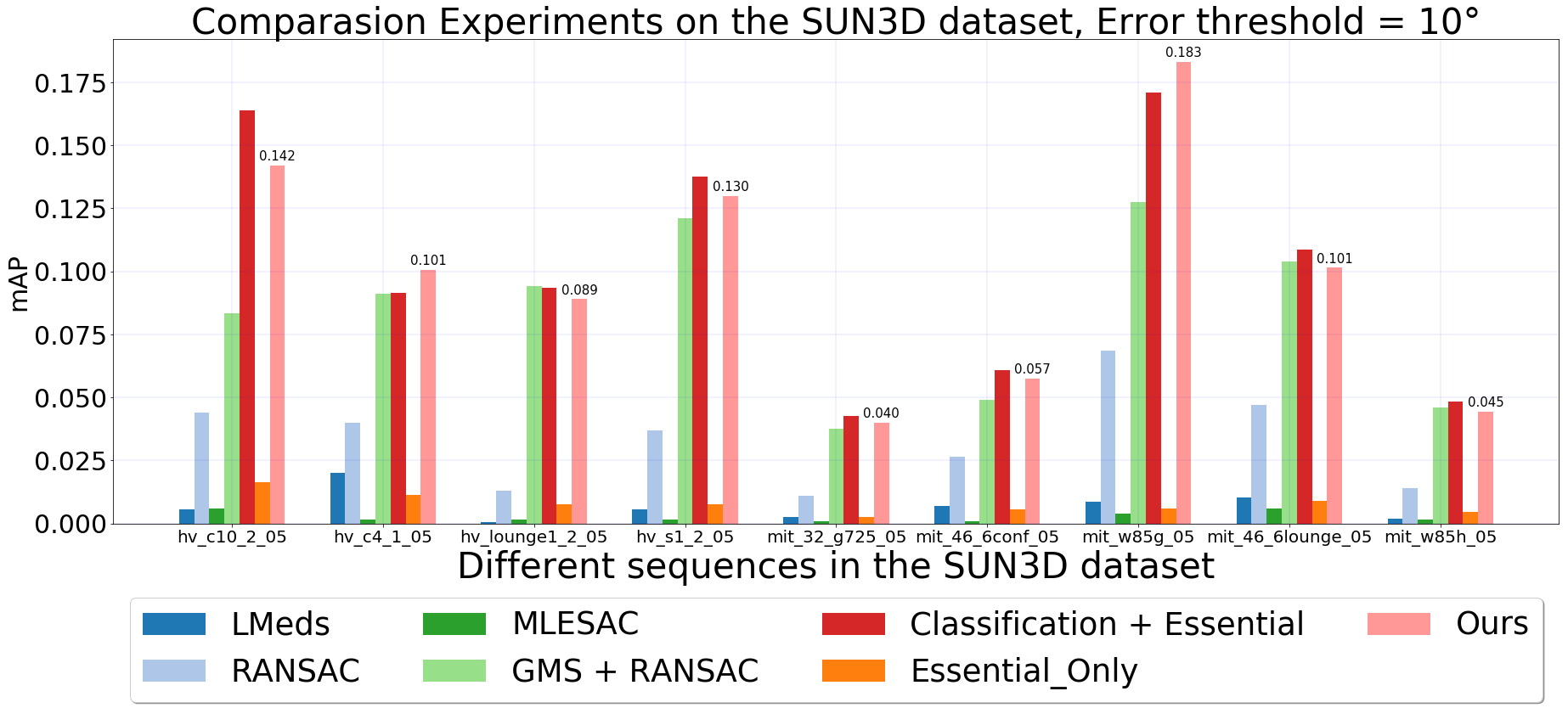}
	\caption{}
	\label{fig:arc_10_9}
\end{subfigure}

 \caption{{\bf Keypoint matching mAP with error threshold 10.} We report the accuracy of the estimated
  relative pose in terms of the mean Average Precision (mAP) measure of~\cite{Yi18a} for the SUN3D dataset and the dataset of~\cite{Strecha08b}.
}
\label{fig:arc10}
\end{figure}

%% file: fig/arc20.tex
\begin{figure}
\begin{subfigure}{1.\textwidth}
    \centering
    \includegraphics[width=1.\linewidth]{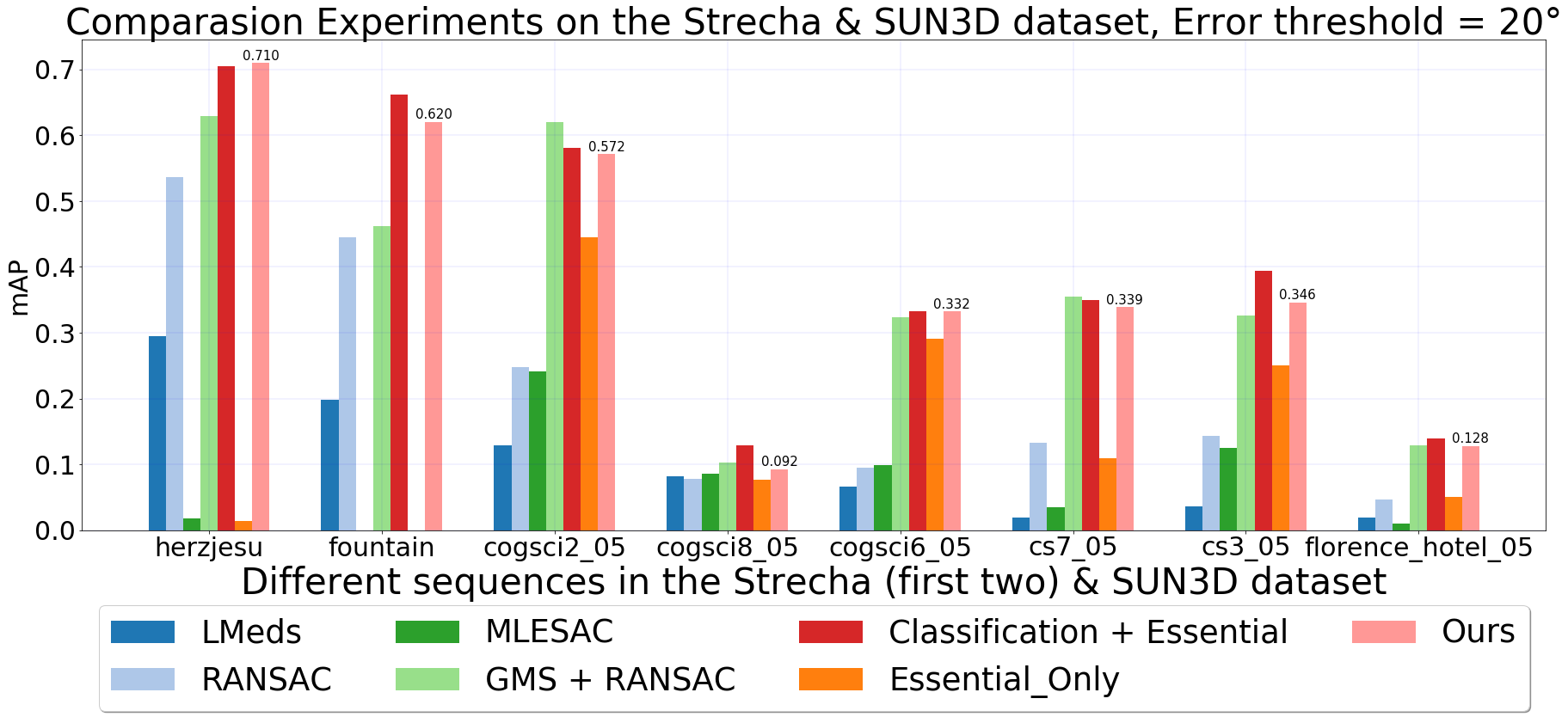}
    \caption{}
    \label{fig:arc_20_8}
\end{subfigure}

\begin{subfigure}{1.\textwidth}
	\centering
	\includegraphics[width=1.\linewidth]{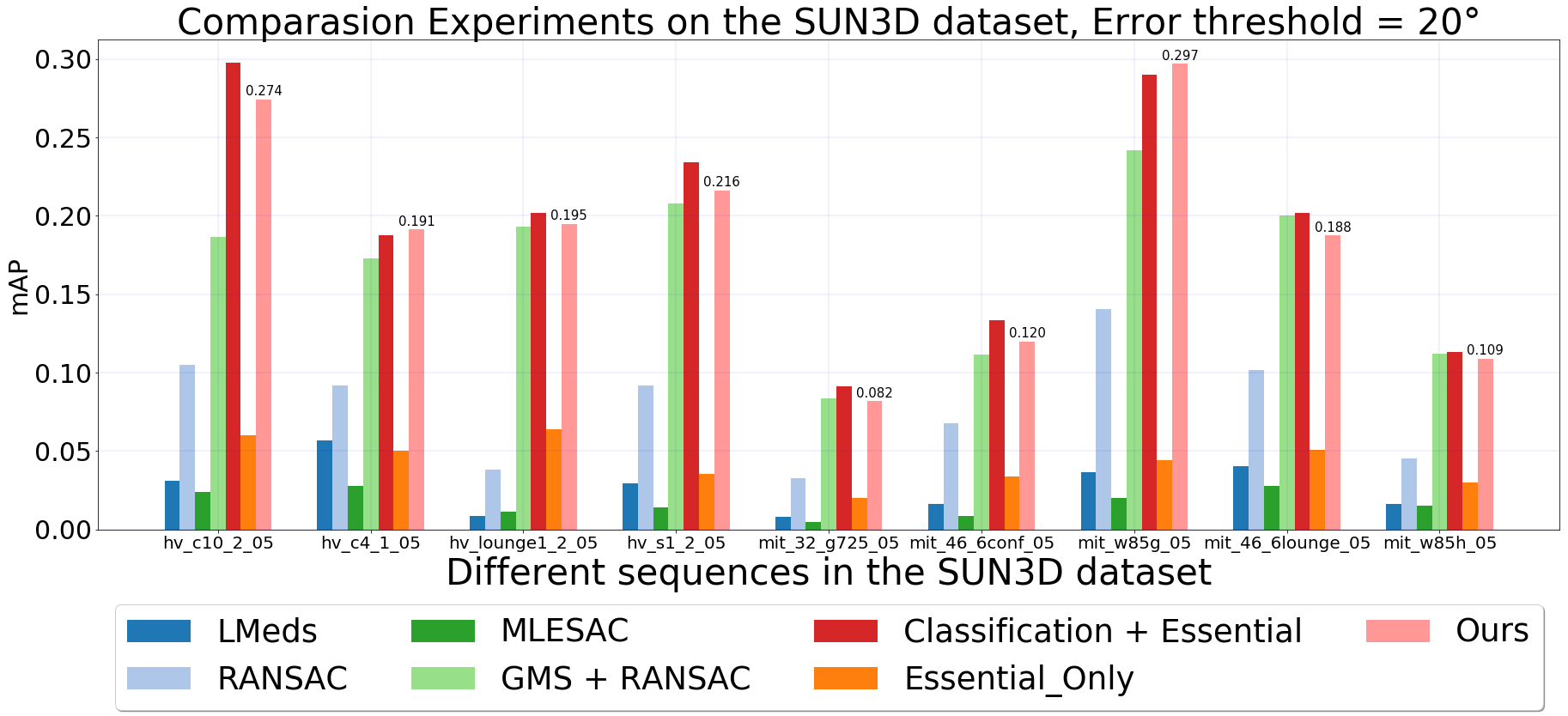}
	\caption{}
	\label{fig:arc_20_9}
\end{subfigure}

\caption{
{\bf Keypoint matching mAP with error threshold 20.} We report the accuracy of the estimated
  relative pose in terms of the mean Average Precision (mAP) measure of~\cite{Yi18a} for the SUN3D dataset and the dataset of~\cite{Strecha08b}.
}
\label{fig:arc20}
\end{figure}

%% file: fig/table.tex
\begin{table}[t]
\caption{{\bf Comparison of the rotation error of our approach with those of the
    baselines.} A $^*$ indicates that RANSAC was used as a postprocessing
  step. Best results are shown in bold.}
\label{tab1}
\begin{center}
\begin{tabular}{l c c c c c c c c c c}
\toprule
Methods             & Ours & REPPnP & EPnP & RPnP & OPnP & PPnP & DLT & EPnP$^*$ & P3P$^*$ \\
\midrule
Reichstag      		& {\bf 0.1319} & 36.0852 & 36.5411 & 12.6649 & 3.7012 & 18.6296 & 35.1560 & 3.1161 & 3.1161\\
Florence         	& {\bf 0.0416} & 32.6205 & 34.6684 & 24.2201 & 1.5771 & 5.0694 & 38.3336 & 2.4004 & 2.4004\\
Prague     			& {\bf 0.0274} & 39.6677 & 39.4239 & 4.6647 & 2.7118 & 16.4462 & 35.1211 & 2.5443 & 2.5443 \\
Notre-dame			& {\bf 0.0293} & 36.9159 & 32.7849 & 16.2304 &  2.8228 & 8.7266 & 32.8611 & 4.2309 & 4.263 \\
\bottomrule
\end{tabular}
\end{center}
\end{table}
\begin{table}[t]
\caption{{\bf Comparison of the translation error of our approach with those of
    the baselines.} A $^*$ indicates that RANSAC was used as a postprocessing
  step. Best results are shown in bold.}
\label{tab2}
\begin{center}
\begin{tabular}{l c c c c c c c c c c}
\toprule
Methods             & Ours & REPPnP & EPnP & RPnP & OPnP & PPnP & DLT & EPnP$^*$ & P3P$^*$\\
\midrule
Reichstag      		& {\bf 0.0110} & 0.2477 & 1.4313 & 0.3885 & 0.2346 & 1.2724 & 0.3687 & 0.1821 & 0.1821 \\
Florence         	& {\bf 0.0135} & 1.9399 & 1.6896 & 1.7971 & 0.8719 & 0.4659 & 1.9069 & 0.7640 & 0.7640 \\
Prague     			& {\bf 0.0069} & 0.6156 & 1.4631 & 1.4440 & 0.8715 & 0.8278 & 1.6928 & 0.8045 & 0.8045 \\
Notre-dame			& {\bf 0.0046} & 1.8663 & 1.7654 & 1.7910 & 0.8382 & 1.0669 & 1.7918 & 1.0605 & 1.0615 \\
\bottomrule
\end{tabular}
\end{center}
\end{table}

%% file: fig/pnp_supply.tex
\begin{figure}[t]
\centering
\begin{subfigure}{.24\textwidth}
    \centering
    \includegraphics[width=1.\linewidth,trim={0 0 0 2.5cm},clip]{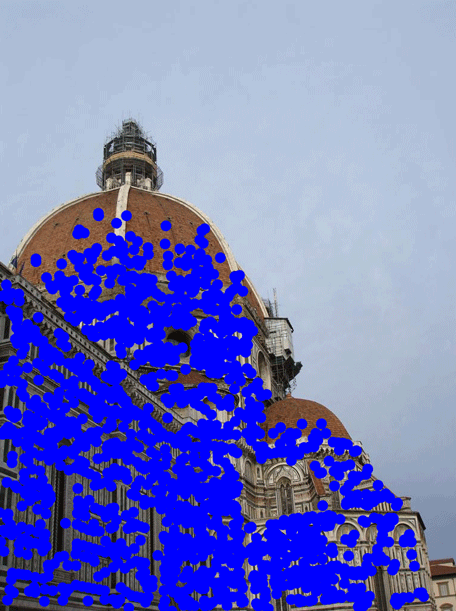}
\end{subfigure}
\begin{subfigure}{.24\textwidth}
    \centering
    \includegraphics[width=1.\linewidth]{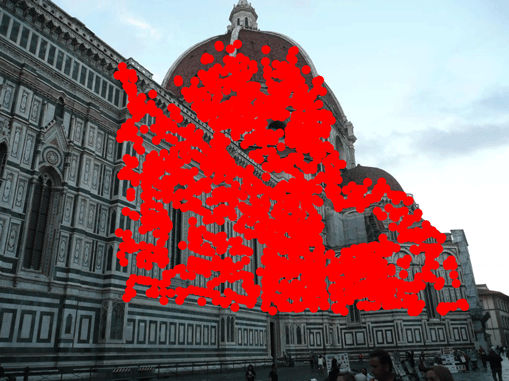}
\end{subfigure}
\begin{subfigure}{.24\textwidth}
    \centering
    \includegraphics[width=1.\linewidth]{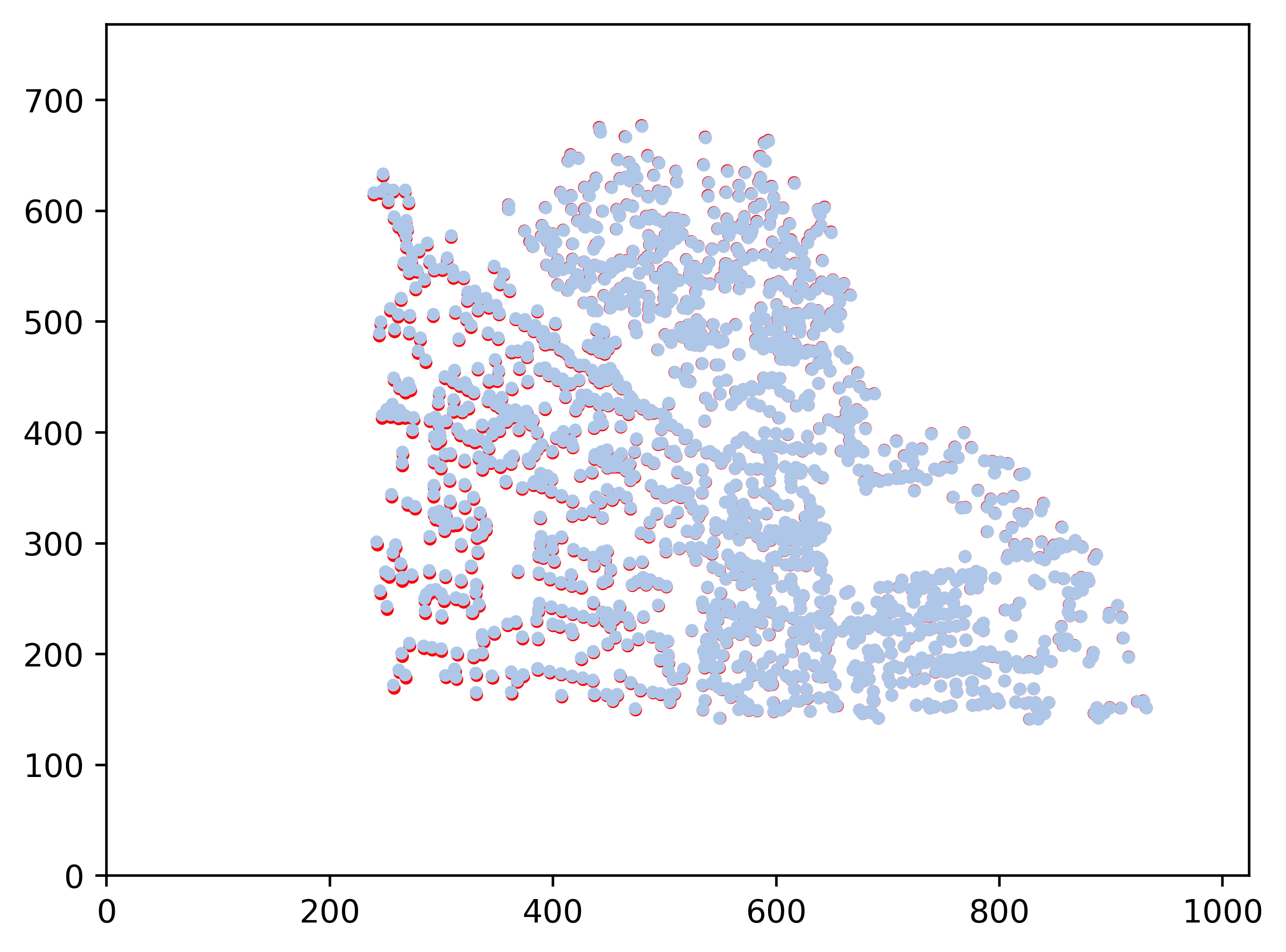}
\end{subfigure}
\begin{subfigure}{.24\textwidth}
    \centering
    \includegraphics[width=1.\linewidth]{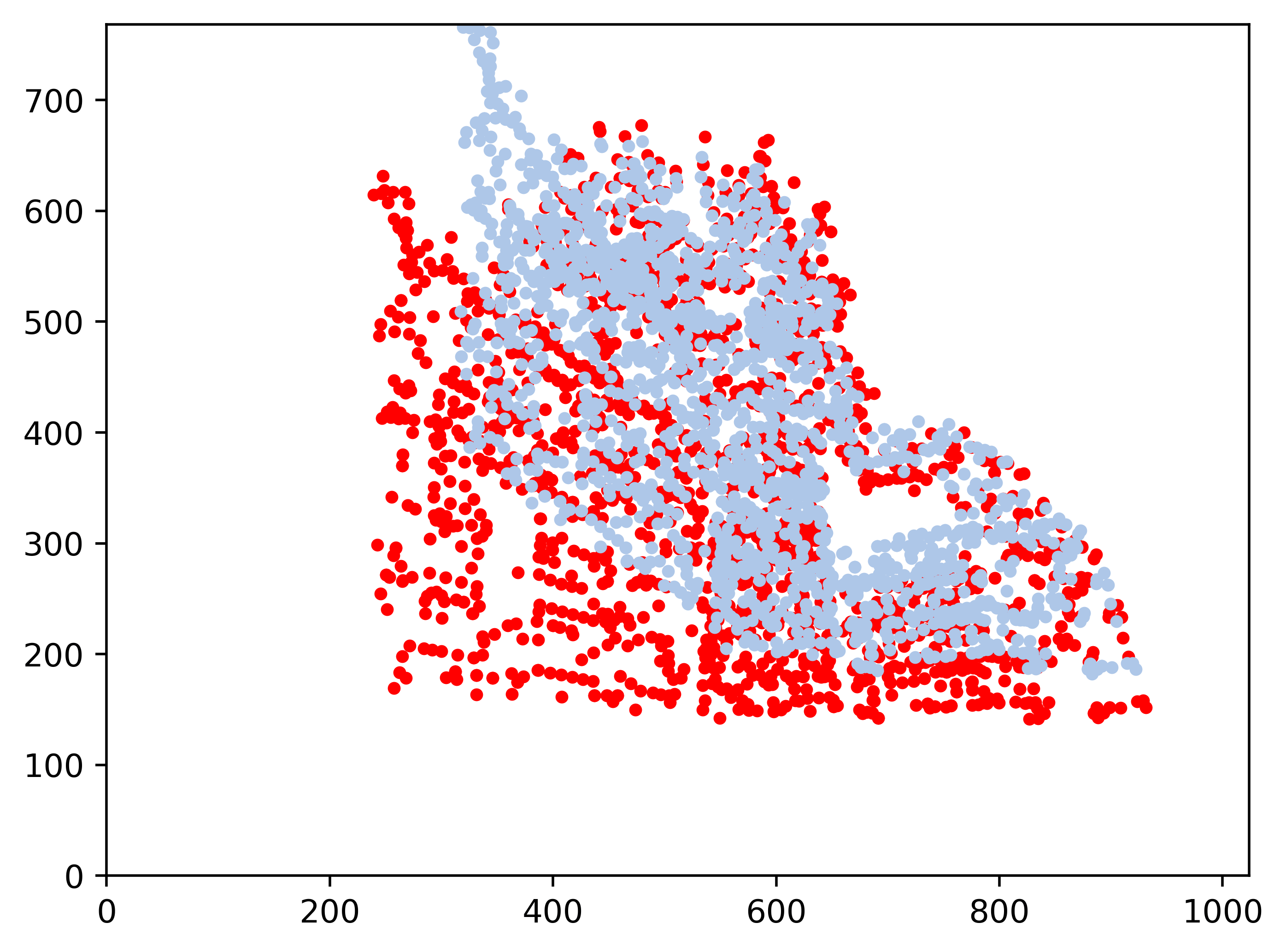}
\end{subfigure}

\begin{subfigure}{.24\textwidth}
    \centering
    \includegraphics[width=1.\linewidth]{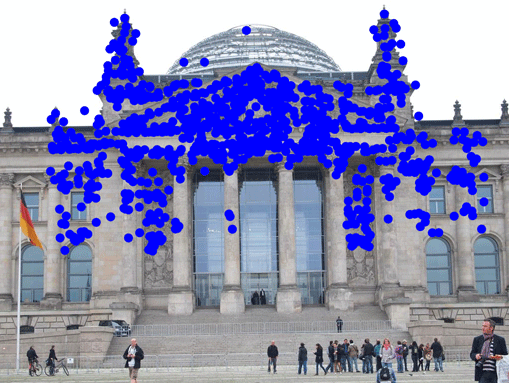}
\end{subfigure}
\begin{subfigure}{.24\textwidth}
    \centering
    \includegraphics[width=1.\linewidth]{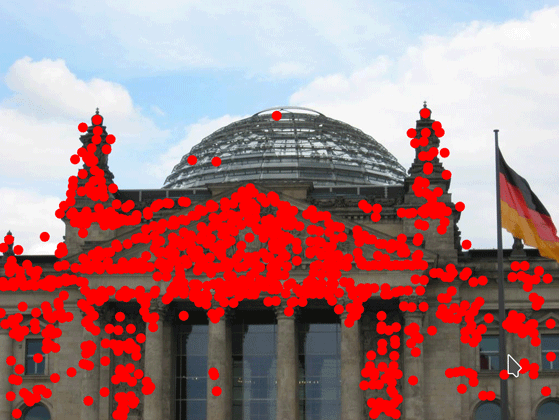}
\end{subfigure}
\begin{subfigure}{.24\textwidth}
    \centering
    \includegraphics[width=1.\linewidth]{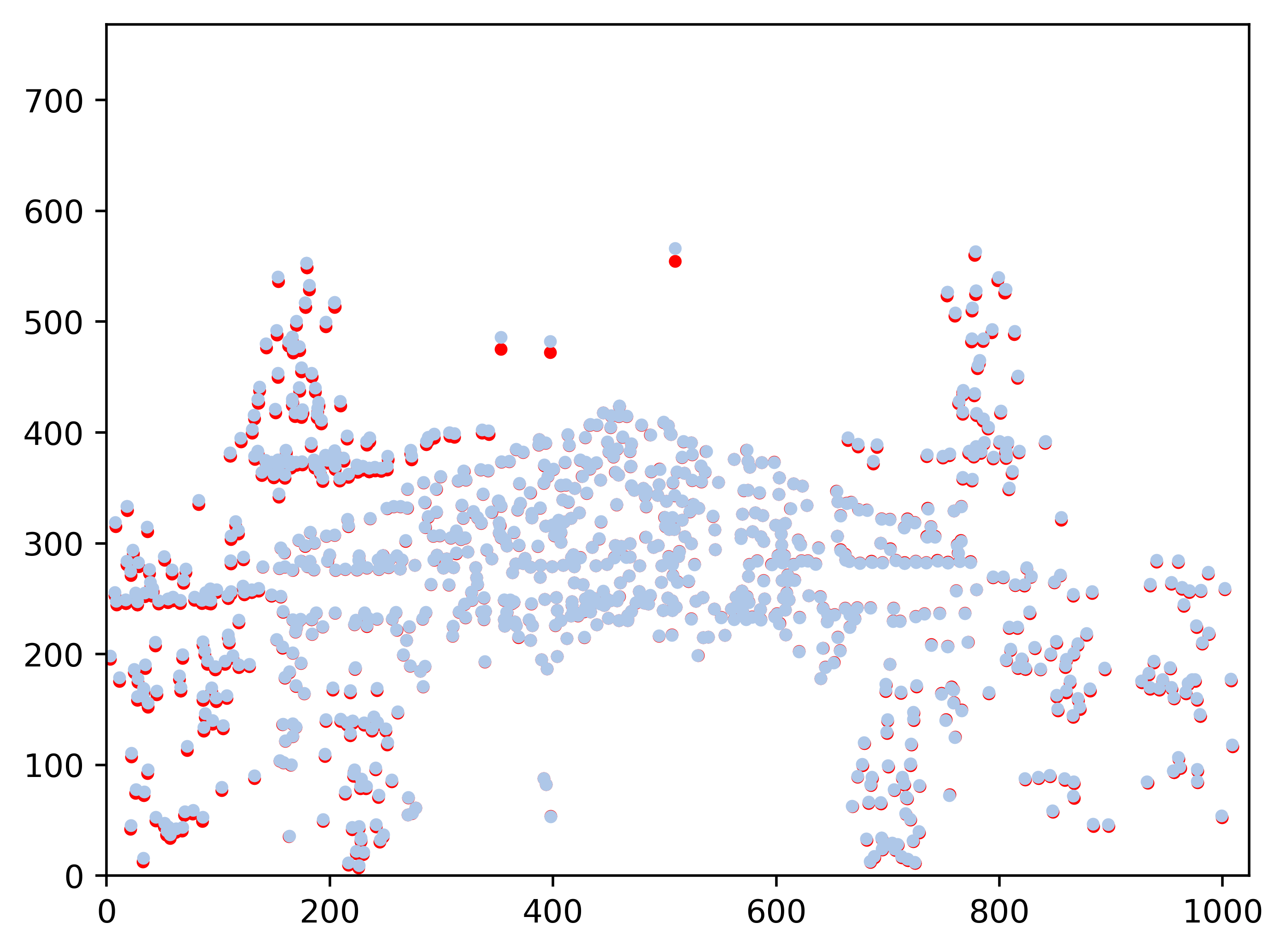}
\end{subfigure}
\begin{subfigure}{.24\textwidth}
    \centering
    \includegraphics[width=1.\linewidth]{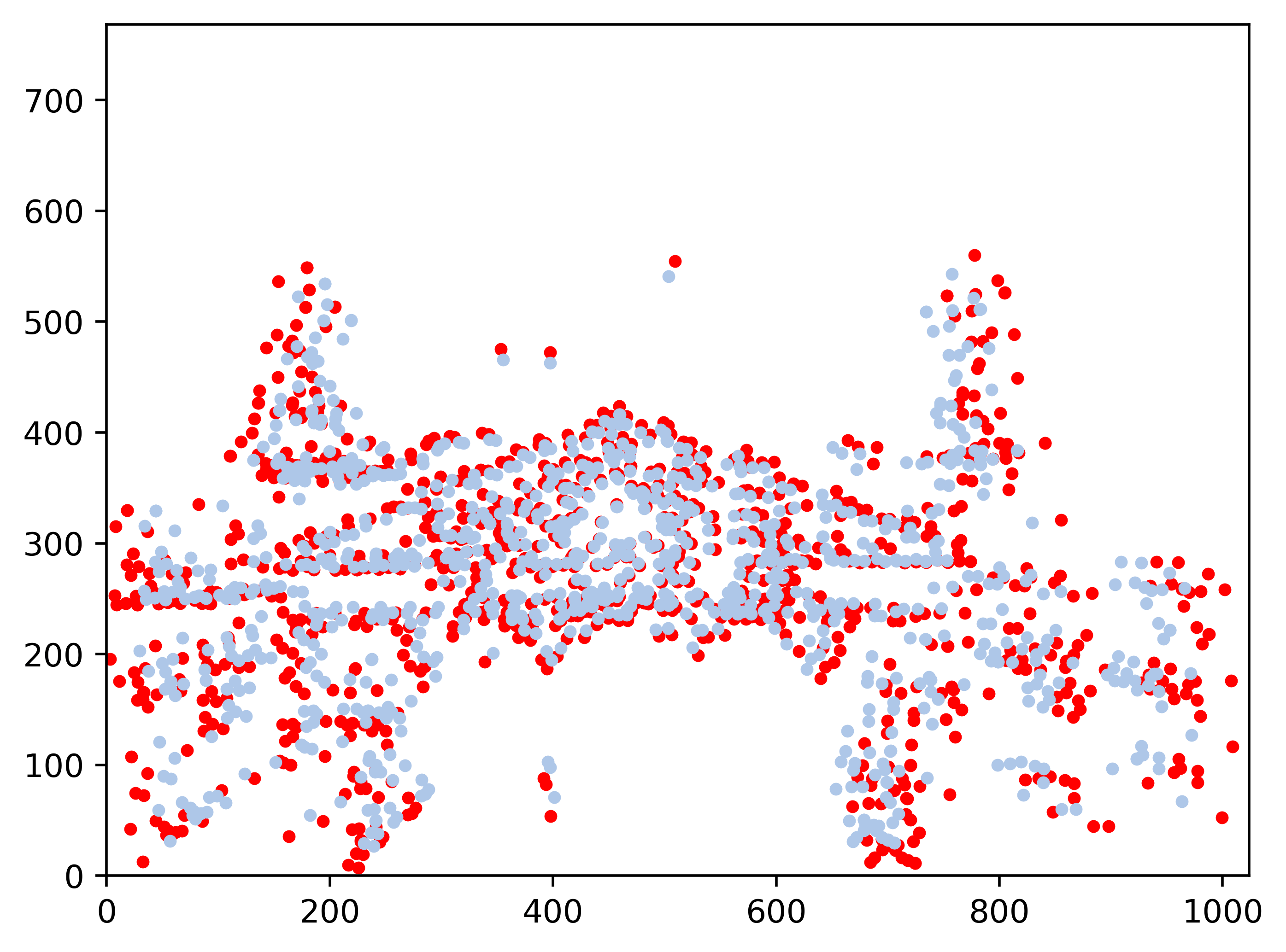}
\end{subfigure}

\begin{subfigure}{.24\textwidth}
    \centering
    \includegraphics[width=1.\linewidth]{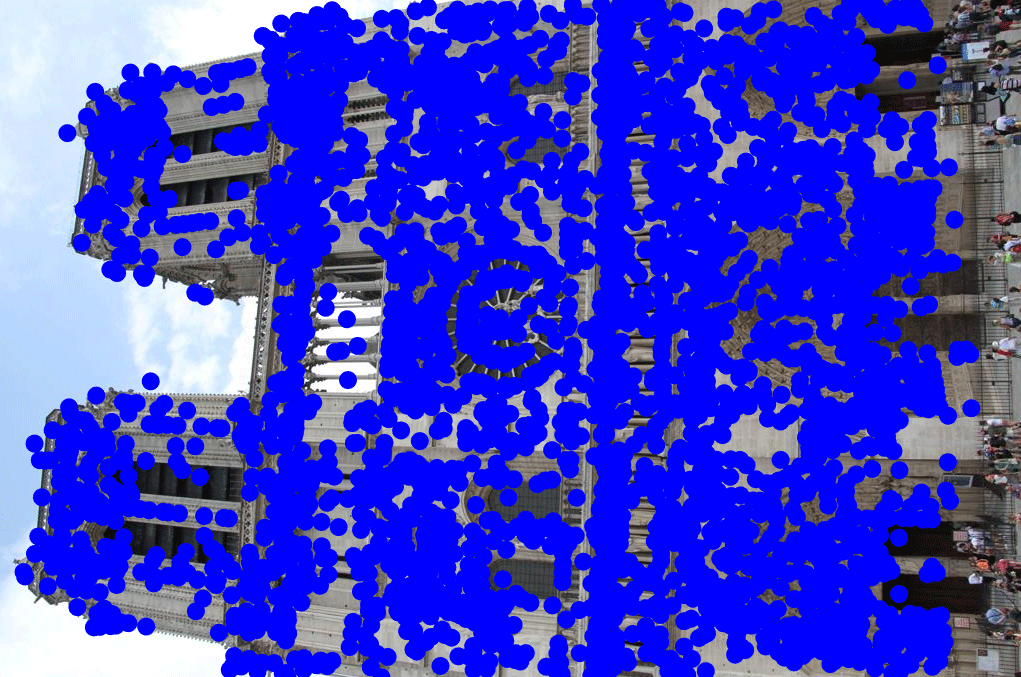}
\end{subfigure}
\begin{subfigure}{.24\textwidth}
    \centering
    \includegraphics[width=1.\linewidth]{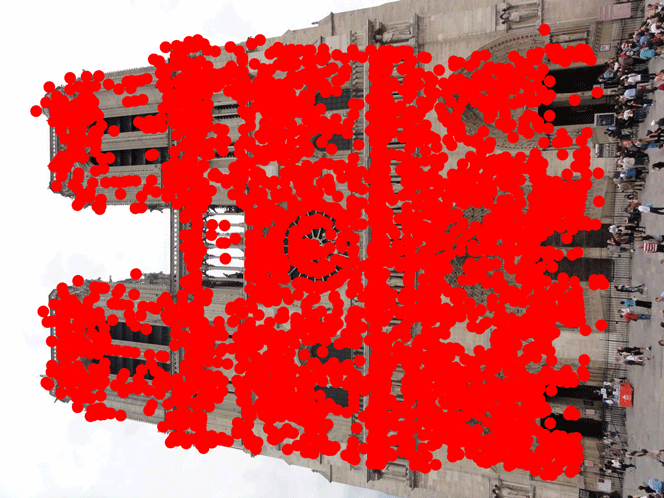}
\end{subfigure}
\begin{subfigure}{.24\textwidth}
    \centering
    \includegraphics[width=1.\linewidth]{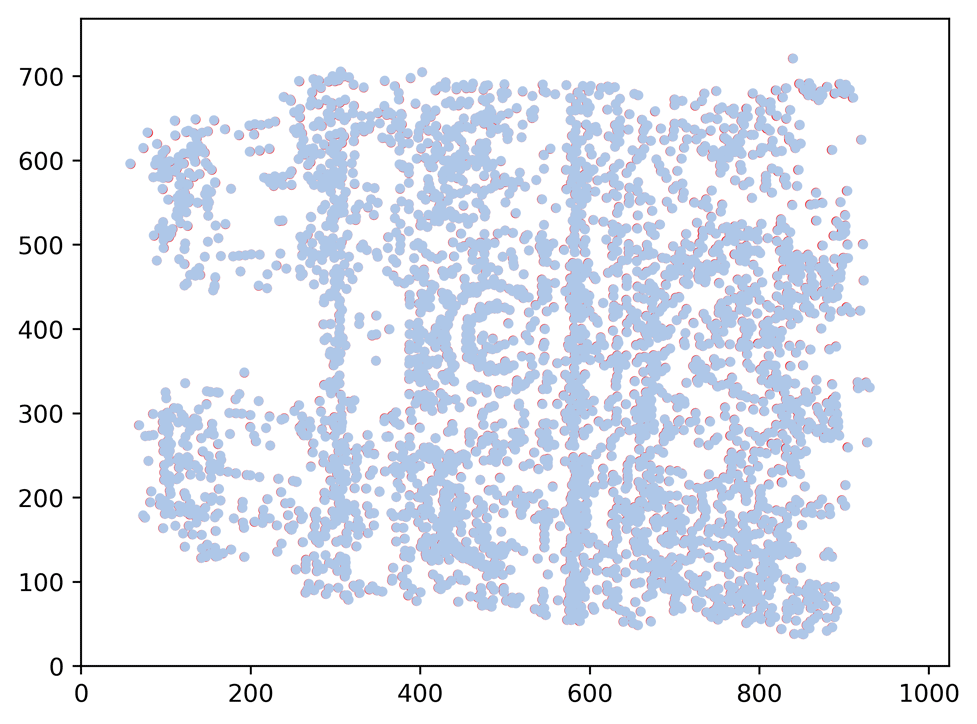}
\end{subfigure}
\begin{subfigure}{.24\textwidth}
	\centering
	\includegraphics[width=1.\linewidth]{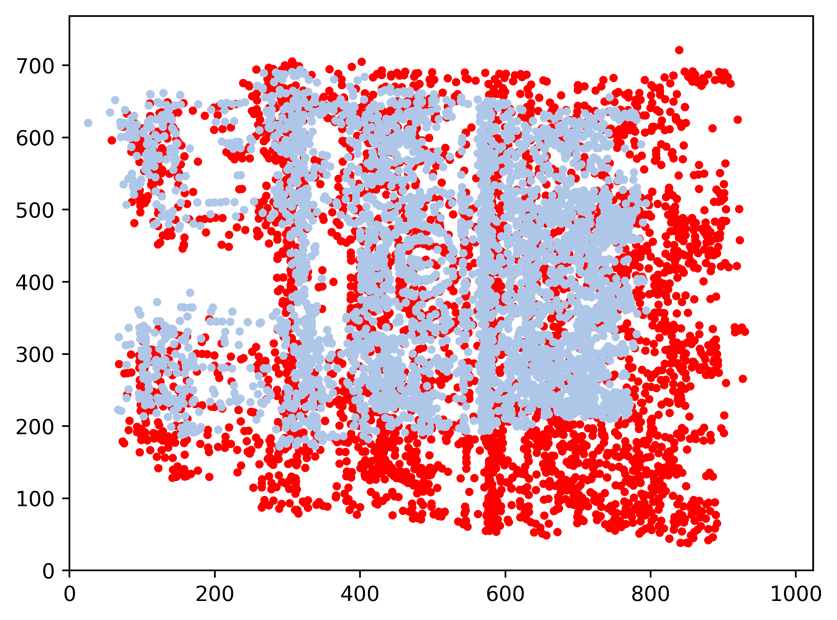}
\end{subfigure}

\begin{subfigure}{.24\textwidth}
    \centering
    \includegraphics[width=1.\linewidth]{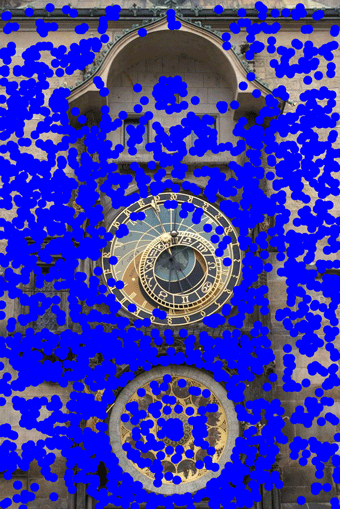}
\end{subfigure}
\begin{subfigure}{.24\textwidth}
    \centering
    \includegraphics[width=1.\linewidth]{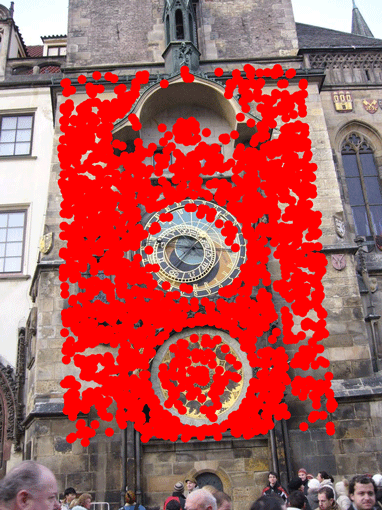}
\end{subfigure}
\begin{subfigure}{.24\textwidth}
    \centering
    \includegraphics[width=1.\linewidth, height=4cm]{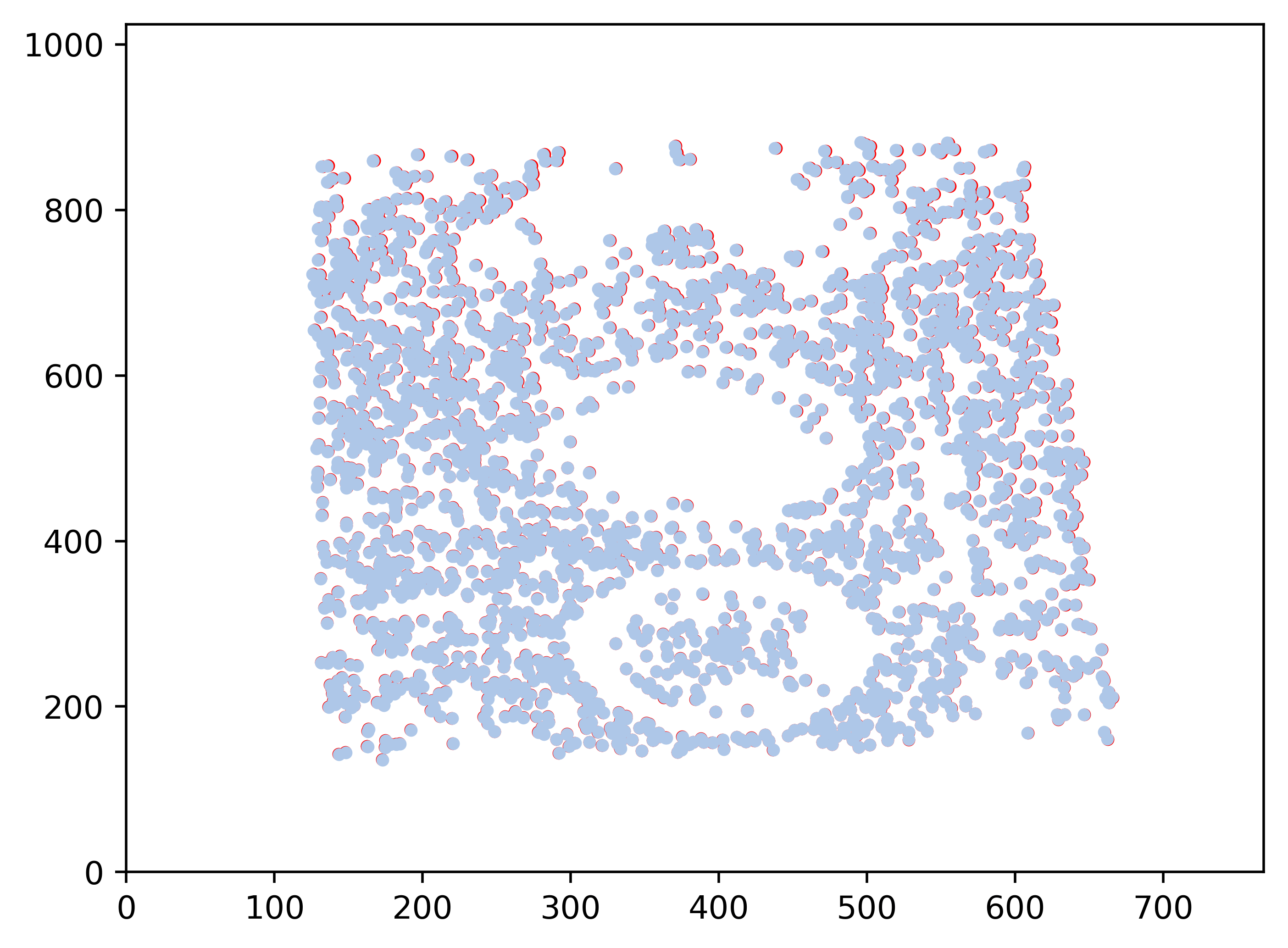}
\end{subfigure}
\begin{subfigure}{.24\textwidth}
	\centering
	\includegraphics[width=1.\linewidth, height=4cm]{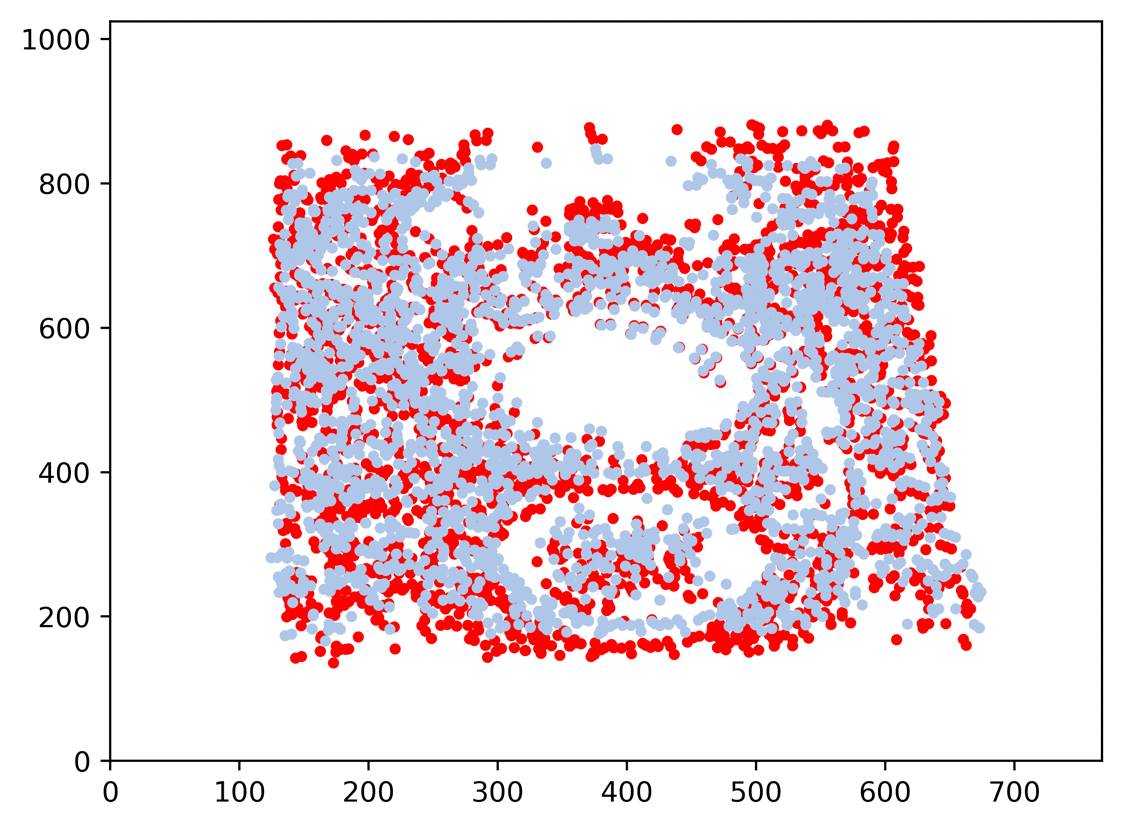}
\end{subfigure}

\caption{{\bf Qualitative PnP results.} The first two columns show the two images in the pair, the second of which we seek to estimate the pose from. In the third and fourth columns, we show the reprojection of the 3D point cloud after applying the rotation and translation predicted by our model and by EPnP+RANSAC, respectively. The red dots correspond to the ground-truth locations and the gray ones to our predictions. Note that our results match the ground truth much more closely than the baseline. Top to bottom: Florence, Reichstag, Notre-dame, Prague.
}
\label{fig:pnp}
\end{figure}

%% file: top.bbl
\begin{thebibliography}{10}

\bibitem{Yi18a}
Yi, K., Trulls, E., Ono, Y., Lepetit, V., Salzmann, M., Fua, P.:
\newblock {Learning to Find Good Correspondences}.
\newblock In: CVPR. (2018)

\bibitem{Papadopoulo00}
Papadopoulo, T., Lourakis, M.:
\newblock Estimating the jacobian of the singular value decomposition: Theory
  and applications.
\newblock In: ECCV. (2000)  554--570

\bibitem{Giles08}
Giles, M.:
\newblock {Collected Matrix Derivative Results for Forward and Reverse Mode
  Algorithmic Differentiation}.
\newblock In: Advances in Automatic Differentiation. (2008)  35--44

\bibitem{Ionescu15}
Ionescu, C., Vantzos, O., Sminchisescu, C.:
\newblock {Matrix backpropagation for Deep Networks with Structured Layers}.
\newblock (2015)

\bibitem{Tensorflow}
Abadi, M., Barham, P., Chen, J., Chen, Z., Davis, A., Dean, J., Devin, M.,
  Ghemawat, S., Irving, G., Isard, M., Kudlur, M., Levenberg, J., Monga, R.,
  Moore, S., Murray, D., Steiner, B., Tucker, P., Vasudevan, V., Warden, P.,
  Wicke, M., Yu, Y., Zheng, X.:
\newblock {Tensorflow: A System for Large-Scale Machine Learning}.
\newblock In: USENIX Conference on Operating Systems Design and Implementation.
  (2016)  265--283

\bibitem{PyTorch}
Paszke, A., Gross, S., Chintala, S., Chanan, G., Yang, E., DeVito, Z., Lin, Z.,
  Desmaison, A., Antiga, L., Lerer, A.:
\newblock {Automatic differentiation in PyTorch}.
\newblock In: NIPS Autodiff Workshop. (2017)

\bibitem{Jaderberg15}
Jaderberg, M., Simonyan, K., Zisserman, A., Kavukcuoglu, K.:
\newblock {Spatial Transformer Networks}.
\newblock In: NIPS. (2015)  2017--2025

\bibitem{Handa16}
Handa, A., Bloesch, M., Patraucean, V., Stent, S., McCormac, J., Davison, A.:
\newblock {Gvnn: Neural Network Library for Geometric Computer Vision}.
\newblock In: ECCV. (2016)

\bibitem{Murray16}
Murray, I.:
\newblock {Differentiation of the Cholesky Decomposition}.
\newblock arXiv Preprint (2016)

\bibitem{Longuet-Higgins81}
Longuet-Higgins, H.:
\newblock {A Computer Algorithm for Reconstructing a Scene from Two
  Projections}.
\newblock Nature \textbf{293} (1981)  133--135

\bibitem{Hartley97}
Hartley, R.:
\newblock {In Defense of the Eight-Point Algorithm}.
\newblock PAMI \textbf{19}(6) (June 1997)  580--593

\bibitem{Hartley00}
Hartley, R., Zisserman, A.:
\newblock {Multiple View Geometry in Computer Vision}.
\newblock Cambridge University Press (2000)

\bibitem{Nister03}
Nister, D.:
\newblock {An Efficient Solution to the Five-Point Relative Pose Problem}.
\newblock In: CVPR. (June 2003)

\bibitem{Fischler81}
Fischler, M., Bolles, R.:
\newblock {Random Sample Consensus: A Paradigm for Model Fitting with
  Applications to Image Analysis and Automated Cartography}.
\newblock Communications ACM \textbf{24}(6) (1981)  381--395

\bibitem{Torr00}
Torr, P., Zisserman, A.:
\newblock {{MLESAC}: A New Robust Estimator with Application to Estimating
  Image Geometry}.
\newblock CVIU \textbf{78} (2000)  138--156

\bibitem{Rousseeuw87}
Rousseeuw, P., Leroy, A.:
\newblock {Robust Regression and Outlier Detection}.
\newblock Wiley (1987)

\bibitem{Bian17}
Bian, J., Lin, W., Matsushita, Y., Yeung, S., Nguyen, T., Cheng, M.:
\newblock {GMS: Grid-Based Motion Statistics for Fast, Ultra-Robust Feature
  Correspondence}.
\newblock In: CVPR. (2017)

\bibitem{Raguram13}
Raguram, R., Chum, O., Pollefeys, M., Matas, J., Frahm, J.M.:
\newblock {USAC: A Universal Framework for Random Sample Consensus}.
\newblock PAMI \textbf{35}(8) (2013)  2022--2038

\bibitem{Zamir16}
Zamir, A.R., Wekel, T., Agrawal, P., Malik, J., Savarese, S.:
\newblock {Generic 3D Representation via Pose Estimation and Matching}.
\newblock In: ECCV. (2016)

\bibitem{Ummenhofer17}
Ummenhofer, B., Zhou, H., Uhrig, J., Mayer, N., Ilg, E., Dosovitskiy, A., Brox,
  T.:
\newblock {Demon: Depth and Motion Network for Learning Monocular Stereo}.
\newblock In: CVPR. (2017)

\bibitem{Lepetit09}
Lepetit, V., Moreno-noguer, F., Fua, P.:
\newblock {{EP$n$P}: An Accurate $o(n)$ Solution to the {P$n$P} Problem}.
\newblock IJCV (2009)

\bibitem{Kneip11}
Kneip, L., Scaramuzza, D., Siegwart, R.:
\newblock {A Novel Parametrization of the Perspective-Three-Point Problem for a
  Direct Computation of Absolute Camera Position and Orientation}.
\newblock In: CVPR. (2011)  2969--2976

\bibitem{Zheng13}
Zheng, Y., Kuang, Y., Sugimoto, S., Astrom, K., Okutomi, M.:
\newblock {Revisiting the {PnP} Problem: A Fast, General and Optimal Solution}.
\newblock In: ICCV. (2013)

\bibitem{Ferraz14}
Ferraz, L., Binefa, X., Moreno-noguer, F.:
\newblock {Very Fast Solution to the {PnP} Problem with Algebraic Outlier
  Rejection}.
\newblock In: CVPR. (2014)  501--508

\bibitem{Brachmann16b}
Brachmann, E., Krull, A., Nowozin, S., Shotton, J., Michel, F., Gumhold, S.,
  Rother, C.:
\newblock {DSAC -- Differentiable RANSAC for Camera Localization}.
\newblock ARXIV (2016)

\bibitem{Huang17a}
Huang, G., Liu, Z., Weinberger, K., van~der Maaten, L.:
\newblock {Densely Connected Convolutional Networks}.
\newblock In: CVPR. (2017)

\bibitem{Huang17b}
Huang, Z., Wan, C., Probst, T., Gool, L.V.:
\newblock {Deep learning on lie groups for skeleton-based action recognition}.
\newblock In: CVPR. (2017)  6099--6108

\bibitem{Law17}
Law, M., Urtasun, R., Zemel, R.S.:
\newblock {Deep spectral clustering learning}.
\newblock In: ICML. (2017)  1985--1994

\bibitem{Zhang98}
Zhang, Z.:
\newblock {Determining the Epipolar Geometry and Its Uncertainty: A Review}.
\newblock IJCV \textbf{27}(2) (1998)  161--195

\bibitem{Garro12}
Garro, V., Crosilla, F., Fusiello, A.:
\newblock {Solving the PnP Problem with Anisotropic Orthogonal Procrustes
  Analysis}.
\newblock In: 3DPVT. (2012)  262--269

\bibitem{Schonemann66}
Sch{\"o}nemann, P.:
\newblock {A Generalized Solution of the Orthogonal Procrustes Problem}.
\newblock Psychometrika \textbf{31}(1) (1966)  1--10

\bibitem{Kingma15}
Kingma, D., Ba, J.:
\newblock {Adam: {A} Method for Stochastic Optimisation}.
\newblock In: ICLR. (2015)

\bibitem{Xiao13}
Xiao, J., Owens, A., Torralba, A.:
\newblock {SUN3D: A Database of Big Spaces Reconstructed Using SfM and Object
  Labels}.
\newblock In: ICCV. (2013)

\bibitem{Strecha08b}
Strecha, C., Hansen, W., {Van~Gool}, L., Fua, P., Thoennessen, U.:
\newblock {On Benchmarking Camera Calibration and Multi-View Stereo for High
  Resolution Imagery}.
\newblock In: CVPR. (2008)

\bibitem{Cantzler}
Cantzler, H.:
\newblock {Random Sample Consensus (RANSAC)} (2005) CVonline.

\bibitem{Simpson97}
Simpson, D.:
\newblock {Introduction to Rousseeuw (1984) Least Median of Squares
  Regression}.
\newblock In: Breakthroughs in Statistics.
\newblock Springer (1997)  433--461

\bibitem{Li12c}
Li, S., Xu, C., Xie, M.:
\newblock {A Robust {O}(n) Solution to the Perspective-N-Point Problem}.
\newblock PAMI (2012)  1444--1450

\bibitem{Crivellaro17}
Crivellaro, A., Rad, M., Verdie, Y., Yi, K., Fua, P., Lepetit, V.:
\newblock {Robust 3D Object Tracking from Monocular Images Using Stable Parts}.
\newblock PAMI (2017)

\bibitem{Heinly15}
Heinly, J., Schoenberger, J., Dunn, E., Frahm, J.M.:
\newblock {Reconstructing the World in Six Days}.
\newblock In: CVPR. (2015)

\bibitem{Wu13}
Wu, C.:
\newblock {Towards Linear-Time Incremental Structure from Motion}.
\newblock In: 3DV. (2013)

\end{thebibliography}
